%% file: main.tex
\documentclass{article}

\usepackage{iclr2024_conference,times}
\iclrfinalcopy

\usepackage[utf8]{inputenc} 
\usepackage[T1]{fontenc}    
\usepackage{hyperref}       
\usepackage{url}            
\usepackage{booktabs}       
\usepackage{algorithm}
\usepackage[noend]{algpseudocode}
\usepackage{amsfonts}       
\usepackage{nicefrac}       
\usepackage{microtype}      
\usepackage{xcolor}         
\usepackage{xcolor,colortbl}
\usepackage{tabularx}
\usepackage{makecell}
\usepackage{multirow}
\usepackage{multicol}
\usepackage{comment}
\usepackage{adjustbox}
\usepackage{dashrule}
\usepackage[utf8]{inputenc}
\usepackage[titletoc]{appendix}
\usepackage{titletoc}
\usepackage{nicematrix}
\usepackage{xcolor}
\usepackage{subcaption}
\usepackage{tikz}
\usepackage{pgfplots}
\usepackage{enumitem}
\usepackage{wrapfig}
\usepackage{enumitem}
\usepackage{xcolor}
\usepackage{ulem}
\usepackage[utf8]{inputenc}
\usepackage{xcolor}
\usepackage{amsmath}
\usepackage{listings}
\usepackage{tcolorbox}
\usepackage{varwidth}

\usepackage{caption}
\usetikzlibrary{shapes,arrows,positioning,patterns,calc}

\definecolor{Gray}{gray}{0.93}
\definecolor{uclagold}{rgb}{1.0, 0.7, 0.0}
\definecolor{airforceblue}{rgb}{0.36, 0.54, 0.66}
\definecolor{rosegold}{rgb}{0.72, 0.43, 0.47}
\definecolor{pastelbrown}{rgb}{0.51, 0.41, 0.33}
\definecolor{isabelline}{rgb}{0.96, 0.94, 0.93}
\definecolor{macaroniandcheese}{rgb}{0.98, 0.89, 0.83}
\definecolor{wildblueyonder}{rgb}{0.85, 0.89, 0.95}
\definecolor{mediumtaupe}{rgb}{0.4, 0.3, 0.28}
\definecolor{bluegray}{rgb}{0.4, 0.6, 0.8}
\definecolor{celestialblue}{rgb}{0.29, 0.59, 0.82}
\definecolor{darkorange}{rgb}{1.0, 0.55, 0.0}
\definecolor{cadmiumred}{rgb}{0.89, 0.0, 0.13}
\definecolor{magnolia}{rgb}{0.97, 0.96, 1.0}
\definecolor{pastelblue}{rgb}{0.68, 0.78, 0.81}
\definecolor{persiangreen}{rgb}{0.0, 0.65, 0.58}
\definecolor{steelblue}{rgb}{0.27, 0.51, 0.71}
\definecolor{bluebell}{rgb}{0.64, 0.64, 0.82}
\definecolor{dimgray}{rgb}{0.41, 0.41, 0.41}
\definecolor{splashedwhite}{rgb}{1.0, 0.99, 1.0}
\definecolor{lavendergray}{rgb}{0.77, 0.76, 0.82}
\definecolor{lightgray}{rgb}{0.83, 0.83, 0.83}
\definecolor{lightlightgray}{gray}{0.9}
\definecolor{lavendermist}{rgb}{0.9, 0.9, 0.98}
\definecolor{lightgreen}{HTML}{f8fcf4}
\definecolor{lightblue}{HTML}{dfebf7}
\definecolor{zeroshot}{rgb}{0.9, 0.9, 0.9}
\definecolor{fourshot}{rgb}{0.8, 0.9, 0.8}
\definecolor{eightshot}{rgb}{0.8, 0.8, 0.9}
\definecolor{sixteenshot}{rgb}{0.9, 0.8, 0.8}
\definecolor{gtbackground}{RGB}{220, 255, 211}
\definecolor{corruptedbackground}{RGB}{255, 253, 193}
\definecolor{corruptedtext}{RGB}{239, 45, 21}

\newcommand{\corrupted}[1]{\textcolor{corruptedtext}{#1}}

\newlength{\alignmentresultlength}
\algnewcommand\algorithmicforeach{\textbf{for each}}
\algdef{S}[FOR]{ForEach}[1]{\algorithmicforeach\ #1\ \algorithmicdo}

\input{tikz_options}
\input{figures/llm_bias_template.tex}

\title{Understanding Alignment in Multimodal LLMs: A Comprehensive Study}

\author{%
  Elmira Amirloo\thanks{Authors contributed equally as first authors.},
  Jean-Philippe Fauconnier\footnotemark[1],
  Christoph Roesmann\footnotemark[1],\\
  \textbf{Christian Kerl}\thanks{Authors contributed equally.},
  \textbf{Rinu Boney}\footnotemark[2],
  \textbf{Yusu Qian},
  \textbf{Zirui Wang}, \\
  \textbf{Afshin Dehghan},
  \textbf{Yinfei Yang},
  \textbf{Zhe Gan},
  \textbf{Peter Grasch} \\
  Apple Inc.
}

\begin{document}
\maketitle
\pagestyle{plain} 
\input{sections/0_abstract}
\input{sections/1_introduction}
\input{sections/2_alignment}
\input{sections/3_data}
\input{sections/4_experiments}

\input{sections/5_conclusion}
\input{sections/acknowledgements}
\bibliographystyle{iclr2024_conference}
\bibliography{egbib}

\appendix
\include{sections/appendix}

\end{document}

%% file: figures/llm_bias_template.tex
	\pgfkeys{/tikz/.cd,
		yshiftbox/.initial=0, 
		yshiftbox/.get=\yshiftbox, 
		yshiftbox/.store in=\yshiftbox, 
		imgname/.initial=0, 
		imgname/.get=\imgname, 
		imgname/.store in=\imgname, 
		prompttext/.initial=0, 
		prompttext/.get=\prompttext, 
		prompttext/.store in=\prompttext, 
		sfttext/.initial=0, 
		sfttext/.get=\sfttext, 
		sfttext/.store in=\sfttext, 
		bdhsatext/.initial=0, 
		bdhsatext/.get=\bdhsatext, 
		bdhsatext/.store in=\bdhsatext, 
		bdhsbtext/.initial=0, 
		bdhsbtext/.get=\bdhsbtext, 
		bdhsbtext/.store in=\bdhsbtext, 
		bdhsctext/.initial=0, 
		bdhsctext/.get=\bdhsctext, 
		bdhsctext/.store in=\bdhsctext, 
		bdhsdtext/.initial=0, 
		bdhsdtext/.get=\bdhsdtext, 
		bdhsdtext/.store in=\bdhsdtext, 
		bdhsnatext/.initial=0, 
		bdhsnatext/.get=\bdhsnatext, 
		bdhsnatext/.store in=\bdhsnatext, 
		bdhsnbtext/.initial=0, 
		bdhsnbtext/.get=\bdhsnbtext, 
		bdhsnbtext/.store in=\bdhsnbtext, 
		povidatext/.initial=0, 
		povidatext/.get=\povidatext, 
		povidatext/.store in=\povidatext, 
		povidbtext/.initial=0, 
		povidbtext/.get=\povidbtext, 
		povidbtext/.store in=\povidbtext, 
	}

	\tikzstyle{exampleframe} = [rounded corners]
	\tikzstyle{textblock} = [rectangle, draw,  minimum size=0.2cm, inner sep=0.1cm, rounded corners]
	\tikzstyle{imgblock} = [minimum width=3cm, minimum height=1cm, inner sep=0.0cm]
	
	\tikzstyle{promptblock} = [rectangle, text width=2.9cm]
	\tikzstyle{sftblock} = [textblock,  fill=gtbackground, text width=10.0cm]
	\tikzstyle{rejectedblock} = [textblock,  fill=corruptedbackground, text width=4cm]
	
	\tikzstyle{bdhsblock} = [textblock,  text width=4.75cm]
	
	\newcommand{\sft}{\textbf{SFT ground truth: }}

    \newcommand{\bdhsa}{$\mathbf{\text{\bf{BDHS}}_{\text{\bf{attn}}}, \boldsymbol\rho_{\text{th}}{=}0}$ \textbf{(ours):} }
	\newcommand{\bdhsb}{$\mathbf{\text{\bf{BDHS}}_{\text{\bf{attn}}}, \boldsymbol\rho_{\text{th}}{=}0}.98$ \textbf{(ours):} }
	\newcommand{\bdhsc}{$\mathbf{\text{\bf{BDHS}}_{\text{\bf{attn}}}, \boldsymbol\rho_{\text{th}}{=}0.99}$ \textbf{(ours):} }
	\newcommand{\bdhsd}{$\mathbf{\text{\bf{BDHS}}_{\text{\bf{attn}}}, \boldsymbol\rho_{\text{th}}{=}1}$ \textbf{(ours):} }

	\newcommand{\bdhsna}{$\mathbf{\text{\bf{BDHS}}_{\text{\bf{noise}}}, N{=}300}$ \textbf{(ours):} }
	\newcommand{\bdhsnb}{$\mathbf{\text{\bf{BDHS}}_{\text{\bf{noise}}}, N{=}500}$ \textbf{(ours):} }
	
	\newcommand{\povida}{\textbf{POVID-style image distortion, $\mathbf{N{=}300}$: }}
	\newcommand{\povidb}{\textbf{POVID-style image distortion, $\mathbf{N{=}500}$: }}

	\newcommand{\placeexample}[1][]{
	\tikzset{#1}
	\begin{scope}[auto, local bounding box=scope1, yshift=\yshiftbox, font=\fontsize{7pt}{7pt}\selectfont]
		\node[imgblock] at (0,0) (img) { \includegraphics[width=3cm] {figures/\imgname.jpg} };
		
		\node[promptblock, below=0.0cm of img] (prompt) {\textbf{Prompt:} \prompttext};	
		
		\node[sftblock, right=0.5cm of img.north east] (sft) {\sft \sfttext};	
		
		\node[bdhsblock, yshift=-0.1cm, anchor=north west] (bdhsa) at (sft.south west)  {\bdhsa \bdhsatext};	
		\node[bdhsblock, below=0.1cm of bdhsa] (bdhsb) {\bdhsb \bdhsbtext};	
		\node[bdhsblock, below=0.1cm of bdhsb] (bdhsc) {\bdhsc \bdhsctext};	
		\node[bdhsblock, below=0.1cm of bdhsc] (bdhsd) {\bdhsd \bdhsdtext};	
		
		\node[bdhsblock, yshift=-0.1cm, anchor=north east] (bdhsna) at (sft.south east)  {\bdhsna \bdhsnatext};	
		\node[bdhsblock, below=0.1cm of bdhsna] (bdhsnb) {\bdhsnb \bdhsnbtext};	
		
		\node[bdhsblock, below=0.1cm of bdhsnb] (povida)  {\povida \povidatext};	
		\node[bdhsblock, below=0.1cm of povida] (povidb) {\povidb \povidbtext};	
		
		\path let \p1=(bdhsd.south),\p2=(povidb.south) in coordinate (miny1) at (\x1, {min(\y1,\y2)});
		\path let \p1=(miny1),\p2=(prompt.south) in coordinate (miny) at (\x1, {min(\y1,\y2)});
		\coordinate (maxy) at (sft.north);
		
		\coordinate (top_left) at (img.north west |- maxy);
		\coordinate (bottom_right) at (povidb.south east);
		\draw[exampleframe] ([shift={(0.0,-0.15)}]miny) -| ([shift={(-0.15,0.15)}]top_left) -| ([shift={(0.15,-0.15)}]bottom_right |- miny) -- ([shift={(0,-0.15)}]miny);
	\end{scope}
 	}

%% file: sections/0_abstract.tex
\begin{abstract}
Preference alignment has become a crucial component in enhancing the performance of Large Language Models (LLMs), yet its impact in Multimodal Large Language Models (MLLMs) remains comparatively underexplored. Similar to language models, MLLMs for image understanding tasks encounter challenges like hallucination. In MLLMs, hallucination can occur not only by stating incorrect facts but also by producing responses that are inconsistent with the image content. A primary objective of alignment for MLLMs is to encourage these models to align responses more closely with image information. Recently, multiple works have introduced preference datasets for MLLMs and examined different alignment methods, including Direct Preference Optimization (DPO) and Proximal Policy Optimization (PPO). However, due to variations in datasets, base model types, and alignment methods, it remains unclear which specific elements contribute most significantly to the reported improvements in these works. In this paper, we independently analyze each aspect of preference alignment in MLLMs. We start by categorizing the alignment algorithms into two groups, offline (such as DPO), and online (such as online-DPO), and show that combining offline and online methods can improve the performance of the model in certain scenarios. 
We review a variety of published multimodal preference datasets and discuss how the details of their construction impact model performance. Based on these insights, we introduce a novel way of creating multimodal preference data called Bias-Driven Hallucination Sampling (BDHS) that needs neither additional annotation nor external models, and show that it can achieve competitive performance to previously published alignment work for multimodal models across a range of benchmarks.
\end{abstract}

%% file: sections/1_introduction.tex
\section{Introduction}\label{section:introduction}
Recent advancements in Multimodal Large Language Models (MLLMs) have significantly improved our understanding of vision-language tasks. By integrating visual signals with Large Language Models (LLMs), these models have demonstrated enhanced capabilities in multimodal understanding, reasoning, and interaction \citep{Liu_2023,Liu_2024b,Bai_2023b,Mckinzie_2024}.

Typically, MLLMs are pre-trained on large image-text datasets to develop foundational multimodal knowledge and skills, then undergo post-training  for conversational capabilities, instruction following, helpfulness, and safety. Despite rapid advancements in recent years, significant challenges persist.

A notable problem is the tendency of MLLMs to produce responses that are not factually grounded in the visual input, commonly referred to as hallucinations, leading to inaccuracies such as incorrect descriptions of non-existent visual elements~\citep{liu2023mitigating,Cui_2023}. This undermines the trustworthiness of MLLMs in many practical applications.

Preference alignment methods have proven effective in reducing hallucinations and generating responses more closely aligned with human preferences for LLMs \citep{Zhao_2023e,Rafailov_2023,Azar_2024,Guo_2024,Yuan_2024,Ahmadian_2024,Tang_2024}.  These methods utilize pairwise preference data to fine-tune the model, which can be based on Reinforcement Learning from Human Feedback (RLHF) \citep{Christiano_2017,Stiennon_2020}, Direct Alignment from Preferences (DAP) \citep{Rafailov_2023,Azar_2024,Zhao_2023e}, or Online Direct Alignment from Preferences (Online-DAP)~\citep{Yuan_2024,Guo_2024}.

While alignment in LLMs has been extensively studied, alignment for MLLMs has not yet been investigated to the same extent. \citet{Sun_2023} and \citet{Zhou_2024} aligned LLaVA 1.5 \citep{Liu_2023b} using Proximal Policy Optimization (PPO) and Direct Preference Optimization (DPO), respectively, while \citet{Li_2023d} and \citet{Yu_2023} employed DPO and its variations to align Qwen-VL \citep{Bai_2023b} and Muffin \citep{Yu_2023b} models. Notably, besides different alignment strategies and often different base models, all these works also introduce novel preference datasets for alignment with various sizes, collection, and generation schemes. As a result, while each of these studies offers valuable insights into alignment for MLLMs, it can sometimes be difficult to strongly attribute reported improvements to the individual proposed choices.

\begin{figure}[t!]
   \centering
    \resizebox{0.85\textwidth}{!}{\input{figures/main_figure}}
     \caption{Illustration of the alignment objective for multimodal LLMs. The alignment problem is formulated as reward maximization w.r.t.~the parameters of the policy $\pi_\theta$. A regularization term provides a positive reward for staying close to a reference policy $\pi_\text{ref}$. For DAP methods, the reward maximization problem is transformed to a supervised learning problem by using a closed-form solution for preference pairs.  Both online and offline methods draw from a prompt distribution $ x \sim p$ that comprises text and image prompts. However, for the sampling distribution $(y^+, y^-) \sim \mu$, offline methods draw from a fixed preference dataset while online ones sample from the policy, $\pi_\theta$, and rank/score with an annotator or reward model.}
\label{fig:main_figure}
\vspace{-0.20cm}
\end{figure}
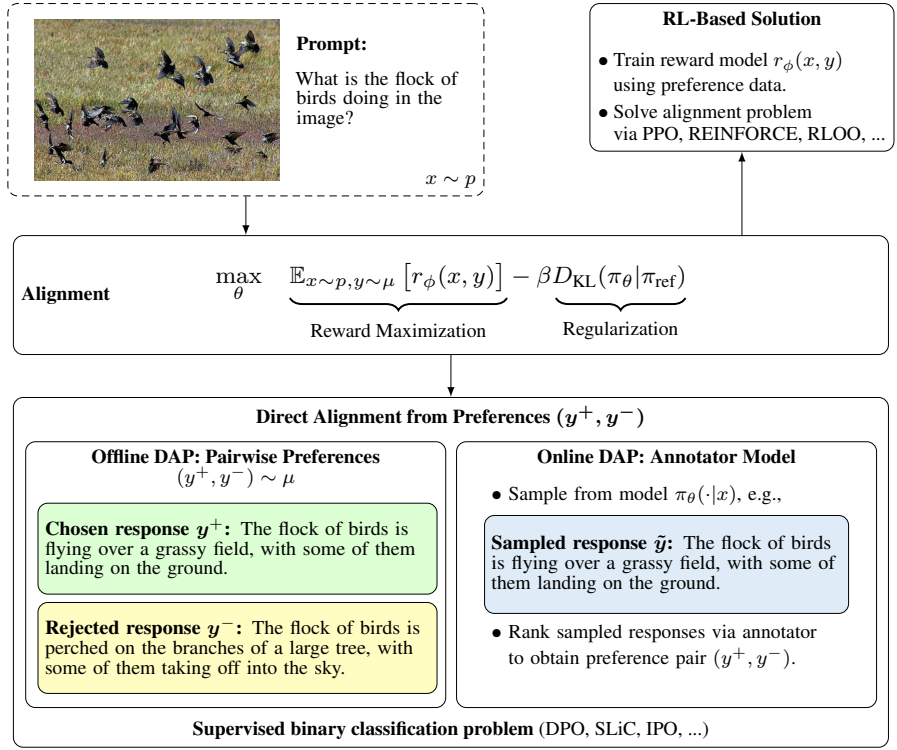

In this paper, we examine each component of multimodal alignment independently.
First, we categorize alignment methods into two types (see Figure \ref{fig:main_figure}): offline methods, which utilize preference pairs collected prior to training (e.g., DPO), and online methods, which involve sampling from the model during policy optimization (e.g., RLHF and Online-DAP). We conduct a comprehensive study over popular online and offline alignment methods, all aligning the popular LLaVA 1.6 model~\citep{Liu_2024b} using a fixed data regiment and study their benefits and shortcomings. To our knowledge, this is the first time that such study is conducted with MLLMs.

Further, we study the different methods for building pairwise preferences using public datasets. We break down the main components of preference data into three parts: prompts, chosen responses and rejected responses (Table \ref{table:sota_datasets}). For each of those components, we investigate how their source, diversity, and quality can affect the resulting alignment. Additionally, we examine how the size of the alignment dataset impacts downstream performance.

Based on our comprehensive ablations, we identify a few key desiderata in alignment strategies for MLLMs and introduce a simple, novel preference data sampling scheme we call Bias-Driven Hallucination Sampling (BDHS). Despite not utilizing any human annotation nor the input of any external teacher model such as GPT4-V, we show that BDHS can achieve competitive performance against even much larger preference datasets constructed under different regimes.

%% file: figures/main_figure.tex
\begin{tikzpicture}[font=\fontsize{8pt}{8pt}\selectfont]

\tikzstyle{block} = [rectangle, draw, minimum width=3cm, minimum height=1cm, text width=3cm, rounded corners];

\node[] at (0, 0) (img) { \includegraphics[width=3.5cm] {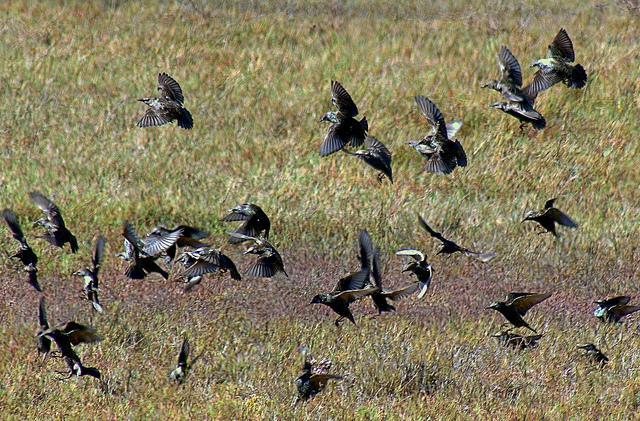}};
\node[anchor=west, text width=2.3cm, align=left] at (img.east) (prompt) {What is the flock of birds doing in the image?};
\node[anchor=south west] at (prompt.north west) {\textbf{Prompt:}};

\node[densely dashed, block, minimum width=6.8cm, minimum height=2.75cm, yshift=0.0cm] (input) at ($(img.west)!0.5!(prompt.east)$) {};
\node[anchor=south east] at (input.south east) {$x \sim p$};

\node[draw, minimum width=4cm, anchor=north west, rounded corners, xshift=1.5cm, align=left] at (input.north east) (rlbased) {
	\\[0.5cm]
				$\bullet$  Train reward model $r_\phi(x, y)$ \\[1pt]  \phantom{$\bullet$} using preference data. \\[3pt]
				$\bullet$ Solve alignment problem \\[1pt] \phantom{$\bullet$} via PPO, REINFORCE, RLOO, ...
};
\node[anchor=north] at (rlbased.north) {\textbf{RL-Based Solution}};

\coordinate (topcenter) at ($(input.west)!0.5!(rlbased.east)$);

\node[scale=1.2, align=center, text width=0.6\textwidth, below=2.0cm of topcenter] (rewardmax)  {
	\begin{equation*}
		\max_{\theta} \quad  \mathbb{E}_{x \sim p, y \sim \mu} \left[ r_\phi(x, y)\right] - \beta  D_{\text{KL}}(\pi_\theta| \pi_{\text{ref}})
	\end{equation*}
};

\draw[decorate,decoration={brace,amplitude=5pt,mirror,raise=4pt}, thick,yshift=0pt] 
($ (rewardmax) + (-2.3,-0.25) $) -- +(3.1, 0) node [midway,yshift=-0.6cm] (bracerew) {Reward Maximization};
\draw[decorate,decoration={brace,amplitude=5pt,mirror,raise=4pt}, thick,yshift=0pt] 
($ (rewardmax) + (1.5,-0.25) $) -- +(1.85, 0) node [midway,yshift=-0.6cm] (bracerew) {Regularization};

\node[block, minimum width=12.5cm, minimum height=1.7cm, yshift=-0.35cm] (alignment) at (rewardmax) {};
\node[anchor=west] at (alignment.west) {\textbf{Alignment}};

\draw[-latex] (input) -- (input |- alignment.north);
\draw[-latex] (alignment.north -| rlbased) -- (rlbased);

\node[draw, rounded corners, align=left, inner sep=3mm, minimum width=12.5cm,  minimum height=5.0cm, anchor=north west, yshift=-0.6cm] (dap) at (alignment.south west){ };
\node[anchor=north] at (dap.north) {\textbf{Direct Alignment from Preferences $\boldsymbol{(y^+, y^-)}$}};

\draw[-latex] (alignment) -- (dap);


\node[draw, rounded corners, inner sep=0mm, minimum width=6.0cm, minimum height=3.8cm,
      anchor=north west, at={(dap.north west)}, xshift=5pt, yshift=-18pt] (offline) {};

\node[draw, rounded corners, inner sep=1mm, text width=5.5cm, minimum height=1.3cm,
anchor=north west, at={(offline.north west)}, fill=gtbackground, xshift=5pt, yshift=-25pt] (chosen) {
				      \textbf{Chosen response $\boldsymbol y^+$:} The flock of birds is flying over a grassy field, with some of them landing on the ground.
		};
\node[draw, rounded corners, inner sep=1mm, text width=5.5cm, minimum height=1.3cm,
      anchor=north west, at={(chosen.south west)}, xshift=0pt, yshift=-3pt, fill=corruptedbackground] (rejected) {
				          \textbf{Rejected response $\boldsymbol y^-$:} The flock of birds is perched on the branches of a large tree, with some of them taking off into the sky.
		      };
\node[anchor=north, align=center] (label_pairwise_preferences) at (offline.north) {\textbf{Offline DAP: Pairwise Preferences} \\ $(y^+, y^-) \sim \mu$};

\node[draw, rounded corners, inner sep=0mm, minimum width=6.0cm, minimum height=3.8cm,
      anchor=north west, at={(offline.north east)}, xshift=4pt, yshift=0pt] (online) {};
\node[anchor=north, align=center] (label_reward) at (online.north) {\textbf{Online DAP: Annotator Model}};

\node[text width=5cm, anchor=north, rounded corners, yshift=-0.5cm, align=left] (sampletext) at (online.north) {
	$\bullet$  Sample from model $\pi_\theta(\cdot | x)$, e.g., 
};

\node[draw, rounded corners, inner sep=1mm, text width=5.0cm, minimum height=1.4cm,
anchor=north, fill=lightblue, yshift=-30pt] at (online.north) (sampled) {
		\textbf{Sampled response $\boldsymbol{\tilde{y}}$:} The flock of birds is flying over a grassy field, with some of them landing on the ground.
};

\node[text width=5cm, anchor=north, rounded corners, yshift=-2.5cm, align=left] (sampletext2) at (online.north) {
	$\bullet$ Rank sampled responses via annotator \\[1pt] \phantom{$\bullet$}  to obtain preference pair $(y^+, y^-)$.
};

\node[anchor=south] at (dap.south) {\textbf{Supervised binary classification problem} (DPO, SLiC, IPO, ...)};
\end{tikzpicture}

%% file: sections/2_alignment.tex
\section{Alignment}\label{section:alignment}

Preference alignment uses pairwise preference data. Each pair is linked to a text prompt, denoted as $x_{\text{text}}$, and an associated image, $x_{\text{img}}$, together forming the input \( x = (x_{\text{img}}, x_{\text{text}}) \). The responses include a preferred one, \( y^+ \), and a non-preferred or rejected one, \( y^- \). See Section~\ref{section:preference_datasets} for a more thorough discussion of these components.
In this section, we focus on the various ways that preference dataset, $\mathcal{D} = \{(x, y^+, y^-)\}_{i=1}^{N}$, is used by alignment approaches.

\subsection{Reinforcement Learning from Human Feedback (RLHF)}

RLHF was the initial method used for alignment \citep{Christiano_2017,Stiennon_2020}, involving the training of a reward model (RM) from pairwise preferences and then optimizing a policy using the RM via reinforcement learning (RL). In RLHF, a reward model is initially trained on the preference pairs as described in \citet{Stiennon_2020}. The training of this reward model uses a straightforward cross-entropy loss, treating the binary choice -- preferred or rejected -- as a categorical label. The objective function for training the reward model, $r_\phi$,  is as follows:
\begin{equation}
L_{\text{RM}} = - \log \left( \sigma \left( \log \left( r_{\phi}(x, y^+) - r_{\phi}(x, y^-) \right) \right) \right)\,,
\end{equation}
where $\sigma$ is the logistic function.

Next, the model (i.e., policy), $\pi_{\theta}$, is fine-tuned through RL using the trained reward model to optimize the following objective:
\begin{equation}
\max_{\pi_\theta} \mathbb{E}_{x \sim D, y \sim \pi_\theta(y|x)} \left[ r_\phi(x, y) - \beta D_{\text{KL}}(\pi_\theta(y|x) | \pi_{\text{ref}}(y|x)) \right]\,.
\label{eq:rl-objective}
\end{equation}
An additional KL penalty term $D_{\text{KL}}(\cdot)$ is incorporated to discourage significant deviations of $\pi_{\theta}$ from the initial model, $\pi_{\text{ref}}$ \citep{Stiennon_2020}, and $\beta$ is a hyperparameter which adjusts the effect of this term in the overall objective.

Since the RM is learned in all RL-based approaches, even if it is a decent approximation of what we are truly optimizing for, such as human preferences, it remains an imperfect approximation. Previous work has shown that if not handled carefully, over-optimizing for the RM can hurt the performance of the aligned model \citep{Gao_2023}. This complexity adds a significant layer of challenge to RLHF methods.

Different RL algorithms apply unique strategies to optimize the RL objective (Equation~\ref{eq:rl-objective}). In Section~\ref{section:rl-experiments} we investigate the complexities of RL-based alignment for MLLMs, examining how different algorithms affect model performance. Specifically, we evaluate the impact of using PPO \citep{Schulman_2017,Stiennon_2020,Ouyang_2022} and REINFORCE Leave-One-Out (RLOO) \citep{Williams_1992, Ahmadian_2024} in comparison to other alignment methods.

\subsection{Direct Alignment from Preference}
\label{sec:offline_dap}

This family of approaches directly utilizes preference data, $D$, to optimize the policy, $\pi_{\theta}$. By eliminating the need to train a reward model, these methods significantly simplify the preference optimization pipeline. Furthermore, the gradient of all objectives can be precisely computed, distinguishing these methods from traditional RLHF approaches. The most widely used objective in MLLM alignment is DPO \citep{Rafailov_2023} (Equation \ref{eq:dpo_loss}). We have conducted the majority of our experiments using DPO to ensure comparability with other studies in MLLM alignment. In Section~\ref{section:offline}, we also examine DPO alongside two other common offline methods, IPO \citep{Azar_2024} and SLiC \citep{Zhao_2023e}. For a unified derivation of common direct alignment methods refer to \cite{tang2024generalized} and for brevity, we only recap the DPO loss function:
    \begin{equation} \label{eq:dpo_loss}
      L_{\text{DPO}}(\pi_\theta; \pi_{\text{ref}}) =\mathbb{E}_{(x, 
 y^+, y^-) \sim D} \left[ -\log \sigma \left( \beta \log \frac{\pi_{\theta}(y^+|x)\pi_{\text{ref}}(y^-|x)}{\pi_{\text{ref}}(y^+|x)\pi_{\theta}(y^-|x)} \right)\right]\,.
    \end{equation}
 
 For simplicity, we will omit the dependency on $\pi_{\text{ref}}$ from subsequent equations. 
 
 It is important to note that most preference datasets are not derived from the model being aligned and are collected offline. Even when the data is constructed based on the model that is undergoing alignment, the samples encountered during training do not account for changes in the model over training. This leads to a distribution shift between the model that generated the data and the model being aligned, which can be considered a disadvantage of these methods.

\subsection{Online Direct Alignment from Preference}
\label{sec:online_dap}

Recently, a new family of algorithms has been proposed for aligning LLMs. These methods do not train a separate reward model. Instead, they employ either the model that is being aligned \citep{Yuan_2024} or a different LLM \citep{Guo_2024} to obtain online feedback to create preference pairs. These pairs are then used to optimize the objective function via for example DPO. This approach eliminates the complexity of training a separate reward model while still taking advantage of online samples from the model, thereby avoiding distribution shifts.

We explore the use of LLaVA 1.6-34B \citep{Liu_2024b} as annotator to generate online preference pairs, motivated by its strong performance on a multitude of multimodal benchmarks.
Additionally, we investigate a hybrid approach that combines online and offline approaches. This method involves sampling from the offline preference data with a probability \( p \), \( (y^+, y^-) \), and sampling from the model with a probability \( 1-p \), \( (\tilde{y}^+, \tilde{y}^-) \). Equation~\eqref{eq:mixed_dpo} details this approach.
\begin{equation} \label{eq:mixed_dpo}
\begin{aligned}
L_{\text{Mixed-DPO}}(\pi_\theta) &= \mathbb{E}_{\substack{(x, y^+, y^-) \sim D \\ (\tilde{y}^+, \tilde{y}^-) \sim \pi_\theta}} \left[ \alpha L_{\text{DPO}}(y^+, y^-, x ; \pi_\theta) + (1-\alpha) L_{\text{DPO}}(\tilde{y}^+, \tilde{y}^-, x; \pi_\theta)\right]\,, \\
\end{aligned}
\end{equation}
where $\alpha \sim \text{Bernoulli}(p)$. In our experiments we use $p = 0.5$. This algorithm is similar to techniques used in off-policy RL methods like Q-learning \citep{hester2018deep}, where a replay buffer includes samples from both the model and expert demonstrations. We found this approach particularly effective when the online and offline methods have complementary effects on the model's final performance.

%% file: sections/3_data.tex
\section{Multimodal Preference Data}\label{section:preference_datasets}

Multimodal preference data is usually constructed by using responses generated by one or more MLLMs, typically excluding the model being aligned. In this section, we explain the structure and components of the data and its generation as well as analyze these elements within the context of recently published datasets. While preference data is often discussed in the offline setting of collected and stored datasets, online approaches such as online DPO inherently generate preference data online, following the same structure and components as introduced below.

\subsection{Elements of Preference Data}

Examples of preference data for multimodal alignment generally involve three different elements, which we categorize here. Also see Figure~\ref{fig:main_figure} for how these components are used during alignment.

\begin{itemize}[nosep,leftmargin=0.5cm]
    \item \textbf{Prompts} comprise the text instruction and the corresponding image. They can be general (e.g., \textit{What is the title of the book mentioned in the image?}) or domain specialized (e.g., \textit{You are a driving assistant. Based on the current image, what is the best action to take when you are driving on the road?}).
    \item \textbf{Chosen responses} are responses that a well aligned policy model shall prefer over the rejected responses. They may comprise hallucination free responses, accurate factual information and instruction-following responses.
    \item \textbf{Rejected responses} shall score low / shall not be preferred over chosen responses, given a well aligned policy.
\end{itemize}

\subsection{Background}
\label{sec:recent_works}

Recently, multiple works have proposed preference datasets for MLLMs. The preference pairs are constructed based on human annotations or derived from synthetic techniques.
In Table~\ref{table:sota_datasets}, we report recently published datasets.

\input{tables/datasets_related_works}

\textbf{Human annotations}~~In LLaVA-RLHF \citep{Sun_2023}, authors collect human preferences by asking crowdworkers faced with two responses to prioritize responses that exhibit the best multimodal alignment and minimize hallucinations. Using this process, the authors built a 10k preferences dataset. The prompts are from LLaVA-Instruct-150k \citep{Liu_2023}, while responses are sampled from LLaVA base model. As is common for many other preferences datasets, the source of the images is COCO \citep{Lin_2014}.

In RLHF-V,~\citet{Yu_2023} propose to collect human preferences at the segment level by asking annotators to correct mistakes in model responses. As a result, the preference is expressed on token spans and not at the response level. Used in conjunction with a token-weighted DPO training, authors reported a reduced hallucinations level. Prompts and images are originally from the UniMM-Chat SFT dataset introduced by~\citet{Yu_2023b}, and responses sent for annotation and correction are sampled from Muffin \citep{Yu_2023b}. In the latest iterations of this dataset, both the sources of data and the samples have become more diverse.

\textbf{Synthetic annotations}~~In DRESS \citep{Chen_2023e}, authors introduce NLF, a 63k pairwise preference dataset built from LLaVA-Instruct-150k images and prompts. Authors leverage GPT-4 to provide critique and refinement on the responses of their in-house DRESS model. In VLFeedback \citep{Li_2023d}, authors sample responses from a pool of 12 multimodal MLLMs --- including GPT-4V, the LLaVA 1.5 series models, Qwen-VL-Chat, InstructBLIP --- on a pool of datasets. This synthetic approach allows to importantly scale up both the number of examples generated and the diversity of the response. Totalling 80k samples, the prompts and images are from 9 diverse datasets such as LLaVA-Instruct-150k (COCO images), SVIT \citep{Zhao_2023b} (Visual Genome images), LLaVAR \citep{Zhang_2023b} (LAION-5B images), etc. GPT-4V is used to grade and select the best answer among the MLLM responses.

In POVID,~\citet{Zhou_2024} propose to generate dispreferences from a ground truth dataset directly, removing the need for ranking responses. Specifically, 17k examples are selected randomly from the LLaVA-Instruct-150k dataset, with the original answers assumed to be a preferred response, while the dispreferred response is derived by prompting GPT-4V to introduce mistakes in the preferred response. The authors make a distinction between hallucinations added into captioning tasks (e.g., wrong visual attributes, incorrect relationships) and into reasoning tasks (e.g., wrong reasoning based on a hallucination). They explicitly prompt GPT-4V to incorporate these specific types of inaccuracies into the chosen responses.

\textbf{Preference signal}~~In addition to considering these works in terms of using human or synthetic supervision as above, another key distinction among them is how the preference signal is composed. The preference signal is obtained by \textbf{ranking} in LLaVA-RLHF \citep{Sun_2023} and VLFeedback \cite{Li_2023d}: responses are sampled, then ranked by humans or GPT-4V. For other works, the preference signal is a \textbf{construction} obtained either by correcting/refining responses to produce chosen responses, as done by RLHF-V \citep{Yu_2023} and DRESS \citep{Chen_2023e}, or by corrupting responses to produce rejected ones as POVID \citep{Zhou_2024}.

\subsection{POVID-Style Image Distortion}
\label{sec:povid}

A particular reason for hallucinations in MLLMs is that the underlying language model is pre-trained in isolation. 
Therefore, the model tends to prefer memorization from training or textual context $x_{\text{text}}$ over the associated image $x_{\text{img}}$ \citep{Zhou_2024}. 
As part of the POVID work, \citet{Zhou_2024} suggests to trigger inherent hallucination patterns directly by presenting noisy images $\tilde{x}_{\text{img}}$ to the model when generating the non-preferred response $\tilde{y}^-$. 
Hereby, each generated token $t$ in $\tilde{y}^-$ is conditioned on the prior tokens from the preferred response $y^+_{<t}$, i.e., $\pi_\theta(\tilde{y}^-_t|\tilde{x}, y^+_{<t})$ with modified input $\tilde{x}=(x_{\text{text}}, \tilde{x}_{\text{img}})$ (teacher-forcing). The tilde notation emphasizes that the response is driven by the model with restricted access to the image.
$\tilde{x}_{\text{img}}$ is created through a diffusion process that incrementally adds Gaussian noise to the image $x_{\text{img}}$ for a predefined number of steps $N$, which is set to 500 by default (see Section~\ref{sec:appendix:diffusion} for implementation details).

The online response $\tilde{y}^-$ is combined with existing preference pairs $(y^+, y^-)$ by averaging pairwise losses\footnote{The loss presented here refers to the version in the public source code of POVID \citep{Zhou_2024} which averages the individual DPO losses, opposed to the paper \citep{Zhou_2024} that combines the non-preferred responses by a weighted sum of their log-probabilities.}, i.e.,
\begin{equation} \label{eq:avg_dpo}
\begin{aligned}
L_{\text{Avg-DPO}}(\pi_\theta) &= \mathbb{E}_{\substack{(x, y^+) \sim D \\  \tilde{y}^- \sim \pi_\theta(\tilde{x}_{\text{img}}, y^+)}} \left[ \gamma L_{\text{DPO}}(y^+, y^-, x ; \pi_\theta) + (1-\gamma) L_{\text{DPO}}(y^+, \tilde{y}^-, x; \pi_\theta)\right]\,, \\
\end{aligned}
\end{equation}
with $\gamma=0.5$.
While this method has desirable characteristics, such as not requiring external teacher models or human annotation to construct preference pairs, as well as generating samples that are at least partially informed by the policy under alignment, it carries some notable drawbacks.

\cite{Zhou_2024} shows that selecting too few diffusion steps can yield insufficient corruption, whereas too many diffusion steps negatively impacts the generated responses, as the model mainly identifies noise respective to \textit{pixels}. In our experiments, we further found that the amount of noise to be added for each image might be difficult to control universally with a single parameter. 

Additionally, \cite{Zhou_2024} suggests that the proposed teacher-forcing strategy can help to yield samples that exhibit corruption only in few, key tokens most informed by the visual content, thus focusing the feedback signal for alignment. However, since the method operates token by token, we found that this can introduce non-sensical responses, e.g., only corrupting some parts of multi-token noun phrases. Such constructions would presumably already achieve a low generation probability, limiting the learning signal for the DPO-based alignment.
An example showing non-sensical responses is provided in Section~\ref{sec:bdhs_example} with Figure~\ref{fig:llm_bias_example}.

\subsection{Bias-Driven Hallucination Sampling (BDHS)} 
\label{section:BDHS}

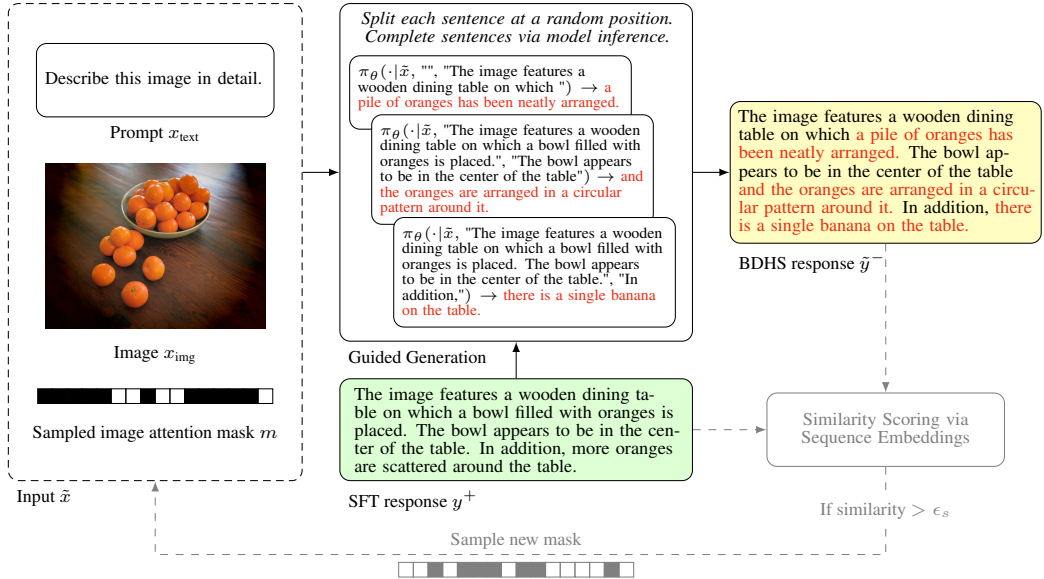
\begin{figure}
    \centering
   \resizebox{0.99\textwidth}{!}{\input{figures/bdhs_overview_oranges}}
    \caption{Overview of the BDHS method including the optional iterative variant in gray. For each re-generation, both the image attention mask and the sentence split positions are resampled. The image is taken from the LLaVA Instruct dataset.}
    \label{fig:bdhs_overview}
\end{figure}

Inspired by Section~\ref{sec:povid}, we aim to address its main identified shortcomings.
First, we propose to rethink the method of corrupting the signal from the input image from a pixel-based approach to one that limits access in the latent space via attention masking, which we argue more directly achieves the underlying motivation of triggering the inherent bias of the underlying language model.

Second, we introduce a new reference-guided generation strategy that allows corrupted responses to remain largely true to the chosen response while still introducing meaningful divergence, without introducing non-sensical continuations introduced by token-based teacher forcing.

Third, we use an off-the-shelf sentence embedding to verify that the generated rejected response is meaningfully distinct from the original reference to focus the resulting feedback signal on hallucinations over mere stylistic difference.

We refer to our novel technique as \emph{Bias-Driven Hallucination Sampling (BDHS)}. 
BDHS is annotation free and computationally efficient to the point that rejected responses can be generated online, which we explore in Section~\ref{section:ablations-bdhs}.
An overview of the method can be found in Figure~\ref{fig:bdhs_overview}, with further details provided in the following subsections.

\subsubsection{Eliciting LLM Bias via Attention Masking}

As widely discussed, hallucinations in MLLMs often express the underlying language models' inherent biases, for example towards frequently cooccuring objects or object attributes \citep{Li_2023f,qian2024easy,Zhou_2024}. In other words, the MLLM may choose to draw from its parametric knowledge when instead it should have more strongly considered information from the image in question.

We propose to directly induce this failure mode by simply masking attention to image tokens to induce hallucination.
Let $\tilde{x}=(x_{\text{text}}, \tilde{x}_{\text{img}}, m)$ denote the modified input with (optional noisy) image $\tilde{x}_{\text{img}}$ and attention mask $m$. Suppose the MLLM encodes image $\tilde{x}_\text{img}$ to $k$ embedding vectors, each vector with dimension~$d$. Then, $m$ is defined as a boolean mask of dimension $k$.
We suggest to randomly sample the mask $m = (m_1, m_2,\dotsc, m_k)$ according to a uniform distribution $\mathcal{U}(0,1)$ and threshold $\rho_\text{th} \in [0,1]$ where each element follows 
\begin{equation} \label{eq:attention_mask}
m_i =
\begin{cases}
    1 & \text{if } \rho_i \geq \rho_\text{th} \text{ for } \rho_i \sim \mathcal{U}(0,1)\,;\\
    0 & \text{else}\,.
\end{cases}
\end{equation}

By masking the image embeddings using $m$, the model only pays attention to a subset of the $k$ embedding vectors to generate the response $\tilde{y}^-$. Where the remaining signal is not sufficient, the MLLM can only draw on its parametric knowledge to answer, thus inducing hallucination. By allowing access to some part of the image, we encourage more realistic hallucinations.

In our experiments with LLaVA 1.6, we found that $\rho_\text{th}$ values close to 1 empirically gave the strongest results\footnote{Final results are reported at $\rho_\text{th}=0.99$.}. We argue that this is likely a result of the ``AnyRes'' technique used in LLaVA 1.6, which leads to significant redundancy across image tokens.

\subsubsection{Reference-Guided Generation}

Keeping the generated corrupted response $\tilde{y}^-$ close to the preferred one, $y^+$, supports the optimizer in paying more attention to the image as only the non-overlapping portion is affected by the modified input $\tilde{x}$. Otherwise, responses $\tilde{y}^-$ and $y^+$ could diverge early on or $\tilde{y}^-$ could even represent a generic response hinting on missing image information. Instead, in order to maintain consistency in style and structure we propose the following reference-guided sampling strategy, where we ``diverge'' and ``rejoin'' from $y^+$ at random points to form $\tilde{y}^-$.

We assume that the preferred response $y^+$ can be split into $k=1,2,\dots,S$ sentences with $y^+_k$ denoting the $k$-th sentence of $y^+$.
Each sentence is decomposed into two parts $y^+_{k,1}$ and $y^+_{k,2}$, respectively, at a randomly sampled position, i.e., $y^+_k = (y^+_{k,1}, y^+_{k,2})$. The model $\pi_{\theta}$ is then invoked to generate a corresponding corrupted sentence $\tilde{y}^-_k = (y^+_{k,1}, \tilde{y}^-_{k,2})$ whereas
\begin{equation}
   \tilde{y}^-_{k,2} \sim \pi_{\theta}(\cdot | \tilde{x}, y^+_{<k}, y^+_{k,1})\,.
   \label{eq:bdhs_generation}
\end{equation}
Note that this is an abuse of notation for better readability, as $\tilde{y}^-_{k,2}$ denotes the full response sampled from multiple model invocations until the first full stop or end of sequence token.
Every sentence is based on the full ground truth from the previous sentence $y^+_{<k}$ and not the previously generated output $\tilde{y}^-_{<k}$ to improve consistency. Finally, the full BDHS response is given by concatenation of the individual sentences, i.e. $\tilde{y}^- = (\tilde{y}^-_1, \tilde{y}^-_2, \dotsc, \tilde{y}^-_S)$.
Note, the partitioning into sentences is a design decision to keep the non-overlapping portion between $y^+$ and $\tilde{y}^-$ reasonably small and to improve consistency when switching forth and back between responses.
In the implementation, the generation of responses for several sentences and preference pairs can be highly parallelized as~\eqref{eq:bdhs_generation} does not depend on any previously generated output for all $k$.

For question answering tasks, several ground truth responses $y^+$ consist of only one or few words and often start with \textit{yes} or \textit{no}. In these cases $\tilde{y}^-_{k,2}$ can often easily inferred from $y^+_{k,1}$ even without image access at all and therefore we extend the previous strategy by a simple heuristic: whenever $y^+_{k,1}$ starts with a \textit{yes} or \textit{no} it is substituted by its counterpart with a probability of 50\%.

\subsubsection{Ensuring Semantically Meaningful Differences} \label{sec:bdhs:similarity}

Similar to our observation in Online-DAP, BDHS responses $\tilde{y}^-$ can still be very similar to the ground truth $y^+$, especially when the pivot position in $y^+_k$ is late in the sentence, i.e. the length of $y^+_{k,2}$ is small. To maximize learning utility of BDHS preference pairs, this is undesirable.

While further increasing $\rho_\text{th}$ or biasing towards early pivot positions in the reference-guided generation could minimize such trivial generations, this introduces additional hyperparameters and can lead to less realistic dispreferred responses.
Instead, we realize BDHS in an iterative fashion.

Once $\tilde{y}^-$ is generated, a semantic similarity score w.r.t.~$y^+$ is computed using an off-the-shelf sentence embedding model\footnote{We use the \textit{all-mpnet-base-v2} sentence embedding model \citep{Reimers_2019} with $\epsilon_\text{s}=0.97$.}. A new response $\tilde{y}^-$ is sampled if the cosine similarity is above a pre-defined threshold $\epsilon_\text{s}$. After reaching the maximum number of iterations $N_\text{BDHS}$, $\tilde{y}^-$ is generated according to input $\tilde{x}$ without any reference guidance.
Appendix~\ref{section:appendix:bdhs} provides the actual algorithm for BDHS including similarity scoring.

This additional semantic comparison avoids $\tilde{y}^-$ responses that are trivial rephrasings. Moreover, measuring the number of examples that need re-generation allows intuitive tuning of the $\rho_\text{th}$ hyper parameter.

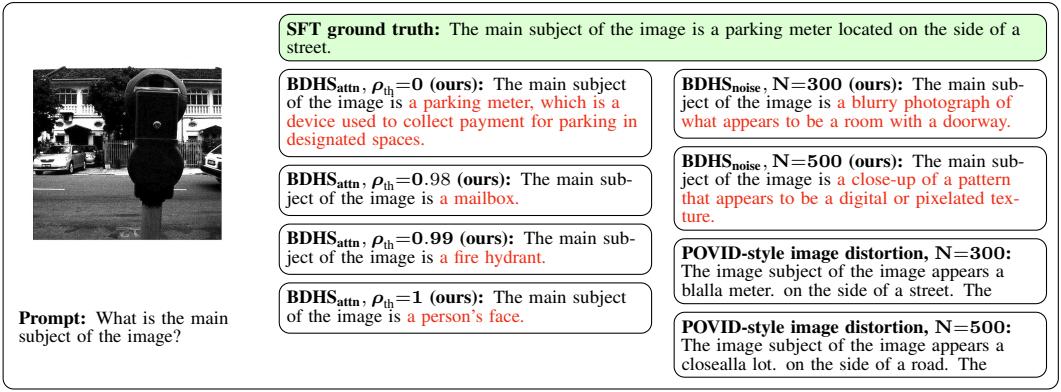
\begin{figure}
	\centering
	\begin{tikzpicture}		
		\placeexample[
		yshiftbox=0cm,
		imgname=COCO_train2014_000000490055,
		prompttext=What is the main subject of the image?,
		sfttext=The main subject of the image is a parking meter located on the side of a street.,
		bdhsatext={The main subject of the image is \corrupted{a parking meter, which is a device used to collect payment for parking in designated spaces.}},
		bdhsbtext={The main subject of the image is \corrupted{a mailbox.}},
		bdhsctext={The main subject of the image is \corrupted{a fire hydrant.}},
		bdhsdtext={The main subject of the image is \corrupted{a person's face.}} ,
		bdhsnatext={The main subject of the image is \corrupted{a blurry photograph of what appears to be a room with a doorway.}} ,
		bdhsnbtext={The main subject of the image is \corrupted{a close-up of a pattern that appears to be a digital or pixelated texture.}},
		povidatext={The image subject of the image appears a blalla meter. on the side of a street. The},
		povidbtext={The image subject of the image appears a closealla lot. on the side of a road. The}
		]
	\end{tikzpicture}

	\caption{Example of generated responses for different hyperparameters and approaches. The image, prompt and SFT ground truth are taken from LLaVA Instruct. For guided generation, actual model completions are shown in bold face.}
	\label{fig:llm_bias_example}
\end{figure}

\subsubsection{Example}
\label{sec:bdhs_example}

Figure \ref{fig:llm_bias_example} presents generated responses for a selected example defined by image, prompt and SFT ground truth from the LLaVA Instruct dataset.
This example should particularly demonstrate the difference between attention masking and noisy images. BDHS with attention masking ($N_\text{BDHS}=5$) is referred to as as $\text{BDHS}_\text{attn}$ 
and BDHS with noisy images in the input as $\text{BDHS}_\text{noise}$.
For $\rho_\text{th}=0$ attention masking is disabled but still guided along the ground truth response. The model is able to properly identify the parking meter in the image. 
With increased attention masking the model starts to hallucinate as desired. Even with fully masked image embeddings the model still hallucinates, while
for $\text{BDHS}_\text{noise}$ the generated responses tend to refer to the blurriness of the images.
The example includes responses for the teacher-forced POVID-style image distortion as described in Section~\ref{sec:povid}.
Due to the token-based, teacher-forced predictions, the generated responses often are non-sensical and inconsistent which worsens for higher noise levels. 

\textbf{Discussion}~~Concurrent to us \citep{Yu_2024} also emphasizes the significance of generating model samples with minimal differences. While their insights on annotation strategy are interesting, their proposed "Deconfounded Candidate Response Generation" approach appears similar to common sampling techniques using higher temperatures in online pipelines, which do not necessarily create pairs of minimal differences. In another concurrent work, \cite{deng2024enhancing} proposes generating "rejected responses" through image corruption. Despite the conceptual resemblance, we find that both using an attention mask and SFT-guided corruption are crucial in our final BDHS design (see Section~\ref{section:ablations-bdhs}). 

%% file: tables/datasets_related_works.tex
\begin{table}[t!]
    \centering
    \resizebox{\textwidth}{!}{
    \begin{tabular}{p{0.5cm} ll | lr | lr  r | r}
        \toprule
        Type                                                                   & Name                       & Size                 & \multicolumn{2}{c|}{Prompt}                                 & \multicolumn{2}{c}{Response}                                                                                                                   & Judge           & \multirow{2}{*}{\centering \parbox{1.4cm}{\centering Preference Signal}} \\

                                                                               &                            &                      & Text                                                       & Image                                              & Chosen                      & Rejected                                                    &                 & \\
        \midrule
        \multirow{3}{*}{\rotatebox[origin=c]{90}{ ~~~~ Human}}                 & LLaVA-RLHF                 & 10k                  & LLaVA-Instruct-150k                                        & COCO                                               & LLaVA 1.5                   & LLaVA 1.5                                                   & human           & ranking\\
                                                                               & RLHF-V                     & 5.7k\textdagger      & UniMM-Chat                                                 & Various\textdagger                                 & Muffin/Various\textdagger (corrected)          & Muffin/Various \textdagger                                                      & human           & construction\\
                                                                               &                            &                      &                                                            &                                                    &                             &                                                             &                 &\\
        \midrule
        \parbox[t]{2mm}{\multirow{2}{*}{\rotatebox[origin=c]{90}{ Synthetic}}} & VLFeedback                 & 80k                  & \multicolumn{2}{l|}{9 datasets (LLaVA, SVIT, etc.)}                                                             & \multicolumn{2}{l}{12 MLLMs (LLaVA 1.5, GPT-4V, etc.)}                      & GPT-4V          & ranking\\
                                                                               & NLF                        & 63k                  & LLaVA-Instruct-150k                                        & COCO                                               & DRESS$_\text{ft}$ (refined) &  DRESS$_\text{ft}$                                          & GPT-4V          & construction\\
                                                                               & POVID                      & 17k                  & LLaVA-Instruct-150k                                        & COCO                                               & \multicolumn{2}{r}{SFT Ground truth ~~~~~SFT Ground truth (corrupted)}     & GPT-4V                  & construction\\
                                                                               &                            &                      &                                                            &                                                    &                             &                                                             &                 & \\
        \bottomrule
    \end{tabular}}
    \caption{Recently published multimodal preference datasets. \textdagger~denotes the updated dataset version of RLHF-V published on Hugging Face Hub.}
    \label{table:sota_datasets}
\end{table}

%% file: figures/bdhs_overview_oranges.tex
\begin{tikzpicture}[font=\fontsize{7pt}{7pt}\selectfont]

\pgfmathsetseed{43}
\definecolor{sfeedback}{gray}{0.5}
\tikzstyle{attention_cell} = [rectangle, draw,  minimum size=0.2cm, inner sep=0cm];
\tikzstyle{masked_attention_cell} = [attention_cell, fill];
\tikzstyle{block} = [rectangle, draw, minimum width=3cm, minimum height=1cm, text width=3cm, rounded corners];
\tikzstyle{sentence_block} = [block, fill=white, text width=3.6cm, text depth=0.6cm, font=\fontsize{6pt}{6pt}\selectfont];
\tikzstyle{img_block} = [rectangle, minimum width=3cm, minimum height=1cm];
\tikzstyle{text_block} = [rectangle, minimum width=3cm, minimum height=1cm];

\node[block, label=below:Prompt $x_\text{text}$] (prompt) at (0,0) {Describe this image in detail.};

\node[img_block, label=below:Image $x_\text{img}$, below=0.5cm of prompt] (img) { \includegraphics[width=3cm] {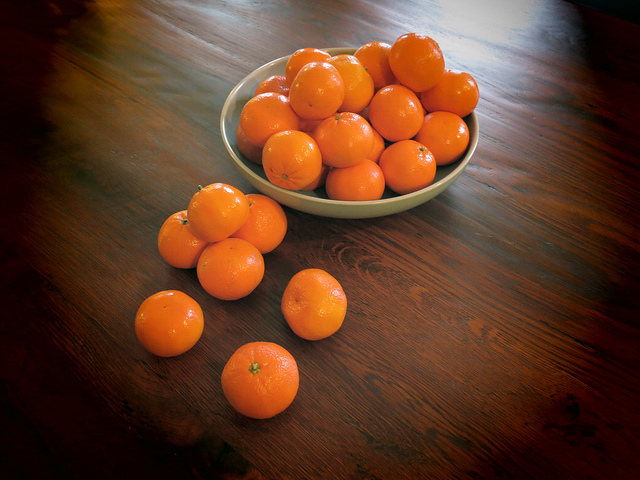} };

\node[text_block, below=0.8cm of img] (attn_mask) {Sampled image attention mask $m$};
\foreach \x in {-1.5,-1.3,...,1.5}{
	\pgfmathsetmacro{\shouldmask}{random(0,10)};%
	\pgfmathsetmacro\switch{\shouldmask > 3 ? "masked_attention_cell" : "attention_cell"};
	\coordinate (pos) at (\x,0);
	\node[\switch] at (pos |- attn_mask.north) {  };
}

\node[densely dashed, block, minimum width=4cm, minimum height=6.5cm, yshift=0.2cm] (input) at ($(prompt.north)!0.5!(attn_mask.south)$) {};
\node[anchor=north west] at (input.south west) {Input $\tilde{x}$};

\node[block, minimum width=4.8cm, minimum height=4.6cm, right=0.5cm of input.north east, anchor=north west] (generation)  {};
\node[anchor=north west] at (generation.south west) {Guided Generation};
\node[text width=4.2cm, align=center, anchor=north] (gen_desc) at (generation.north) {\textit{Split each sentence at a random position. Complete sentences via model inference.}};
\node[sentence_block, anchor=north west, xshift=-0.05cm, yshift=-0.0cm] (s1) at (gen_desc.south west) {$ \pi_\theta(\cdot \vert \tilde{x}, $ "", "The image features a wooden dining table on which "$) \rightarrow$ \corrupted{a pile of oranges has been neatly arranged.}};
\node[sentence_block, xshift=0.3cm, yshift=-1.05cm, text depth=1.1cm] (s2) at (s1) {$ \pi_\theta(\cdot \vert \tilde{x}, $ "The image features a wooden dining table on which a bowl filled with oranges is placed.", "The bowl appears to be in the center of the table"$) \rightarrow$ \corrupted{ and the oranges are arranged in a circular pattern around it.}};
\node[sentence_block, xshift=0.3cm, yshift=-1.4cm, minimum height = 1.4cm,  text depth=1.1cm] (s3) at (s2) {$ \pi_\theta(\cdot \vert \tilde{x}, $ "The image features a wooden dining table on which a bowl filled with oranges is placed. The bowl appears to be in the center of the table.", "In addition,"$) \rightarrow$ \corrupted{there is a single banana on the table.}};

\node[block, fill=corruptedbackground, text width=4cm, right=0.5cm of generation] (output) {
The image features a wooden dining table on which \corrupted{a pile of oranges has been neatly arranged.} 
The bowl appears to be in the center of the table \corrupted{and the oranges are arranged in a circular pattern around it.} 
In addition, \corrupted{there is a single banana on the table.}};
\node[anchor=north west] at (output.south west) {BDHS response $\tilde{y}^-$};

\node[block, minimum width=4.8cm, text width=4.4cm, anchor=south, fill=gtbackground] (gt) at (input.south -| generation)  {The image features a wooden dining table on which a bowl filled with oranges is placed. The bowl appears to be in the center of the table. In addition, more oranges are scattered around the table.};
\node[anchor=north west] at (gt.south west) {SFT response $y^+$};

\node[block, align=center, color=sfeedback] (sc) at (gt -| output)  {Similarity Scoring via Sequence Embeddings};

\draw[-latex] (input.east |- generation) -- (generation);
\draw[-latex] (generation) -- (output);
\draw[-latex] (gt) -- (generation);
\draw[-latex, dashed, sfeedback] (output) -- (sc);
\draw[-latex, dashed, sfeedback] (gt) -- (sc);

\coordinate[below=of gt] (feedback);
\draw[-latex, dashed, sfeedback] (sc) |- (feedback)  node [fill=white, pos=0.15, below] {If similarity $> \epsilon_s$} -| (input) coordinate[name=feedbackmid, pos=0];

\foreach \x in {-1.5,-1.3,...,1.5}{
	\pgfmathsetmacro{\shouldmask}{random(0,10)};%
	\pgfmathsetmacro\switch{\shouldmask > 3 ? "masked_attention_cell" : "attention_cell"};
	\coordinate (pos) at ($ (feedbackmid) + (\x,0) $);
	\node[\switch, sfeedback, below=0.1cm] at (pos) {  };
}
\node [above=-0.05cm, sfeedback] at (feedbackmid) {Sample new mask};

\end{tikzpicture}

%% file: sections/4_experiments.tex
\vspace{-2mm}
\section{Experimental Setup}
\vspace{-1mm}
\label{section:experimental-setup}

\subsection{Model}
We conduct our ablations on LLaVA 1.6 as this series of models is both well studied and exhibits strong performance across a range of multimodal tasks~\citep{Liu_2024b}. Particularly, we focus on aligning the LLaVA 1.6-7B Vicuna model variant as this scale of parameters is particularly widely used in the community. Notably, LLaVA 1.6-7B provides a significantly stronger baseline performance over the more common choice of LLaVA 1.5-7B in the multimodal alignment literature.

\subsection{Evaluation}
\label{section:eval_sets}

\textbf{Benchmarks}~~We adopt multiple benchmarks to assess the capabilities of MLLMs, centered around both measuring the models visual faithfulness, i.e.~its tendency to hallucinate, as well as overall helpfulness, i.e.~the overall quality of its responses. Results have been obtained using an internal fork of lm-eval-harness \citep{eval-harness,Mckinzie_2024,Li_2024}.

\texttt{LLaVABench-in-the-Wild}~\citep{Liu_2023b}, \texttt{TextVQA}~\citep{Singh_2019}, and \texttt{GQA}~\citep{Hudson_2019} help measure the model helpfulness, i.e. the effectiveness at following instructions and the completeness of the responses. \texttt{LLaVABench-in-the-Wild} expects free-form answers while both \texttt{TextVQA} and \texttt{GQA} require concise responses. We additionally report \texttt{MMVet}~\citep{Yu_2023c}, which evaluates the knowledge and visual reasoning capabilities of the MLLM. Such capabilities are not a direct target for most MLLM alignment strategies to improve. Nevertheless, \texttt{MMVet} offers a useful indicator for ensuring that such capabilities are not lost due to a possibly too simple or not sufficiently diverse alignment regiment.

\texttt{POPE}~\citep{Li_2023f} and \texttt{MMHALBench}~\citep{Sun_2023} evaluate the visual faithfulness of a model by identifying hallucinations in model responses. For \texttt{POPE}, we noticed that most of our experiments would reach a seeming plateau between 86\% and 88\% despite improvements in the other benchmarks. We conducted an initial manual review of 100 reported losses and observed incorrect or disputable ground truth on as many as 20 \% of those samples (see Appendix \ref{section:appendix-pope}). While re-annotating those examples is beyond the scope of this work, we invite the community to consider it as many recent SOTA models exhibit such plateau\footnote{See Table 4 in \citet{Mckinzie_2024} where all the models reported are demonstrating a plateau on \texttt{POPE}.}.

Additionally, we noticed unexpected results on \texttt{MMHALBench}, and subsequent analysis showed limitations in its scoring. Specifically, \texttt{MMHALBench} uses text-only GPT-4 to detect hallucinations by comparing model responses to a reference response and a short list of objects known to be in the image. Sometimes this leads to entirely correct model responses to be marked as hallucinations when they included more detail than the provided references. To mitigate this issue, we introduce a straightforward derivative we call \texttt{MMHALBench-V(ision)}, which relies on GPT-4o, i.e. provides the input image as additional context to the judge, to more reliably evaluate model capabilities. Data and evaluation prompts are unchanged. We empirically found this to be more reflective of true hallucinations in a human comparison. See Appendix~\ref{section:appendix-mmhalbenchv_motivation} for our review. Throughout experiments, we mainly focus on \texttt{MMHALBench-V} numbers and report \texttt{MMHALBench} primarily for reference.

While responses that have fewer hallucinations are often also inherently more helpful, we observe that these dimensions are nevertheless distinct and optimizing for reduction in hallucination crucially does not necessarily imply a more helpful model. In fact, in some instances, we even observed an inverse relationship. For example, as discussed in~\citep{Zhu_2023c}, a given model would be more likely to hallucinate when asked to produce longer captions than shorter ones. This implies that models could learn to hallucinate less simply by providing more concise, arguably less useful, responses, and that models that aim to provide more detailed responses may find it more difficult to remain faithful to visual context in all respects\footnote{To some extent, one could argue this mirrors the tension between helpfulness and safety as reported in \citet{Touvron_2023}, where a highly safe model may be less helpful.}. For this reason, we report the recall metric from \texttt{Object HalBench} \cite{Yu_2023}, styled Recall$^{\text{coco}}$ in our tables. This measures how many objects known to be in an image based on CoCo annotations are mentioned in a comprehensive caption given by the model. We considered as well reporting the \texttt{CHAIR}~\citep{Rohrbach_2018} metrics from \texttt{Object HalBench}~\citep{Yu_2023}. However, during our experiments, we found that those measurements were not always correlated with the quality of the models evaluated (see Appendix \ref{section:appendix-chair}).

\vspace{-2mm}
\section{Experiments}
\vspace{-1mm}
\label{section:experiments}

In this section, we empirically evaluate different aspects of aligning MLLMs. We start by summarizing our key findings in Section~\ref{section:combined_learnings}. Then, we proceed with an in-depth ablation study on the components we have discussed in the paper, offering a clearer view of effect. We begin with equalizing the experimental conditions on public preference datasets (Section~\ref{section:dataset_baselines}). We then highlight desiderata for a high-quality preference dataset (Section~\ref{section:dataset_ablations}) and show that BDHS can be a simple and effective mechanism following such best practices (Section~\ref{section:ablations-bdhs}). Subsequently, we compare various alignment techniques, such as RL-based methods (Section~\ref{section:rl-experiments}), Online and Mixed-DPO strategies (Section~\ref{section:onlinedpo-mixeddpo}), as well as various offline approaches (Section~\ref{section:offline}).

\subsection{Key Components in MLLM Alignment Pipeline}
\label{section:combined_learnings}

\input{tables/big_picture}

We summarize our main findings and compare results with other SOTA models in Table~\ref{table:main_table}.
First, we fixed the base model (LLaVA 1.6-7B) and studied the effects of online vs. offline methods using the POVID alignment data~\citep{Zhou_2024}. While offline DPO shows more significant improvement on benchmarks that consider hallucination, such as \texttt{POPE} and \texttt{MMHALBench-V}, the Online-DPO enhances benchmarks evaluating the quality of answers in an open question answering setup, like \texttt{LLaVABench-in-the-Wild}. This is intuitive, as the preference pairs in the POVID dataset are specifically designed to reduce hallucinations. In contrast, the online samples from the model may not always provide as strong a signal for reducing hallucinations. Mixed-DPO allows to incorporate the benefits of both approaches and the results show consistent improvement over both online and offline methods.

When using Online-DPO or Mixed-DPO strategies, we typically depend on advanced models like LLaVA 1.6-34B to rank the online samples generated by the model. However, access to such models is not always guaranteed. We discuss this limitation in more detail in Section~\ref{section:anno-ablations}. Additionally, the construction of the POVID dataset also involves using a superior model such as GPT-4V to inject noise into SFT data. Our proposed BDHS method does not require additional annotators or preference data, and relies exclusively on SFT data already available from the instruction tuning of the base model. Despite this simplicity, it consistently outperforms the models that utilize the larger POVID dataset (i.e. both offline and Mixed-DPO) in most benchmarks. Implementing BDHS in an online format further closes this performance gap in \texttt{MMHALBench-V}, establishing BDHS as a compelling and cost-effective alternative to other more resource-intensive approaches. Combining the POVID dataset with the online-BDHS approach (referred to as Online-BDHS $\cup$ POVID), with the exception of \texttt{MMHALBench-V}, consistently outperforms the model that uses only the POVID dataset across all benchmarks. It also surpasses STIC \citep{deng2024enhancing} and RLAIF-V \citep{Yu_2024} on the reported benchmarks. We further discuss the enhanced efficacy of our approach over \citet{Zhou_2024} in Section~\ref{section:ablations-bdhs}.

While Section~\ref{section:dataset_ablations} provides a detailed analysis of various preference datasets, we highlight key findings from the VLFeedback dataset here, as they contribute significantly to building an effective alignment strategy. Unlike POVID, both VLFeedback and its variant, VLFeedbackCorrupted(5k), select the ``chosen response'' in the preference pairs from the top responses ranked by GPT-4V, selected from a pool of model-generated responses. Compared to re-using SFT data, this approach potentially offers an additional supervisory signal to the model, leading to enhanced performance on benchmarks like \texttt{LLaVABench-in-the-Wild}, where such aligned models even outperform the unaligned 13B and 34B models from the same family.

Notably, we introduce VLFeedbackCorrupted (5k), a small dataset leveraging corruption injection to generate the ``rejected response'', which performs competitively to the much larger rank-based VLFeedback (full) dataset. These experiments demonstrate the effectiveness of two strategies in constructing preference data: First, learning from strong (highly-ranked) responses seems to yield a distillation-like benefit. Second, using subtle differences between ``chosen'' and ``rejected'' responses, as opposed to just rank-based pairs (like in VLFeedback (full)), can significantly reduce hallucinations, even in a limited data regiment.

Finally, we replace the noise injection strategy using GPT-4 with our proposed BDHS. We observe a slight reduction of the \texttt{MMHALBench-V} and \texttt{LLaVABench-in-the-Wild} scores compared to the GPT-4V based approach, but note that the achieved result still represents meaningful improvements over the baseline. On all other metrics, BDHS shows comparable or even superior results, establishing BDHS as a strong alternative to GPT-4V in this pipeline.

In the remainder of this section, we conduct a comprehensive ablation study on each of the components discussed earlier, aiming to offer insights into the typical trade-offs encountered in alignment strategies.

\subsection{Removing Confounding Factors for Previously Published Datasets}
\label{section:dataset_baselines}

We analyze RLHF-V~\citep{Yu_2023}, VLFeedback~\citep{Li_2023d} and POVID~\citep{Zhou_2024} as they offer a fair blend between human and synthetic sources, and between constructed and ranked preference signal composition. As it is challenging to determine what are the properties that characterize a high-quality preference dataset, we first replicate alignment using the published datasets against LLaVA 1.6-7B with DPO. Additionally, we sub-sample all datasets to a consistent size of 5,000 examples to mitigate effect sizes. Results are summarized in Table~\ref{table:datasets_baselines}. When available, we additionally report the results published by the original authors, highlighted in gray in the table.

\cite{Zhou_2024} have conducted a similar experiment using LLaVA 1.5, however they do not control for dataset size. We were successful in replicating certain observations published by these authors. POVID reaches the highest score on \texttt{POPE}. \cite{Zhou_2024} also reports the highest \texttt{MMHALBench} scores with POVID, which we were able to reproduce using LLaVA 1.6, although this is only true when size correction is not applied. Upon normalizing for size, POVID's performance equaled that of VLFeedback and was lower than RLHF-V.

\input{tables/datasets_baselines.tex}

In other domains, our experiment have shown divergent trends. While~\cite{Zhou_2024} demonstrated that all preference datasets improved LLaVA 1.5 on \texttt{MMVet}, our findings with LLaVA 1.6 exhibited a reverse trend: all our runs did not match up to the baseline. Interestingly, as the datasets grew larger, we witnessed a further deviation from the baseline. We hypothesize that these preference datasets lack the necessary information to improve \texttt{MMVet} over the notably stronger baseline LLaVA 1.6 introduced, which necessitates specialized knowledge (see Section~\ref{section:eval_sets}). VLFeedback, to a certain extent, may possess some of this knowledge thanks to its diverse prompts. However, the other datasets appear to fall short. By restricting dataset sizes, we further limit the potential alterations on the non-aligned model, as the results stay closer to that baseline.

Oppositely, VLFeedback on \texttt{LLaVABench-in-the-Wild} shows an uplift bump that is only limited when the size restriction limit is applied. When aligning on the complete VLFeedback, the largest dataset in these experiments, we can achieve the highest score on that benchmark.

\subsection{Desiderata for Preference Datasets}
\label{section:dataset_ablations}

We examine the components of a preference dataset for multimodal alignment, as introduced in Section~\ref{section:preference_datasets}, and investigate the following options in constructing this preference data. The explored choices are further summarized in Table~\ref{table:datasets_settings}.

\begin{itemize}[nosep,leftmargin=0.5cm]
    \item \textbf{Prompts}~~We compared (i) a diverse prompt strategy mixing multiple datasets to (ii) prompts only from LLaVA-Instruct-150k, which was already seen during the SFT stage of the base model.
    \item \textbf{Chosen responses}~~We introduced 3 settings: (i) diverse responses from multiple MLLMs; (ii) LLaVA responses only, (iii) GPT-4V responses only.
    \item \textbf{Rejected responses}~~We introduced 2 settings: (i) diverse responses from multiple MLLMs, and (ii) corruption of the chosen responses.
\end{itemize}

In order to construct these preference dataset ablations cheaply and reproducibly, we leverage the size and diversity of the VLFeedback dataset~\citep{Li_2023d}. VLFeedback possesses several properties that makes it a good sandbox: (a) the prompts, derived from 9 datasets (LLaVA-Instruct-150k, SVIT, LLaVAR, etc.), are diverse, (b) the chosen and rejected responses are sampled from 12 MLLMs making them very diverse too -- $\sim37$\% responses are from GPT-4V, and $\sim$35\% from the LLaVA 1.5 series, (c) finally, the large size of VLFeedback, 80,000 quadruplets of responses that can be paired together, makes it simpler to isolate specific aspects.

\input{tables/datasets_settings.tex}

\textbf{Corruption strategy}~~Reranking is originally used to determine chosen and rejected responses in VLFeedback (see Section~\ref{section:preference_datasets}).
In order to remove variation introduced by the original rejected responses (e.g., style change between MLLMs) and permit a tighter control on ablations, we replace rejected responses from the original VLFeedback samples with corrupted versions of the preferred responses. Similar to the method in \citep{Zhou_2024}, we leverage GPT-4 to specifically introduce realistic hallucinations, assisted by a few shots for illustration (see Appendix~\ref{section:appendix-dataset-prompt-corruption}).

\input{tables/ablation_datasets}

\textbf{Results}~~Following Section~\ref{section:dataset_baselines}, we apply DPO alignment on the LLaVA 1.6-7B model, and we limit all the datasets to 5,000 samples. In Table~\ref{table:ablation_datasets}, we report the results of this experiment.
First, we show that our corruption strategy achieves improvements over the baseline comparable in magnitude to the ranking-based preference signal in the original VLFeedback data. In some benchmarks, like \texttt{MMHAL-Bench-V}, we even observe improvements, while notably \texttt{MMVet} shows some regressions. Nevertheless, we argue that this represents a reasonable baseline to adopt for easier iteration on the following ablations. In Appendix~\ref{section:appendix-dataset-size-ablation}, we provide more analysis on this strategy.

Next, we explore the impact of novelty of the prompts used for alignment, by sampling another 5k preference data generated with the same corruption mechanism solely from prompts that are a part of the LLaVA SFT mixture. These are examples that the base model would have already been trained on during the SFT stage. Interestingly, it appears that using novel prompts does not offer substantial benefits. We still observe comparable lift on \texttt{LLaVABench-in-the-Wild}, and while \texttt{MMHAL-Bench-V} shows less dramatic improvement over the baseline compared to the more diverse corruption-based sample, this may be due to more verbose responses, as indicated by higher recall. \texttt{POPE} even improves somewhat significantly and the regression in \texttt{MMVet} is also less pronounced.

Finally, we explore the impact of the construction of the accepted response in the alignment data. One could argue that for responses derived from stronger model such as GPT-4V, improvements may also be the result of learning from this stronger teacher model. Therefore, we conduct two experiments: one, where we sample data where the preferred response comes from GPT-4V only, and one where the preferred response comes from LLaVA 1.5-7B, a model generally weaker than the base model under alignment in this experiment. Interestingly, we do not observe any benefit from learning from GPT-4V generated responses, in fact, our results suggest that positive samples derived from LLaVA 1.5-7B led to a slightly stronger model post alignment.

These findings suggests that useful preference data can be derived cheaply, even from responses from relatively weaker models, as long as one can effectively sample and identify relatively desirable answers from the model as their preferred response\footnote{In this ablation, preferred responses were selected from model responses that were ranked as the best among the sampled model responses per example in VLFeedback. In this setting the chosen response can be assumed to be of reasonable quality as it was chosen to be at least better than other models'. While this still indirectly exploits the ranking in the VLFeedback data, all that is required is a way to sample reasonable model answers, which is generally much more readily available in practical scenarios than paired preference data, for example via cheap user feedback (thumbs up / down).}, and introduce targeted corruption to create dispreferred responses.

In the following Section~\ref{section:ablations-bdhs}, we will discuss how one can avoid both the need for sampling preferred model responses as well as the need for an external model to introduce corruption with BDHS.

\subsection{Ablations on BDHS}
\label{section:ablations-bdhs}

Section~\ref{section:BDHS} introduces BDHS as a technique to generate corrupted responses directly using the model subject to alignment.
While our proposed approach is purely based on image attention masking, we also evaluate a variant that consumes noisy images instead, motivated by the teacher-forced \textit{POVID-style image distortion} introduced in~\citet{Zhou_2024} (see Section~\ref{sec:povid}).
In the following, BDHS with attention masking ($\rho_\text{th}=0.99$ and $N_\text{BDHS}=5$) is denoted as $\text{BDHS}_\text{attn}$
and BDHS with noisy images in the input as $\text{BDHS}_\text{noise}$. The number of additive noise steps for $\text{BDHS}_\text{noise}$ is set to $N=500$ similar to the image distortion in \cite{Zhou_2024}.

All ablations in Table~\ref{tab:bdhs_ablation} are based on our 5k subset of POVID as introduced in Section~\ref{section:dataset_baselines}.
As described in Section~\ref{sec:recent_works}, POVID contains LLaVA Instruct responses $y^+$ as well as GPT-4V corrupted non-preferred responses $y^-$.
While $y^+$ is shared between all ablations, we start with substituting~$y^-$ from external supervision by the BDHS model response $\tilde{y}^-$ and invoke standard DPO as shown in the first 3 rows after the LLaVA 1.6-7B baseline results.
The proposed variants consistently improve over the baseline for \texttt{POPE} and \texttt{LLaVABench-in-the-Wild}.
They regress on \texttt{MMHALBench}, however, as discussed in Section~\ref{section:eval_sets}, this benchmark has limitations so we mainly focus on \texttt{MMHALBench-V} instead for which all $\text{BDHS}_\text{attn}$ variants perform comparable to the baseline while the online rollout of $\tilde{y}^-$ even improves over it.
Notably, we also observe significantly higher \texttt{Recall$^{\text{coco}}$}, suggesting richer responses.
$\text{BDHS}_\text{noise}$ results in lower scores for \texttt{LLaVABench-in-the-Wild} while the attention masking approach $\text{BDHS}_\text{attn}$ almost maintains the baseline scores.

\input{tables/llm_bias}

The lower partition of Table~\ref{tab:bdhs_ablation} starts with plain DPO on the POVID (5k) dataset as reference and then each subsequent approach incorporates both the existing response $y^-$ from external supervision as well as $\tilde{y}^-$ derived from the policy.
Hereby, the two non-preferred responses are incorporated into the DAP framework by averaging the losses of ($y^+, y^-$) and ($y^+, \tilde{y}^-$) according to Equation~\eqref{eq:avg_dpo}.
Therefore, the Online-BDHS method uses Online-DPO in a considerable simplified setting compared to the full Online-DPO realization~\eqref{eq:mixed_dpo}, as the formulation presented here does not depend on a dedicated external annotator (see Section~\ref{sec:online_dap}).

All the BDHS ablations improve significantly on \texttt{LLaVABench-in-the-Wild} compared to the DPO baseline and POVID-style image distortion.
The $\text{BDHS}_\text{attn}$ with attention masking performs significantly better on \texttt{MMVet} compared to $\text{BDHS}_\text{noise}$.
Notably, $\text{BDHS}_\text{attn}$ consistently outperforms the POVID-style image distortion across all benchmarks.
We follow the published implementation of \cite{Zhou_2024}, however, surprisingly the \textit{POVID-style image distortion} performs worse compared to plain DPO via POVID (5k), which differs from the LLaVA 1.5-7B alignment results in their paper.
Presumably, the non-sensical responses from teacher-forcing could lower the performance while trading off with the existing GPT4-V preference pairs.

While online approaches with BDHS improve on certain benchmarks, we emphasize that even the offline dataset created with $\text{BDHS}_\text{attn}$ and without additional response from external supervision already constitutes a cost-effective baseline that consistently performs well across all benchmarks.
Unless otherwise stated, BDHS in the following sections generally refers to $\text{BDHS}_\text{attn}$.

\subsection{RL-based Alignment}
\label{section:rl-experiments}

To evaluate RL-based alignment methods we followed the established recipe of training a reward model on a preference dataset and then using an RL algorithm to optimize the MLLM to maximize the reward of responses sampled from the policy. We chose PPO and RLOO due to their popularity in the LLM literature.

\textbf{Reward Model Training and Evaluation}~~We analyze the utility of datasets available in the community for reward model training by training on POVID, RLHF-V and VLFeedback preference datasets. To evaluate such created reward models in isolation, we hold out a small validation set split from the original dataset and report classification accuracy of the trained reward model, i.e. its ability to differentiate the chosen from the rejected response in POVID, RLHF-V, and VLFeedback sets. These held out validation sets are not used for reward model training.

\input{tables/rm_validation_accuracy}

Table~\ref{table:rm_validation_accuracy} shows the performance of the reward models trained on different datasets across all validation sets. The model trained on VLFeedback shows the best generalization across the different datasets, as may be expected given its significantly larger size and higher diversity. In contrast, reward models trained on POVID and RLHF-V show notably poor generalization to their respective counterpart, while achieving high scores on their own held out portions. We hypothesize that the reward model may learn to recognize and prefer the respective (original) policy response before corruption (POVID) or enhancement (RLHF-V), which could explain the performance being significantly below a random choice baseline. To strengthen our hypothesis, we also combine the POVID and RLHF-V sets for reward model training and observe that both LLaVA 1.5-7B and LLaVA 1.6-7B are able to learn a more balanced objective, although even for such a combined training set we still observe limited generalization to VLFeedback.

\textbf{RL Training and Evaluation}~~We used the POVID and VLFeedback based reward models for PPO and RLOO training. Table \ref{table:alignment_rl} shows the scores of the best models trained via PPO and RLOO.

\input{tables/rl_alignment}

Mirroring the observed lack in generalization in our reward model experiments, we found that using POVID-based reward model resulted in collapse of responses during the RL training. Only the use of the reward model trained on the much larger VLFeedback dataset allowed for stable RL training without model collapse. We hypothesize that besides the larger size, VLFeedback may be more aligned with the downstream objective of the reward model due to its construction by ranking sampled model responses, compared to POVID, which aims to produce minimally different preference pairs.  Nevertheless, even the stronger VLFeedback-based reward model did not allow us to reliably outperform a much simpler DPO baseline\footnote{We also found that models achieving higher reward during RL training, did not perform better than models with lower reward and less KL divergence, i.e., models with higher $\beta$ parameter performed better on the benchmarks.  None of the RL algorithms clearly outperformed the others.}.

These observations indicate that reward model training with subsequent RL alignment could perhaps require more carefully curated data, e.g., with more focus on diversity, than direct alignment methods where both POVID and VLFeedback individually achieve strong improvements.
In addition to inherently stronger reward models, perhaps basing them on more powerful base models, it also suggests that the approach introduced in the concurrent work of \citet{Yu_2024}, which introduces a symbolic reward formulation based on scores from a VQA model verifying statements made by the policy may be a promising avenue for future research.

Another interesting observation is that the RL aligned models show similar evaluation trends as the DAP aligned models, where both use POVID prompts and images for the training of the policy. For example, compared to the base model they show some improvement in \texttt{POPE}, and \texttt{MMHalBench} (both variants), with some regressions in \texttt{LLaVABench-in-the-Wild}, \texttt{TextVQA}, \texttt{GQA}, and \texttt{MMVet}. These trends are distinct to what is seen when using direct preference alignment on VLFeedback data as shown in Table~\ref{table:datasets_baselines}. This is remarkable as the RL aligned models do of course not use the chosen and rejected responses present in the POVID dataset, instead getting their feedback signal entirely from the reward model which is trained on VLFeedback data. We observe a similar trend in Section~\ref{section:onlinedpo-mixeddpo}, where in a purely online setting, the choice of input prompts and images significantly impacts alignment results.

\subsection{Online-DPO \& Mixed-DPO}
\label{section:onlinedpo-mixeddpo}

We apply Online-DPO and Mixed-DPO to both the POVID and the RLHF-V dataset. The results are summarized in Table \ref{table:mixed-DPO}. Consistent with our observations on the POVID dataset, applying Mixed-DPO -- which combines elements of DPO and Online-DPO -- typically results in a moderating effect on performance outcomes. The results often span a range slightly broader than the highest and lowest performances achieved by DPO and Online-DPO. This variability is attributed to the probabilistic nature of the online sampling in Online-DPO.

On the RLHF-V dataset, where Online-DPO consistently outperforms DPO across all benchmarks, the moderating effect of Mixed-DPO proves not beneficial, as the offline DPO component contributes minimally to the overall model performance. Nevertheless, Mixed-DPO remains a valuable strategy in scenarios where, as observed in the experiments on the POVID dataset, offline and Online-DPO show complementary improvements, leveraging the strengths of both to optimize overall performance.

\input{tables/mixed_dpo}

\subsubsection{How does a stronger annotator affect the performance of aligned model?} \label{section:anno-ablations}

\textbf{Annotator Evaluation} We used LLaVA 1.6-34B as the annotator. To verify its capability to accurately judge different responses, we evaluated it on the sample held-out part of three datasets we used for evaluating the reward model in Section~\ref{section:rl-experiments}. Results are summarized in Table \ref{table:anno_eval}. For further details on the prompts used and qualitative examples of the annotator's outputs, please refer to Appendix~\ref{section:appendix-annotator}.

Table \ref{table:anno_ablation} presents a comparison of the effects of Online-DPO with two different annotators.
\input{tables/anno_eval}
While using Online-DPO with LLaVA 1.6-7B as the judge can enhance the overall performance of the model, the stronger annotator seems to provide more consistent improvements across various benchmarks.

Concurrent to us, \citet{Yu_2024} proposed an annotation approach that segments the annotation process into easier sub-tasks, with each task being individually scored. These scores are then aggregated to form an overall score that rates the responses. This method can potentially enable weaker models to still provide strong supervision signals during the alignment process. Moreover, exploring the use of stronger base models and diverse datasets, both in terms of size and variety, could further enhance the effectiveness of the online approach. We leave the detailed investigation of these aspects for future work.

\input{tables/annotator_ablation}

\subsection{Comparison of Different Offline Alignment Methods}
\label{section:offline}

While we conducted most of our experiments using DPO for comparability with other works in the community, we also ran a few experiments to investigate whether other popular offline methods could improve the results. Results are summarized in Table~\ref{table:alignment_offline}.

Our results indicate that both IPO and SLiC, similar to DPO, boost the model's performance across most hallucination benchmarks. Additionally, these methods demonstrate improvements in more open question-answering benchmarks. We anticipate that Online-IPO and Online-SLiC will yield enhancements over their offline counterparts — similar to the improvements observed with Online-DPO over DPO — as examined in \citet{Guo_2024}. However, this study is beyond the scope of this paper and is left for future work. Primarily, we aim to highlight the importance of considering different alignment objectives, emphasizing that the choice between offline objectives in different setups can impact the effect of the alignment pipeline.

\input{tables/alignment_offline}

%% file: tables/big_picture.tex
\definecolor{blue-c}{rgb}{0.8, 0.85, 0.95}
\begin{table*}[t!]
\centering
\resizebox{\textwidth}{!}{%
\begin{tabular}{l l c c c c c c c c c}
\toprule
Model         & Alignment  & Dataset                                  & POPE $\uparrow$ & MMHAL      $\uparrow$ & MMHAL$^\text{v}\uparrow$ & LLaVA$^\text{W}\uparrow$ & VQA$^\text{T}\uparrow$      & GQA $\uparrow$        & MMVet $\uparrow$  & Recall$^{\text{coco}}\uparrow$ \\
\midrule
LLaVA 1.6-7B  & --         & --                                       & 86.40           & 2.95                  & 2.75                     & 80.85                    & 64.85                       & 64.23                 & 43.94             & 68.13                          \\
LLaVA 1.6-13B & --         & --                                       & 86.23           & \textbf{\uline{3.23}} & \textbf{\uline{3.18}}                   & 86.10                    & \textbf{\uline{65.7}}       & \textbf{\uline{64.8}} & 48.26             & 68.13                          \\
LLaVA 1.6-34B & --         & --                                       & 87.73           & \textbf{\uline{3.50}} & \textbf{\uline{3.46}}    & 88.35                    & \textbf{\uline{69.5}}       & \textbf{\uline{67.1}} & 53.90             & 71.17                          \\
OmniLMM-12B\textdagger    & --         & --                                       & --              & 3.14                  & --                       & 74.3                     & --                          & --                    & --                & --                             \\
\midrule
LLaVA 1.6-7B\textdagger   & DPO        & STIC                                     & --              & --                    & --                       & 79.2                     & 65.2                        & --                    & \uline{45.0}      & --                             \\
LLaVA 1.5-7B\textdagger   & RLAIF-V    & RLAIF-V                                  & --              & \uline{3.06}                  & --                       & 64.9                     & --                          & --                    & --                & --                             \\
OmniLMM-12B\textdagger    & RLAIF-V    & RLAIF-V                                  & --              & \textbf{\uline{3.36}} & --                       & 74.3                     & --                          & --                    & --                & --                             \\

\midrule
LLaVA 1.6-7B  & DPO        & POVID  (Full)                            & 88.09           & \textbf{3.16}         & \uline{3.07}             & 78.63                    & 64.56                       & 64.12                 & 40.60             & 73.48                          \\
\rowcolor{blue-c}
LLaVA 1.6-7B  & Online-DPO & POVID  (Full)                            & 86.49           & 2.88                  & 2.94                     & 82.61                    & 64.88                       & 64.31         & 43.26             & 68.45                          \\
\rowcolor{blue-c}
LLaVA 1.6-7B  & Mixed-DPO  & POVID  (Full)                            & 88.03           & 2.83                  & \textbf{3.10}            & 82.75                    & 64.93                       & \textbf{64.47}        & 42.80             & \uline{74.53}                          \\

\midrule
LLaVA 1.6-7B  & DPO        & POVID (Full)                             & 88.09           & \textbf{3.16}         & \uline{3.07}             & 78.63                    & 64.56                       & 64.12                 & 40.60             & 73.48                          \\

\rowcolor{blue-c}
LLaVA 1.6-7B  & DPO        & BDHS (POVID, 5k)                         & \uline{88.75}   & 2.61                  & 2.71                     & 86.33                    & 65.07                       & 63.97                 & 43.4              & \textbf{75.58}                            \\
\rowcolor{blue-c}
LLaVA 1.6-7B  & DPO        & Online-BDHS (POVID, 5k)                  & \textbf{88.83}  & 2.80                  & 2.99                     & 85.03                    & 65.09                       & 63.65                 & 43.12             & 74.09                             \\
\rowcolor{blue-c}
LLaVA 1.6-7B  & DPO        & \textasteriskcentered\ $\cup$ POVID (5k) & 88.38           & 2.82                  & 2.81                     & 84.01                    & \textbf{65.42}              & 64.30                 & \textbf{45.46}    & 74.00                             \\
\midrule
LLaVA 1.6-7B  & DPO        & VLFeedback (Full)                        & 81.84           & 2.96                  & 2.99                     & \textbf{90.75}           & 62.93                       & 62.53                 & 43.85             & 66.67                          \\
\rowcolor{blue-c}
LLaVA 1.6-7B  & DPO        & VLFeedbackCorrupted (5k)                 & 87.52           & 3.03                  & 3.01                     & \uline{88.64}                    & \uline{65.30}               & 64.19                 & 42.16             & 70.13                          \\
\rowcolor{blue-c}
LLaVA 1.6-7B  & DPO        & BDHS (VLFeedback, 5k)                    & 88.10   & 2.77                  & 2.87                   & 86.68                  & 65.27                       & \uline{64.33}                 & 43.39             & 72.43                          \\

\bottomrule
\end{tabular}
}

\caption{Main results.  The best and second best results are
shown in \textbf{bold} and \uline{underlined}, respectively. If a larger model outperforms all aligned 7B models, it is indicated by \textbf{\uline{bold and underline}}. \textdagger\ denotes results reported from referenced papers, and a dash (--) marks benchmarks that are not reported. Rows in blue are contributions of this paper.}
\label{table:main_table}
\vspace{-3mm}
\end{table*}

%% file: tables/datasets_baselines.tex
\begin{table*}[t!]
\centering
\resizebox{\textwidth}{!}{%
\begin{tabular}{l c c c c c c c c  c}
\toprule
Model                                                                               & Dataset                     & POPE $\uparrow$ & MMHAL $\uparrow$              & MMHAL$^\text{V}\uparrow$ & LLaVA$^\text{W}\uparrow$ & VQA$^\text{T}\uparrow$ & GQA $\uparrow$ & MMVet $\uparrow$ & Recall$^{\text{coco}}\uparrow$\\
\midrule
LLaVA 1.6-7B                                                                        & --                          & 86.40           & 2.95                          & 2.75                     & 80.85                    & 64.85                  & 64.23          & 43.94            &  68.13                        \\

\midrule
\multicolumn{8}{l}{\textbf{Public datasets}}                                       \\
\midrule
LLaVA 1.6-7B                                                                        & VLFeedback (80k)            & 81.84           & 2.96                          & 2.99                     & 90.55                    & 62.93                  & 62.54          & 43.85            &  66.67                       \\
LLaVA 1.6-7B                                                                        & POVID (17k)                 & 88.09           & 3.16                          & 3.07                     & 78.05                    & 64.56                  & 64.12          & 40.60            &  73.48                       \\
LLaVA 1.6-7B                                                                        & RLHF-V (5.7k)               & 83.86           & 3.15                          & 3.26                     & 70.58                    & 64.75                  & 62.89          & 37.16            &  64.26                       \\

\midrule
\multicolumn{8}{l}{\textbf{Public datasets, randomly subsampled to 5,000 samples}} \\
\midrule
LLaVA 1.6-7B                                                                        & VLFeedback (5k)             & 86.31           & 2.92                          & 3.00                     & 83.10                    & 65.06                  & 64.09          & 43.21            &  68.03                        \\
LLaVA 1.6-7B                                                                        & POVID (5k)                  & 88.18           & 2.93                          & 2.93                     & 81.89                    & 64.90                  & 64.34          & 43.39            &  71.80                        \\
LLaVA 1.6-7B                                                                        & RLHF-V (5k)                 & 84.39           & 3.25                          & 3.35                     & 72.09                    & 64.85                  & 63.35          & 39.72            &  64.68                        \\

\midrule
\multicolumn{8}{l}{\textbf{Previously published}}                                  \\
\midrule
\rowcolor{lightlightgray}
Qwen-VL-Chat                                                                        & VLFeedback~\citep{Li_2023d} & --              & 3.02                          & --                       & --                       & --                     & --             & 49.9             & --                            \\
\rowcolor{lightlightgray}
Muffin                                                                              & RLHF-V~\citep{Yu_2023}      & --              & (52.1$\downarrow$)\textdagger & --                       & --                       & --                     & --             & --               & --                            \\
\rowcolor{lightlightgray}
LLaVA 1.5                                                                           & POVID~\citep{Zhou_2024}     & 86.90           & 2.69                          & --                       & 68.7                     & --                     & --             & 31.8             & --                            \\
\bottomrule
\end{tabular}
}
\caption{Results for LLaVA 1.6-7B Vicuna~\citep{Liu_2024b} aligned with DPO on VLFeedback, POVID, RLHF-V. Results highlighted in gray are the results reported by the original authors. \textdagger~denotes \texttt{MMHALBench} for which \cite{Yu_2023} strictly reported the human-corrected hallucination rate.}
\label{table:datasets_baselines}
\end{table*}

%% file: tables/datasets_settings.tex
\begin{table*}[!h]
    \centering
    \resizebox{\textwidth}{!}{%
    \begin{tabular}{l | cc | ccc | ccc}
        \toprule
        Datasets                                      & \multicolumn{2}{c}{Prompts} & \multicolumn{3}{c}{Chosen Responses} & \multicolumn{2}{c}{Rejected Responses} \\
                                                      & diverse                     & LLaVA-SFT                            & diverse                                 & LLaVA      & GPT-4V     & diverse    & chosen corrupted by GPT-4 \\
        \midrule
        VLFeedback                                    & \checkmark                  &                                      & \checkmark                              &            &            & \checkmark &                           \\
        + corrupting strategy                         & \checkmark                  &                                      & \checkmark                              &            &            &            & \checkmark                \\
        \midrule
        \multicolumn{8}{l}{\textbf{prompts}}         \\
        \midrule
        LLaVA prompts                                 &                             & \checkmark                           & \checkmark                              &            &            &            & \checkmark                \\
        \midrule
        \multicolumn{8}{l}{\textbf{model responses}} \\
        \midrule
        GPT-4V responses only                         &                             & \checkmark                           &                                         &            & \checkmark &            & \checkmark                \\
        LLaVA responses only                          &                             & \checkmark                           &                                         & \checkmark &            &            & \checkmark                \\
        \bottomrule
    \end{tabular}}
    \caption{Controlled settings for multimodal preference dataset exploration. We decompose the preference datasets into prompts, chosen and rejected responses and we then aim at identifying factors contributing to the dataset quality.}
    \label{table:datasets_settings}
\end{table*}

%% file: tables/ablation_datasets.tex
\begin{table*}[!h]
\centering
\resizebox{\textwidth}{!}{%
\begin{tabular}{l | c c c c c c c c }
\toprule
                                               Dataset  & POPE $\uparrow$ & MMHAL      $\uparrow$ & MMHAL$^\text{v}\uparrow$ & LLaVA$^\text{W}\uparrow$ & VQA$^\text{T}\uparrow$ & GQA $\uparrow$ & MMVet $\uparrow$ & Recall$^{\text{coco}}\uparrow$ \\
\midrule
Baseline                                                                        & 86.40           & 2.95                          & 2.75                     & 80.85                    & 64.85                  & 64.23          & 43.94            &  68.13                        \\

VLFeedback (5k)                                         & 86.31           & 2.92                  & 3.00                     & 83.10                    & 65.06                  & 64.09          & 43.21            & 68.03                          \\
+ corrupting strategy                                   & 85.59           & 3.39                  & 3.33                     & 86.65                    & 65.20                  & 63.87          & 37.98            & 68.66                          \\
\midrule
\multicolumn{8}{l}{\textbf{prompts}}                   \\
\midrule
LLaVA prompts                                           & 87.63           & 2.85                  & 2.96                     & 86.55                    & 65.13                  & 64.25          & 41.47            & 70.44                          \\
\midrule
\multicolumn{8}{l}{\textbf{model responses}}           \\
\midrule
GPT-4V responses only                                   & 86.78           & 3.30                  & 3.02                     & 86.77                    & 65.06                  & 64.02          & 40.14            & 69.08                          \\
LLaVA responses only                                    & 87.52           & 3.03                  & 3.01                     & 88.64                    & 65.30                  & 64.19          & 42.16            & 70.13                          \\
\bottomrule
\end{tabular}
}
\caption{Dataset ablations. We started from the public VLFeedback dataset with its diverse prompts and responses, and we then applied targeted sampling and corruption to isolate the factors contributing to the quality of a preference dataset. }
\vspace{-2mm}
\label{table:ablation_datasets}
\end{table*}

%% file: tables/llm_bias.tex
\begin{table*}[!h]
\centering
\resizebox{\textwidth}{!}{%
\begin{tabular}{c c |c c c c c c c c} 
\toprule
$y^-$ from external supervision & $\tilde{y}^-$ derived from policy            & POPE$\uparrow$ & MMHAL$\uparrow$ & MMHAL$^V\uparrow$ & LLaVA$^\text{W}$$\uparrow$ & VQA$^T\uparrow$ & GQA$\uparrow$  & MMVet$\uparrow$ & Recall$^{\text{coco}}\uparrow$ \\
\midrule
--                              & --                                           & 86.40          & \textbf{2.95}   & 2.75              & 80.85                      & 64.85           & {64.23}        & 43.94           & 68.13                          \\
\midrule
--                              & $\text{BDHS}_{\text{noise}}$ (Offline, ours) & 88.60          & 2.37            & 2.48              & 84.53                      & 65.05           & 64.14          & 41.38           & \uline{75.16}                  \\
--                              & $\text{BDHS}_{\text{attn}}$ (Offline, ours)  & \uline{88.75}  & 2.61            & 2.71              & \textbf{86.33}             & 65.07           & 63.97          & 43.4            & \textbf{75.58}                  \\
--                              & $\text{BDHS}_{\text{attn}}$ (Online, ours)   & \textbf{88.83} & 2.80            & \textbf{2.99}     & 85.03                      & 65.09           & 63.65          & 43.12           &        74.09                        \\
\midrule
\midrule
GPT-4V (POVID)                   & --                                           & 88.18          & \uline{2.93}    & \uline{2.93}      & 81.89                      & 64.90           & \textbf{64.34} & 43.39           & 71.80                          \\
GPT-4V (POVID)                   & POVID-style image distortion                 & 88.33          & 2.84            & 2.64              & 80.15                      & 64.21           & 63.79          & 41.28           &         69.39                       \\
GPT-4V (POVID)                   & $\text{BDHS}_{\text{noise}}$ (Offline, ours) & 88.58          & 2.76            & 2.45              & 84.36                      & 65.31           & 64.26          & \uline{43.95}   & 75.05                           \\
GPT-4V (POVID)                   & $\text{BDHS}_{\text{attn}}$ (Offline, ours)  & 88.56          & 2.85            & 2.72              & 85.35                      & \uline{65.39}   & 64.11          & 43.26           & 75.05                           \\
GPT-4V (POVID)                   & $\text{BDHS}_{\text{attn}}$ (Online, ours)   & 88.38          & 2.82            & 2.81              & 84.01                      & \textbf{65.42}  & \uline{64.30}  & \textbf{45.46}  &    74.00                            \\
\bottomrule
\end{tabular}
}
\caption{Ablation results for BDHS including baseline and reference approaches. All results based on LLaVA 1.6-7B, using DPO and the POVID (5k) sample for the source of images and prompt. Whenever both $y^-$ from external supervision and $\tilde{y}^-$ derived from policy (either online or offline) are incorporated, the average loss is computed using~\eqref{eq:avg_dpo}.}
\label{tab:bdhs_ablation}
\end{table*}

%% file: tables/rm_validation_accuracy.tex
\begin{table*}[t!]
\centering
\captionsetup{font=small}
\resizebox{0.6\textwidth}{!}{
\begin{tabular}{l c c c c}
\toprule
\multirow{2}{*}{Base Model} & \multirow{2}{*}{Train Dataset} & \multicolumn{3}{c}{Held-Out Eval Dataset} \\
                            &                                & POVID                                      & RLHF-V & VLFeedback \\

\midrule
LLaVA 1.5-7B                & POVID                          & 0.99                                       & 0.24   & 0.56       \\
LLaVA 1.5-7B                & RLHF-V                         & 0.12                                       & 0.86   & 0.52       \\
LLaVA 1.5-7B                & POVID + RLHF-V                 & 0.98                                       & 0.76   & 0.53       \\
LLaVA 1.5-7B                & VLFeedback                     & 0.61                                       & 0.54   & 0.81       \\
\midrule
LLaVA 1.6-7B                & POVID                          & 0.99                                       & 0.34   & 0.59       \\
LLaVA 1.6-7B                & POVID + RLHF-V                 & 0.97                                       & 0.68   & 0.63       \\
LLaVA 1.6-7B                & VLFeedback                     & 0.76                                       & 0.53   & 0.82       \\
\bottomrule
\end{tabular}
}
\caption{Reward model accuracy on the held-out validation set.}
\label{table:rm_validation_accuracy}
\end{table*}

%% file: tables/rl_alignment.tex
\begin{table*}[ht]
\centering
\resizebox{\textwidth}{!}{%
\begin{tabular}{l c c c c c c c c c c}
\toprule
Alignment & Dataset$_{RM}$ & Dataset$_{P}$ & POPE$\uparrow$                                                     & MMHAL$\uparrow$ & MMHAL$^V\uparrow$ & LLaVA$^\text{W}$$\uparrow$ & VQA$^T\uparrow$ & GQA$\uparrow$ & MMVet$\uparrow$ & Recall$^{\text{coco}}\uparrow$ \\
\midrule
Baseline  & --                  & --            & 86.40                                                              & 2.95            & 2.75              & 80.85                      & 64.85           & 64.23         & 43.94           & 68.13                          \\

DPO       & --                  & POVID         & 88.09                                                              & 3.16            & 3.07              & 78.05                      & 64.56           & 64.12         & 40.60           & 73.48                          \\

\midrule

\rowcolor{lightgray}
PPO       & POVID               & POVID         &                                                                    &                 &                   &                            &                 &               &                 &                                \\

\rowcolor{lightgray}
RLOO      & POVID               & POVID         & \multicolumn{8}{c}{\multirow{-2}{*}{Policy training not stable} } \\
\midrule
PPO       & VLFeedback          & POVID         & 87.54                                                              & 3.02            & 3.09              & 80.17                      & 63.90           & 64.04         & 40.51           & 67.19                          \\
RLOO      & VLFeedback          & POVID         & 87.17                                                              & 2.94            & 2.72              & 78.72                      & 63.59           & 63.72         & 42.25           & 64.57                          \\

\bottomrule
\end{tabular}
}
\caption{RL-based alignment of LLaVA 1.6-7B, DPO baseline included for reference. RL-based alignment methods use a reward model based on LLaVA 1.6-7B, Dataset$_{RM}$ refers to the dataset used to train the reward model, Dataset$_P$ to the set of images and prompts used for RL alignment.}
\label{table:alignment_rl}
\end{table*}

%% file: tables/mixed_dpo.tex
\begin{table*}[t!]
\centering
\resizebox{\textwidth}{!}{%
\begin{tabular}{c c c c c c c c c c}
\toprule
Alignment  & Dataset & POPE $\uparrow$ & MMHAL$\uparrow$ & MMHAL$^V\uparrow$ & LLaVA$^\text{W}\uparrow$ & VQA$^T\uparrow$ & GQA $\uparrow$ & MMVet $\uparrow$ & Recall$^{\text{coco}}\uparrow$ \\

\midrule
--         & --      & 86.41           & 3.06            & 2.71              & 78.96                    & 64.22           & 64.22          & 43.94            & 68.13                          \\
\midrule
DPO        & POVID   & 88.09           & 3.16            & 3.07              & 78.63                    & 64.56           & 64.12          & 40.60            & 73.48                          \\
Online-DPO & POVID   & 86.49           & 2.88            & 2.94              & 82.61                    & 64.88           & 64.31          & 43.26            & 68.45                          \\
Mixed-DPO  & POVID   & 88.03           & 2.83            & 3.10              & 82.75                    & 64.93           & 64.47          & 42.80            & 74.53                          \\
\midrule
DPO        & RLHF-V  & 83.86           & 3.15            & 3.26              & 70.58                    & 64.75           & 62.89          & 37.16            & 64.26                          \\
Online-DPO & RLHF-V  & 85.40           & 3.10            & 3.27              & 79.66                    & 64.94           & 64.05          & 41.01            & 68.13                          \\
Mixed-DPO  & RLHF-V  & 85.57           & 2.94            & 3.16              & 78.46                    & 65.06           & 64.10          & 41.10            & 67.82                          \\

\bottomrule
\end{tabular}
}

\caption{The effect of Mixed-DPO, using LLaVA 1.6-7B as the base model.}
\label{table:mixed-DPO}
\end{table*}

%% file: tables/anno_eval.tex
\begin{wraptable}{r}{0.5\textwidth}
\captionsetup{belowskip=-5pt, aboveskip=1pt, font=small}
  \centering
  \resizebox{0.5\textwidth}{!}{%
  \begin{tabular}{ccc}
    \toprule
    Dataset           & LLaVA 1.6-7B & LLaVA 1.6-34B \\
    \midrule
    VLFeedback (eval) & 79.10        & 90.91         \\
    RLHF-V (eval)     & 81.88        & 93.90         \\
    POVID (eval)      & 92.96        & 98.55         \\
    \bottomrule
  \end{tabular}
  }
  \caption{Performance of the annotators on different preference datasets.}
  \label{table:anno_eval}
\end{wraptable}

%% file: tables/annotator_ablation.tex
\begin{table*}[ht]
\centering
\resizebox{\textwidth}{!}{%
\begin{tabular}{l c c c c c c c c c ccc}
\toprule
Model        & Dataset & Annotator     & POPE $\uparrow$ & MMHAL$\uparrow$ & MMHAL$^V\uparrow$ & LLaVA$^\text{W}\uparrow$ & VQA$^T\uparrow$ & GQA $\uparrow$ & MMVet $\uparrow$ & Recall$^{\text{coco}}\uparrow$ \\

\midrule
LLaVA 1.6-7B  & --         & --                                       & 86.40           & 2.95                  & 2.75                     & 80.85                    & 64.85                       & 64.23                 & 43.94             & 68.13                          \\         
\midrule
LLaVA 1.6-7B & POVID   & LLaVA 1.6-7B  & 86.54           & 2.52            & 2.72              & 81.55                    & 64.93           & 64.18          & 40.73            & 67.40                          \\
LLaVA 1.6-7B & POVID   & LLaVA 1.6-34B & 86.49           & 2.88            & 2.94              & 82.61                    & 64.88           & 64.31          & 43.26            & 68.45                          \\

\bottomrule
\end{tabular}
}
\caption{Comparison of Online-DPO with a strong annotator (i.e., LLaVA 1.6-34B) and a weak annotator (i.e., LLaVA 1.6-7B).}
\label{table:anno_ablation}
\end{table*}

%% file: tables/alignment_offline.tex
\begin{table*}[ht]
\centering
\resizebox{\textwidth}{!}{%
\begin{tabular}{c c c c c c c c c ccc}
\toprule
Alignment & Dataset & POPE $\uparrow$ & MMHAL$\uparrow$ & MMHAL$^V\uparrow$ & LLaVA$^\text{W}$$\uparrow$ & VQA$^T\uparrow$ & GQA $\uparrow$ & MMVet $\uparrow$ & Recall$^{\text{coco}}\uparrow$ \\

\midrule
--        & --      & 86.40           & 2.95            & 2.75              & 80.85                      & 64.85           & 64.23          & 43.94            & 68.13                          \\
\midrule
DPO       & POVID   & 88.09           & 3.16            & 3.07              & 78.63                      & 64.56           & 64.12          & 40.60            & 73.48                          \\
IPO       & POVID   & 87.62           & 3.11            & 3.11              & 82.34                      & 65.09           & 64.47          & 43.99            & 69.81                          \\
SliC      & POVID   & 88.28           & 3.17            & 3.15              & 81.99                      & 64.59           & 64.11          & 41.51            & 74.32                          \\
\bottomrule
\end{tabular}
}
\caption{Comparison of different offline alignment methods based on LLaVA 1.6-7B.}
\label{table:alignment_offline}
\end{table*}

%% file: sections/5_conclusion.tex
\section{Conclusion and Future Work}

In this study, we explore the role of preference alignment in enhancing the performance of MLLMs, with a particular focus on reducing hallucinations. A commonly proposed explanation for hallucinations in MLLMs is their tendency to overlook image content and instead rely on inherent language biases. To address this, we assess various alignment strategies across different datasets and alignment methods. We categorize alignment algorithms into offline and online strategies and demonstrate that a hybrid approach can offer benefits in specific scenarios. We also do a thorough study on the existing multimodal preference datasets, identifying strengths and weaknesses associated with each, and providing insights into how certain types of preference data can enhance model performance.

Leveraging these insights, we develop our own preference dataset and introduce a novel data sampling strategy, BDHS. When applied to the LLaVA 1.6 model, these methods lead to notable improvements across various benchmarks, confirming the potential of tailored preference alignment strategies in refining the capabilities of MLLMs. A significant advantage of this approach is its ability to operate effectively using only SFT data, eliminating the need for a superior model, human labelers, or other complex means of constructing preference data.

This study not only enhances our understanding of preference alignment but also establishes a foundation for further research into MLLM preference alignment. Specifically, we identify several gaps in the community's approach to aligning MLLMs:

\begin{itemize}[nosep,leftmargin=0.5cm]
\item While considerable research has been conducted on various alignment methods, including both online and offline approaches, for LLMs, these studies are less common in the context of MLLMs. For instance, RLH(AI)F is extensively discussed in LLM literature, highlighting its potential over the more simple methods like DPO \citep{Ahmadian_2024, xu2024dpo}. We have provided some insights into RL-based alignment for MLLMs and the evaluation of reward models, yet we believe there remains a significant gap between LLM and MLLM research in this domain.

\item A better hallucination benchmark can help our understanding of model improvements. We discuss some of the shortcomings of current hallucination benchmarks in Sections~\ref{section:appendix-pope},~\ref{section:appendix-chair} and~\ref{section:appendix-mmhalbenchv_motivation}. However, the development of an effective hallucination benchmark remains an active area of research.

\item We thoroughly analyze various aspects of published multimodal preference data. However, the coverage of these datasets is still insufficiently studied. The lack of comprehensive coverage in these datasets may contribute to the absence of significant improvements in some benchmarks.
\end{itemize}

This paper has highlighted key advancements and existing challenges in preference alignment for MLLMs. Our findings point out important gaps that need addressing. We hope these insights inspires further research and helps the community tackle ongoing challenges in this field.

%% file: sections/acknowledgements.tex
\section*{Acknowledgments}

The authors would like to thank Sebastian Brechtel, Meng Cao, Philipp Dufter, Jiaming Hu, Lukas Jendele, Juan Lao Tebar, Shuang Ma, Dhruti Shah, Will Song, Juergen Wiest, and Haotian Zhang for their feedback and guidance throughout this project. We would also like to thank the authors of \citet{Zhou_2024}, \citet{Li_2023d}, and \citet{Yu_2023} for their prompt responses and effective communication with us during the process.

%% file: sections/appendix.tex
\appendix

\section{Implementation Detail} \label{section:appendix-implementation-detail}
For all offline experiments, as well as for Online-DPO and Mixed-DPO, we conducted a hyper-parameter search. The parameters included learning rates of $10^{-7}$, $5 \cdot 10^{-7}$, $10^{-6}$, $5 \cdot 10^{-6}$; projection layer learning rates of $2 \cdot 10^{-5}$, $2 \cdot 10^{-6}$, $2 \cdot 10^{-7}$; epochs of 3, 5, and 7; and batch sizes of 16, 32, and 48. We reported the best results for each method. Additionally, we set the LoRA rank and scaling factor to 128 and 256, respectively. The $\beta$ values for DPO, IPO, and SLiC were explored at $0.05$, $0.1$, $0.2$, $0.5$ for DPO; $0.8$, $0.9$, $1.0$ for IPO; and $0.02$, $0.1$, $0.2$ for SLiC.

For RL methods (PPO and RLOO), we maintained constant base model parameters while training LoRA adapters for alignment. Specifically, for RLOO, we utilized $k=4$, generating four distinct responses for each prompt at a temperature of 1.0. Training was conducted over two epochs with a batch size of 256 and a learning rate of $3 \cdot 10^{-4}$. Prior to RLOO training, we calculated the mean and standard deviation of rewards using the alignment dataset and normalized the rewards during training to achieve zero mean and unit variance. We determined that a $\beta$ value of $0.4$ provided the best balance between rewards and the KL penalty for RLOO. Gradient clipping was also implemented to cap the maximum gradient norm at 1.0.

For PPO specifically, we trained for 3 epochs with a learning rate of 1.41e-5 using a constant learning rate schedule. We used 1 GPU with a batch of 32. For the reward model, we used a a learning rate of 2e-5 and trained for 4000 steps. The learning rate schedule was also adjusted to be constant but with a warmup phase. The fraction value for the warmup phase is set at 0.03. Training was conducted on 8 GPUs with a batch size of 32.

\section{Evaluation}
\subsection{POPE}
\label{section:appendix-pope}

We noticed the existence of an upper bound on the \texttt{POPE} benchmark, as most of our experiments would reach a plateau between 86\% and 88\% despite improvements on other benchmarks. We manually looked at the losses among 100 responses and present the results in this section.

\begin{figure}[!h]
    \centering
   \resizebox{0.99\textwidth}{!}{\input{figures/pope.tex}}
   \caption{Upon analysis of the losses on \texttt{POPE}, we noticed close to 20\% of cases where the ground truth was either incorrect or disputable. This example is from \texttt{POPE}, which sources images from COCO~\citep{Lin_2014}.}
    \label{fig:appendix-pope}
\end{figure}
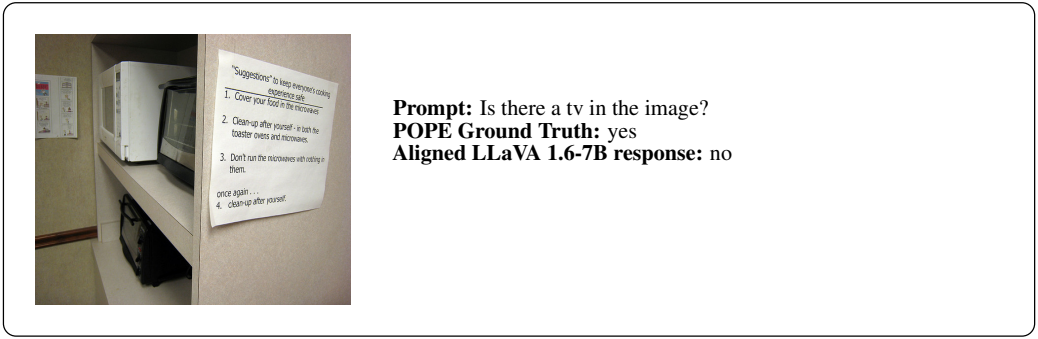

In 20\% cases, we observed that the ground truth was either incorrect or disputable. In some of those cases, it appeared that the ontology used to build \texttt{POPE} could potentially result in differing interpretations. For example, in certain countries, a clear distinction exists between a car and a truck, although this distinction is not as pronounced in other regions of the world\footnote{An example of such distinction between car/truck can be seen on \href{http://images.cocodataset.org/val2014/COCO_val2014_000000210789.jpg}{COCO\_val2014\_000000210789.jpg} where the \texttt{POPE} ground truth expects "no" to the prompt "Is there a car in the image?".}. We provided an example along the response of our aligned model\footnote{We used a LLaVA 1.6-7B DPO-aligned on LLaVA prompts and responses sampled from VLFeedback. See Section \ref{section:dataset_ablations}.} in Figure~\ref{fig:appendix-pope}.

Provided these examples are eliminated, we think it is plausible that performant models could potentially exceed a 90\% accuracy rate on the \texttt{POPE} benchmark. Re-annotating those examples is beyond the scope of this work, however we would like to invite the community to consider it as many recent SOTA models exhibit such plateau. See Table 4 in \citep{Mckinzie_2024} where all the models reported are demonstrating such plateau on \texttt{POPE}.

\subsection{CHAIR and Object HalBench}
\label{section:appendix-chair}
We evaluated two widely used benchmarks in the community for measuring hallucination, focusing on the computation of \texttt{CHAIR} metrics. We investigated approaches described by \cite{Rohrbach_2018}, which uses COCO annotations to compute \texttt{CHAIR} scores, and the more recent method by \cite{Yu_2023}, named \texttt{Object HalBench}, which combines COCO annotations with a GPT model to enhance the detection of hallucinated objects. 

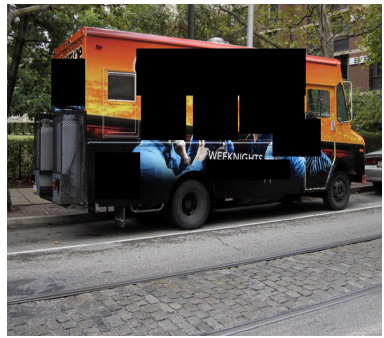
\begin{figure}[t]
    \centering
   \resizebox{0.99\textwidth}{!}{\input{figures/chair.tex}}
   \caption{Examples illustrating instances where the \texttt{CHAIR} and \texttt{Objet HalBench} benchmarks fail to detect hallucinations. Text highlighted in \textcolor{green}{green} identifies hallucinations successfully detected by the benchmarks. In contrast, text highlighted in \textcolor{red}{red} indicates examples where the benchmark failed to identify hallucinations. \textcolor{orange}{Orange} indicates hallucinations that, though not targeted by these benchmarks, degrade response quality. The top example shows the benchmark proposed by~\cite{Yu_2023} while the bottom example follows from \citep{Rohrbach_2018}. Images are from COCO~\citep{Lin_2014}.}
    \label{fig:chair}
\end{figure}

Our analysis reveals that both benchmarks are significantly noisy (Figure \ref{fig:chair}). We also found that any improvements in \texttt{CHAIR} scores strongly depend on the ability of these benchmarks to detect specific types of hallucinations and cannot be attributed solely to the improvement of the model.

Furthermore, it is common to report \texttt{CHAIR} metrics without including recall metrics. Considering the trade-off between \texttt{CHAIR} and recall, omitting recall does not provide a full picture of how much a model has improved in reducing hallucinations. For instance, a model that generates short and conscise responses might not produce many hallucinations, but this may be at the cost of potentially providing an unhelpful answer.

Hence, the recall metric from \citet{Rohrbach_2018} proves particularly informative for comparing different models and helping with our understanding of other benchmarks. We report this metric in our evaluations, styled Recall$^{\text{coco}}$ in our tables.

\subsection{MMHALBench-Vision}
\label{section:appendix-mmhalbenchv_motivation}

The original \texttt{MMHALBench} benchmark \citep{Sun_2023} uses GPT-4 to judge whether model responses introduce hallucinations. In that text-only regime, \texttt{MMHALBench} relies on ground truth information about the pictures, such as the categories of the objects present or a human reference response to the prompt. 

We evaluated manually the common wins and losses obtained on \texttt{MMHALBench} during our experiments and noticed that in $\sim$20\% cases we disagree with the resulting \texttt{MMHALBench} score\footnote{21 cases out of 96 while comparing wins and losses of two models.}. We found cases where responses with hallucinations were considered as correct. Oppositely, we found cases where valid answers were wrongly tagged as containing hallucinations. In many cases, we saw the helpfulness to be under-estimated. See Figure~\ref{fig:appendix-mmhal}.

This can be explained due the ground truth information being only expressed through text causing the judge model, GPT-4, to wrongly tag or miss hallucinations. To mitigate such cases, we introduced \texttt{MMHALBench-Vision}: we rely on the recently introduced GPT4-o to consume the image along the text ground truth information. We kept the evaluation prompt and scoring identical.

\begin{figure}
    \centering
   \resizebox{0.99\textwidth}{!}{\input{figures/mmhal.tex}}
   \caption{Example of incorrect assessments with \texttt{MMHALBench}. The top example shows a response incorrectly assessed as having a hallucination. The second presents a response where the model hallucinates a side view of the backpack, but that is missed by \texttt{MMHALBench}. In both cases, the text ground truth does not give enough information to the text-only evaluator to evaluate the response. Reproduction of \texttt{MMHALBench} images cannot take place here due to usage restrictions. Please refer to \url{https://huggingface.co/datasets/Shengcao1006/MMHal-Bench}.}
    \label{fig:appendix-mmhal}
\end{figure}

\section{Preference Data}
\subsection{Dataset Prompt Corruption}
\label{section:appendix-dataset-prompt-corruption}

We leverage GPT-4 to corrupt chosen responses with realistic and plausible hallucinations. We call \textit{realistic hallucinations} those instances where a human, just by looking at the corrupted response, is unable to recognize it without having to refer back to the image. We have remarked this was an important distinction: the more obvious the corruptions are, the poorer the performance of the resulting policy is. We launched side experiments where we employed a less skilled LLM corrupter and incorporated obvious tags\footnote{An example of obvious rejected response marked with a tag: \textit{There are only two examples of plates that are visible in the image. \textbf{This response is factually incorrect.}}} into the responses. In both scenarios, we noticed a drop in performance as the corruption gets less realistic and readily `hackable' by the policy under alignment. The prompt used to corrupt the chosen responses is reproduced in Table~\ref{table:dataset-prompt-corruption}.

\begin{table*}[!h]
\small
    \centering
    \begin{tcolorbox}
You are provided with a prompt and a response to that prompt. You task is to corrupt the response with realistic hallucinations.\\
\\
Hallucinations can be including objects/persons/animals not present in the original response, changing the positions of the objects/persons/animals, changing the count, changing the attributes (e.g. color, shapes, etc.) or their spatial relationships. Those hallucinations have to be realistic.\\
\\
In other words, your task is to change the provided response in a way that could be correct given the prompt, but is not. A person only seeing the corrupted response shall not be able to identify it as being incorrect, so your change must be a plausible (although again incorrect). Output only the updated, corrupted response, nothing else.\\
\\
Here are a few examples:\\
\#\#\# Example 1\\
prompt: What is the title of the book mentioned in the image?\\
response: The title of the book in the image is "Picture This: A Gallery of Fusible Appliqué Projects for Quilting and Framing" by Jean Wells \& Marina Anderson.\\
corrupted response: The title of the book in the image is "Sewing Masters: A Collection of Innovative Embroidery Designs for Quilting and Framing" by Marvin Wells \& Vincent Anderson.\\
\\
\#\#\# Example 2\\
prompt: You are a driving assistant. Based on current image, what is the best action to take when you are driving on the road? A. Slow down B. Turn around C. Stop the car D. Turn Left E. Keep driving. Please give reasons and the answer\\
response: The best action to take when driving on the road, based on the current image, would be: E. Keep driving.\\Reasons for this action:\\1. The traffic light visible in the image is green, indicating that it is legal and safe to proceed.\\2. The speed limit sign shows "60" which means the car, currently at 20 km/h, is well below the maximum speed limit and can safely continue and even increase speed if necessary.\\3. The road ahead is clear of any immediate hazards or obstructions that would require stopping or turning.\\
corrupted response: The best action to take when driving on the road, based on the current image, would be: C. Stop the car.\\Reasons for this action:\\1. Although the traffic light on the left in the image is green, the one on your side is red, indicating it is necessary to stop\\2. The speed limit sign shows "60" which means the car, currently at 20 km/h, is well below the maximum speed limit and can safely stop before the intersection.\\3. The intersection up ahead indicates the presence of crossing cars, requiring a stop.\\
\\
\#\#\# Example 3\\
prompt: \{original\_prompt\}\\
response: \{original\_response\}\\
corrupted response:\\
    \end{tcolorbox}
\caption{Prompt used to corrupt datasets with GPT-4.}
\label{table:dataset-prompt-corruption}
\end{table*}

\subsection{Dataset Size ablation with the Corrupting Strategy}
\label{section:appendix-dataset-size-ablation}

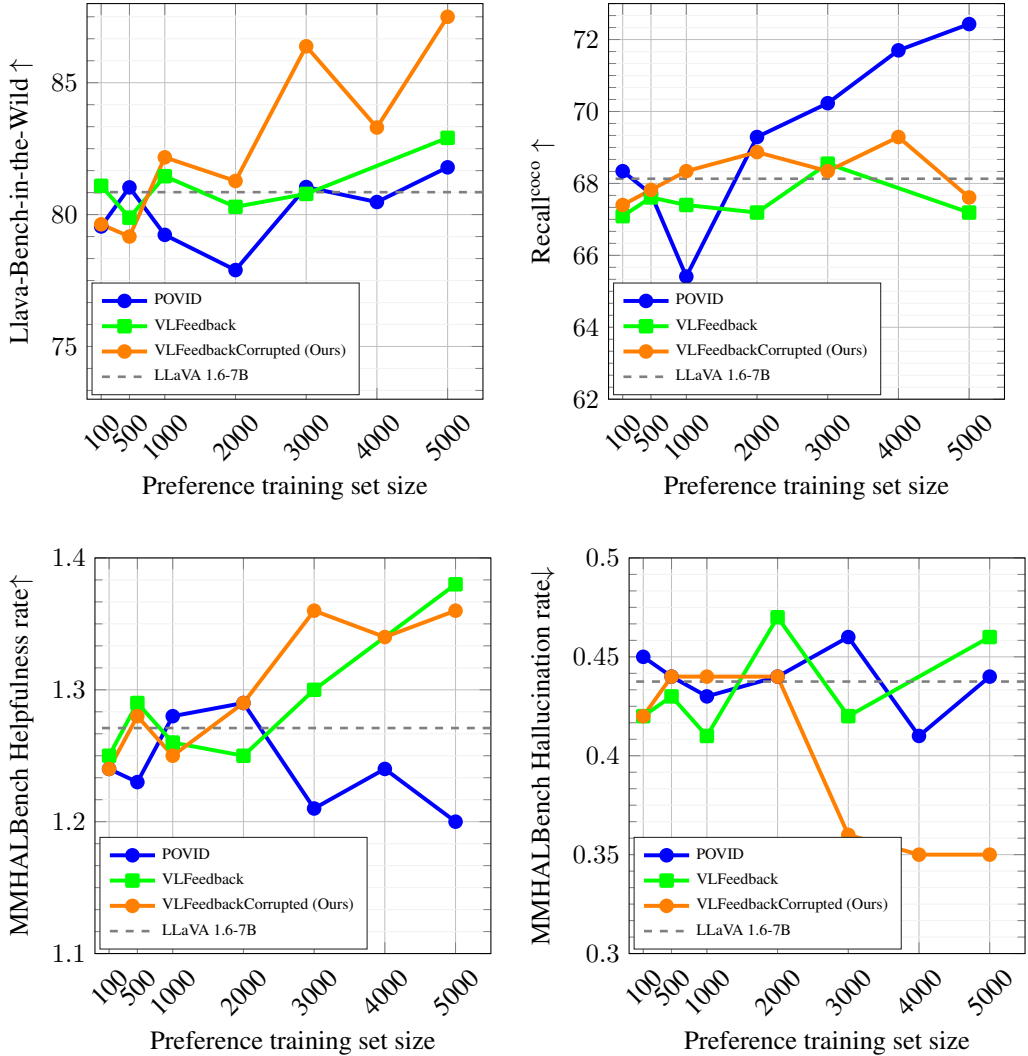
\begin{figure}[!h]
\centering
\begin{subfigure}{.49\textwidth}
\input{figures/plot_impact_size_llava}
\end{subfigure}
\begin{subfigure}{.49\textwidth}
\input{figures/plot_impact_size_recall}
\end{subfigure}
\begin{subfigure}{.49\textwidth}
\input{figures/plot_impact_size_helpfulness}
\end{subfigure}
\begin{subfigure}{.49\textwidth}
\input{figures/plot_impact_size_hallucinations}
\end{subfigure}
\caption{Impact of the preference dataset size. Our corrupting strategy outperforms other datasets on \texttt{LLaVA-in-the-Wild} and \texttt{MMHALBench} hallucination rate. It is on par on the  \texttt{MMHALBench} helpfulness rate against vanilla VLFeedback. Finally, POVID reports the highest Recall$^{\text{coco}}$. The dashed lines are the scores for the LLaVA 1.6-7B baseline.}
\label{fig:dataset_size_ablation}
\end{figure}

We conducted a dataset size ablation on the application of our corrupting strategy on VLFeedback (Figure~\ref{fig:dataset_size_ablation}) . We evaluated 7 checkpoints between 100 and 5,000 training samples, our maximum in this data regime (Section \ref{section:dataset_ablations}). We provide the baseline results with a dashed line. While POVID shows the best result on Recall$^{\text{coco}}$, our simple corruption strategy applied outperforms other datasets on both \texttt{LLavaBench-in-the-Wild} and \texttt{MMHALBench} hallucination rate, while being on par on the \texttt{MMHALBench} helpfulness rate with VLFeedback vanilla.

\section{Annotator in Online-DPO}
\label{section:appendix-annotator}

Table \ref{table:online_prompt} shows the prompt we used to obtain online feedback from the annotator. We conducted multiple experiments with different prompts. In one setup, similar to the approach taken by \mbox{\citet{Guo_2024}} with the rewards model, we included the ground truth response as an additional signal for the annotator to evaluate both responses. We did not observe any significant change in either the evaluation metrics or the final performance of the aligned model. This may be due to the fact that most of the open-source MLLMs we used in this study still lack the ability to follow instructions effectively, especially when the instructions contain multiple components or detailed steps.

We also examined the potential bias of the annotator model in choosing "Response 1" or "Response 2" and found no noticeable bias.

Figure \ref{fig:annotator} shows an example of an annotation made by LLaVA 1.6-34B model.

\input{tables/annotator_prompt}

\begin{figure}
    \centering
   \resizebox{0.99\textwidth}{!}{\input{figures/annotator.tex}}
   \caption{Example of Annotation. Image is from COCO~\citep{Lin_2014}.}
    \label{fig:annotator}
\end{figure}

\section{Bias-Driven Hallucination Sampling} \label{section:appendix:bdhs}

\subsection{Adding noise to images} \label{sec:appendix:diffusion}

This section describes how to gradually add noise to images through a diffusion process.
The derivation follows the public implementation of POVID-style image distortion \citep{Zhou_2024} to enable the proper reproduction of their results.

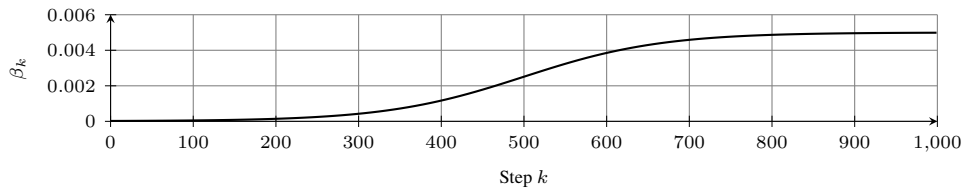
\begin{figure}[!h]
\centering
\input{figures/diffusion_beta}
\caption{Schedule of diffusion parameter $\beta_k$.}
\label{fig:diffusion_beta}
\end{figure}

Let $x_\text{img}(k)$ denote the image after applying noise $k$-times with $x_\text{img}(0)$ referring to the original image and $\mathcal{N}(0,1)$ represent the normal distribution.
Then the forward noise process is defined as:
\begin{equation}
    \label{eq:diffusion}
    x_\text{img}(k) = \sqrt{1-\beta_k} x_\text{img}(k-1) + \sqrt{\beta_k} \epsilon \quad \text{with } \epsilon \sim \mathcal{N}(0,1).
\end{equation}
Hereby, $\beta_k$ denotes a time-variant parameter which is set to $\beta_k = \sigma(-6 + \frac{12 k}{1000}) \cdot (0.5\cdot 10^{-2} - 10^{-5}) + 10^{-5}$
to gradually increase noise between $k=0$ and $k=1000$ (refer to Figure~\ref{fig:diffusion_beta}).

The recursive equation~\eqref{eq:diffusion} can be reformulated to apply $k$ steps of noise at once.
Setting $\alpha_k = 1-\beta_k$ and $\bar{\alpha}_N=\prod_{k=1}^N \alpha_k$,
the following equation applies $N$ steps of noise to image~$x_\text{img}(0)$:
\begin{equation}
    \label{eq:diffusion_final}
    \tilde{x}_\text{img}(N) = \sqrt{\bar{\alpha}_N} x_\text{img}(0) + \sqrt{1-\bar{\alpha}_N} \epsilon \quad \text{with } \epsilon \sim \mathcal{N}(0,1).
\end{equation}
The default for $N$ in \cite{Zhou_2024} is $N=500$.

\subsection{Ensuring Semantically Meaningful Differences}

Section~\ref{sec:bdhs:similarity} describes an iterative technique for BDHS that evaluates similarity scores between the generated response $\tilde{y}^-$ and the ground truth $y^+$.
If both responses are identified as similar according to the sentence embeddings model, a new BDHS response is sampled until a maximum number of iterations $N_\text{BDHS}$ is reached.
The last iteration waives the ground truth reference and generates a full response which is then taken as $\tilde{y}^-$ regardless of the similarity score.
\begin{figure}
\centering
\input{figures/bdhs_similarity}
\caption{Number of resolved similar responses for BDHS generation based on POVID (5k). Parameters are $\epsilon_\text{s}=0.97$ and $N_\text{BDHS}=5$.}
\label{fig:bdhs_similarity}
\end{figure}
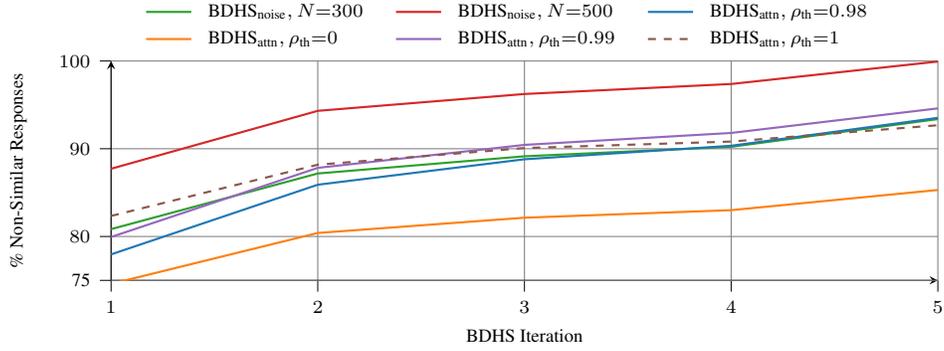
Figure~\ref{fig:bdhs_similarity} shows the number of non-similar responses, i.e. ensuring $\epsilon_\text{s}<0.97$, over the number of BDHS iterations for the full POVID (5k) dataset.
As expected all BDHS variants result in a larger number of non-similar responses compared to the model response without image attention blocking or noisy images.
Running BDHS with a single iteration already results in more than $74\,\%$ semantically different responses.
After four iterations, BDHS variants with restricted image access differ in over $90\,\%$ while the last iteration is guidance free and only depends on the sampled attention mask resp. noise.
Interestingly, in iteration 5, $\text{BDHS}_{\text{attn}}, \rho_{\text{th}}{=}1$ corresponds to guidance-free response generation with fully blocked image tokens which still results in $7\,\%$ similar responses w.r.t.~the SFT ground truth.
Probable reasons for this saturation are either that the correct answer is easy to guess even without access to the image, or that the answer is memorized from the training data.
Note that prompts and images in POVID (5k) are extracted from LLaVA Instruct which served as training data for fine-tuning LLaVA-1.6.

BDHS with noisy images in the input and $N=500$ diffusion steps results in more than $99\,\%$ semantically different responses after five iterations, surpassing the score for the fully blocked response.
This is misleading, as although the responses are indeed semantically different, they mostly mention that the prompt cannot be evaluated due to blurry and noisy images.
Essentially, the noise adds an additional bias towards noise/pixel-referring responses instead of inducing only the desired inherent bias which would saturate at approx.~$93\,\%$ (response with fully masked image tokens).

After three iterations, the score of $\text{BDHS}_{\text{attn}}, \rho_{\text{th}}{=}0.99$ reaches the one from the fully blocked response which is hypothetically implied due to increased diversity by subsampling a distinct attention mask.

Section~\ref{section:appendix:bdhs:examples} presents several examples with actual responses.

\subsection{BDHS Algorithm}

The general overview of BDHS is provided in Figure~\ref{fig:bdhs_overview}. This section introduces the corresponding algorithm which is listed in Algorithm~\ref{alg:bdhs}.
This version includes both, noisy images for $\text{BDHS}_\text{noise}$ and attention masking for $\text{BDHS}_\text{attn}$ (refer to the comments in Algorithm~\ref{alg:bdhs}).
We add a straightforward heuristic to swap \textit{yes} and \textit{no} words whenever they occur in the beginning of a sentence. For this purpose line~\ref{alg:bdhs:yesno}
introduce a regular expression which matches any \textit{yes} or \textit{no} at the beginning of each sentence and optionally skips any preceding newline or whitespace characters.
This expression can be extended to further use-cases if desired.
We choose to generate the full response without any SFT ground truth guidance in the very last iteration whenever $N_{\text{BDHS}}>1$ to minimize similarity (refer to line~\ref{alg:bdhs:lastiter}).

\begin{algorithm}
\caption{BDHS}\label{alg:bdhs}
\small
\begin{algorithmic}[1]
\Require Prompt $x_\text{text}$, image $x_\text{img}$, SFT ground truth $y^+$, attention masking parameter $\rho_\text{th}$, image noise level $N$, BDHS iterations $N_{\text{BDHS}}$, similarity threshold $\epsilon_\text{s}$
\For{$i=1,2,\dotsc,N_{\text{BDHS}}$}
    \State $m \gets \text{Sample image attention mask with } \rho_\text{th}$ according to \eqref{eq:attention_mask} \Comment{$\rho_\text{th}>0$ only for $\text{BDHS}_\text{attn}$}
    \State $\tilde{x}_\text{img} \gets \operatorname{AddNoise}(N, x_\text{img})$ via \eqref{eq:diffusion_final} \Comment{$N>0$ only for $\text{BDHS}_\text{noise}$}
    \State $\tilde{x} \gets (x_\text{text}, \tilde{x}_\text{img}, m)$
    \If{$N_{\text{BDHS}}>1$ and $i=N_{\text{BDHS}}$} \label{alg:bdhs:lastiter}
        \State $y^- \gets$ Generate full model response via $\pi_\theta(\cdot | \tilde{x})$
        \State \Return $y^-$
    \EndIf
    \State $\mathcal{S} \gets$ Split $y^+$ into S sentences
    \State $y^-_k \gets \emptyset$  \Comment{Initialize empty string}
    \ForEach{$y^+_k$ in $\mathcal{S}$} \Comment{Parallelizable}
        \State $\xi \gets \xi \sim \mathcal{U}(0, 1)$ \Comment{$\mathcal{U}(0, 1)$ denotes the uniform distribution in $[0,1]$}
        \If{$y^+_k$ matches \textit{r"\^{}[\textbackslash s]\textasteriskcentered (Yes|yes|No|no)"} and $\xi\geq 0.5$} \Comment{\textit{r"$\cdot$"} denotes a regular expression} \label{alg:bdhs:yesno}
            \State $y^+_k \gets$ Swap corresponding \textit{Yes}/\textit{yes} by \textit{No}/\textit{no} and vice versa
        \EndIf
        \State $y^+_{k, 1} \gets$ Sample random position in $y^+_k$ and return first substring
        \State $y^-_{k, 2} \gets$ Complete sentence via \eqref{eq:bdhs_generation} until full stop or \textit{<eos>}
        \State $y^-_k \gets (y^+_{k, 1}, y^-_{k, 2})$ \Comment{Concatenate strings to full sentence}
        \State $y^- \gets (y^-, y^-_k)$ \Comment{Append to overall response}
    \EndFor
    \State $\phi \gets$ Compute similarity score between $y^-$ and $y^+$ in $[0, 1]$ \Comment{Use sentence embeddings}
    \If{$\phi < \epsilon_\text{s}$}
        \State \textbf{break} \Comment{Semantically different according to threshold}
    \EndIf
\EndFor
\State \Return $y^-$
\end{algorithmic}
\end{algorithm}

\subsection{Additional Ablations}

Additional BDHS ablations, especially regarding different hyperparameter choices are shown in Table~\ref{tab:additional_bdhs_ablation}.
We also evaluate SFT guidance-free generation only with attention masking active. The corresponding benchmark results are listed in the first two rows.
All subsequent rows evaluate the full BDHS approach including SFT guidance. We include ablations that rely on noisy images rather than attention masking, following the diffusion process described in ~\ref{sec:appendix:diffusion}.
\input{tables/llm_bias_appendix}

\subsection{Additional Examples}
\label{section:appendix:bdhs:examples}

This section presents further examples of responses generated from LLaVA instruct prompts and images.
The different variants of BDHS are introduced in Section~\ref{section:BDHS}.
Refer to Figure~\ref{fig:llm_bias_examples_appendix} for the examples.
For guided generation, the colored text is generated purely from the model while the standard text is taken from the SFT ground truth.

\begin{figure}[p]
	\centering

	\begin{tikzpicture}
		\placeexample[
		yshiftbox=0.0cm,
		imgname=COCO_train2014_000000520942,
		prompttext=What color is the jetliner in the image?,
		sfttext=The jetliner in the image is blue and red.,
		bdhsatext={The jetliner in the \corrupted{image is red and blue.}},
		bdhsbtext={The jetliner in the \corrupted{image is white.}},
		bdhsctext={The jetliner in the \corrupted{image is white.}},
		bdhsdtext={The jetliner in the \corrupted{image is white.}},
		bdhsnatext={The jetliner in the \corrupted{image is blue.}},
		bdhsnbtext={The jetliner in the \corrupted{image is not visible due to the high-resolution pixelation.}},
		povidatext={The jetliner in the image is blue. red.},
		povidbtext={The imageliner in the image is not. white.}
		]

		\placeexample[
		yshiftbox=-5cm,
		imgname=COCO_train2014_000000373480,
		prompttext=What is the subject of the image?,
		sfttext={The subject of the image is the Big Ben clock tower, which is a famous attraction in Europe, particularly in London, England.},
		bdhsatext={The subject of the image is \corrupted{the Elizabeth Tower, more commonly known as Big Ben, a famous clock tower located in London, England.}},
		bdhsbtext={The subject of the image is \corrupted{a clock tower with a blue sky background.}},
		bdhsctext={The subject of the image is \corrupted{a tall, narrow tower with a pointed roof, which appears to be a church steeple.}},
		bdhsdtext={The subject of the image is \corrupted{a person's hand holding a smartphone.}} ,
		bdhsnatext={The subject of the image is \corrupted{the Elizabeth Tower, which is the clock tower at the north end of the Palace of Westminster in London, England.}} ,
		bdhsnbtext={The subject of the image is \corrupted{a close-up of a pattern that appears to be a digital or pixelated texture.}},
		povidatext={The subject of the image is the Elizabeth Ben clock tower in which is a famous landraction in London, specifically in London, England. The},
		povidbtext={The image of the image appears a pattern Bang clock tower in which is a famous landraction in London, specifically in London, England. The}
		]

		\placeexample[
		yshiftbox=-13.0cm,
		imgname=COCO_train2014_000000276267,
		prompttext=What street names are visible on the street sign in the image?,
		sfttext={The street names visible on the street sign in the image are "Telegraph Hill" and "Greenwich Road."},
		bdhsatext={The \corrupted{street sign in the image shows "Greenwich" and "Telephone Hill.}},
		bdhsbtext={The \corrupted{street sign in the image shows the intersection of "Telephone Pole" and "Newspaper Box" streets.}},
		bdhsctext={The \corrupted{street sign in the image shows the intersection of "Broadway" and "Spring Street.}},
		bdhsdtext={The \corrupted{street sign in the image shows the intersection of "Main Street" and "Elm Street.}},
		bdhsnatext={The \corrupted{image is quite blurry, but I can make out the street sign.}},
		bdhsnbtext={The \corrupted{image you've provided is very blurry and lacks clear details, making it difficult to read any street names.}},
		povidatext={The image sign visible on the street sign in the image are "Hchn"" and "H Street Street."},
		povidbtext={The image sign are on the street sign in the image are "1chn Avenue" and "Bwich Street."}
		]
	\end{tikzpicture}

	\caption{Examples of generated responses from BHDS ablations and POVID-style image distortion. The image, prompt and SFT ground truth are taken from LLaVA-Instruct-150k, which sources them from CoCo \citep{Lin_2014}.}
	\label{fig:llm_bias_examples_appendix}
\end{figure}

%% file: figures/pope.tex
\begin{tikzpicture}[font=\fontsize{7pt}{7pt}\selectfont]

\tikzstyle{combined_block} = [rectangle, draw, rounded corners, inner sep=10pt, align=left, minimum width=10cm]

    \node[combined_block] (upper_block) at (0,0) {
        \raisebox{-0.7\height}{\includegraphics[width=3.5cm, height=3cm]{figures/COCO_val2014_000000461331.jpg}}
        \hspace{10pt}
        \begin{minipage}[t]{0.48\textwidth}
            \textbf{Prompt:} Is there a tv in the image?\\
            \textbf{POPE Ground Truth:} yes\\
            \textbf{Aligned LLaVA 1.6-7B response:} no
        \end{minipage}
    };

\end{tikzpicture}

%% file: figures/chair.tex
\begin{tikzpicture}[font=\fontsize{7pt}{7pt}\selectfont]

\tikzstyle{combined_block} = [rectangle, draw, rounded corners, inner sep=10pt, align=left, minimum width=10cm]

    \node[combined_block] (upper_block) at (0,0) {
        \raisebox{-0.7\height}{\includegraphics[width=4.5cm, height=4cm]{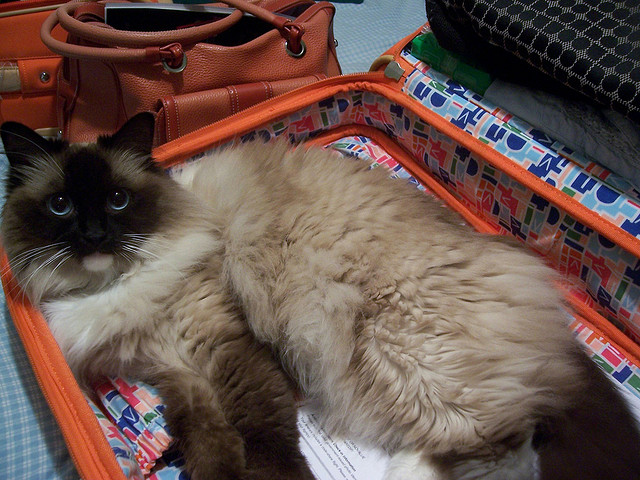}}
        \hspace{10pt}
        \begin{minipage}[t]{0.48\textwidth}
            \textbf{Model Response:} The image features a brown and white cat lying on its side in an \textcolor{orange}{orange suitcase}. The cat appears to be relaxed and comfortable in the suitcase. The suitcase is placed on a \textcolor{red}{bed}. The suitcase is open, revealing its contents, which include a \textcolor{green}{book} and a \textcolor{orange}{handbag}. The cat is the main focus of the image, occupying a significant portion of the frame. The orange suitcase and the handbag are the other main objects in the scene, with the handbag being located on the left side of the image and the suitcase occupying \textcolor{orange}{the right side}. The overall mood of the image is calm and cozy.
        \end{minipage}
    };

    \node[combined_block] (lower_block) at (0,-5) {
        \begin{minipage}[t]{0.48\textwidth}
            \textbf{Model Response:} The image features a large, colorful truck parked on the side of a street. The truck is parked next to a brick building, and \textcolor{orange}{there is a car parked behind} it.

            There are several \textcolor{red}{people} in the scene, with one \textcolor{red}{person} standing near the truck and \textcolor{red}{two others} further away. \textcolor{red}{One of the individuals} is carrying a \textcolor{green}{handbag}, which is visible in the scene. The overall atmosphere of the image is lively and engaging, with the eye-catching truck.
        \end{minipage}
        \hspace{10pt}
        \raisebox{-0.7\height}{\includegraphics[width=4.5cm, height=4cm]{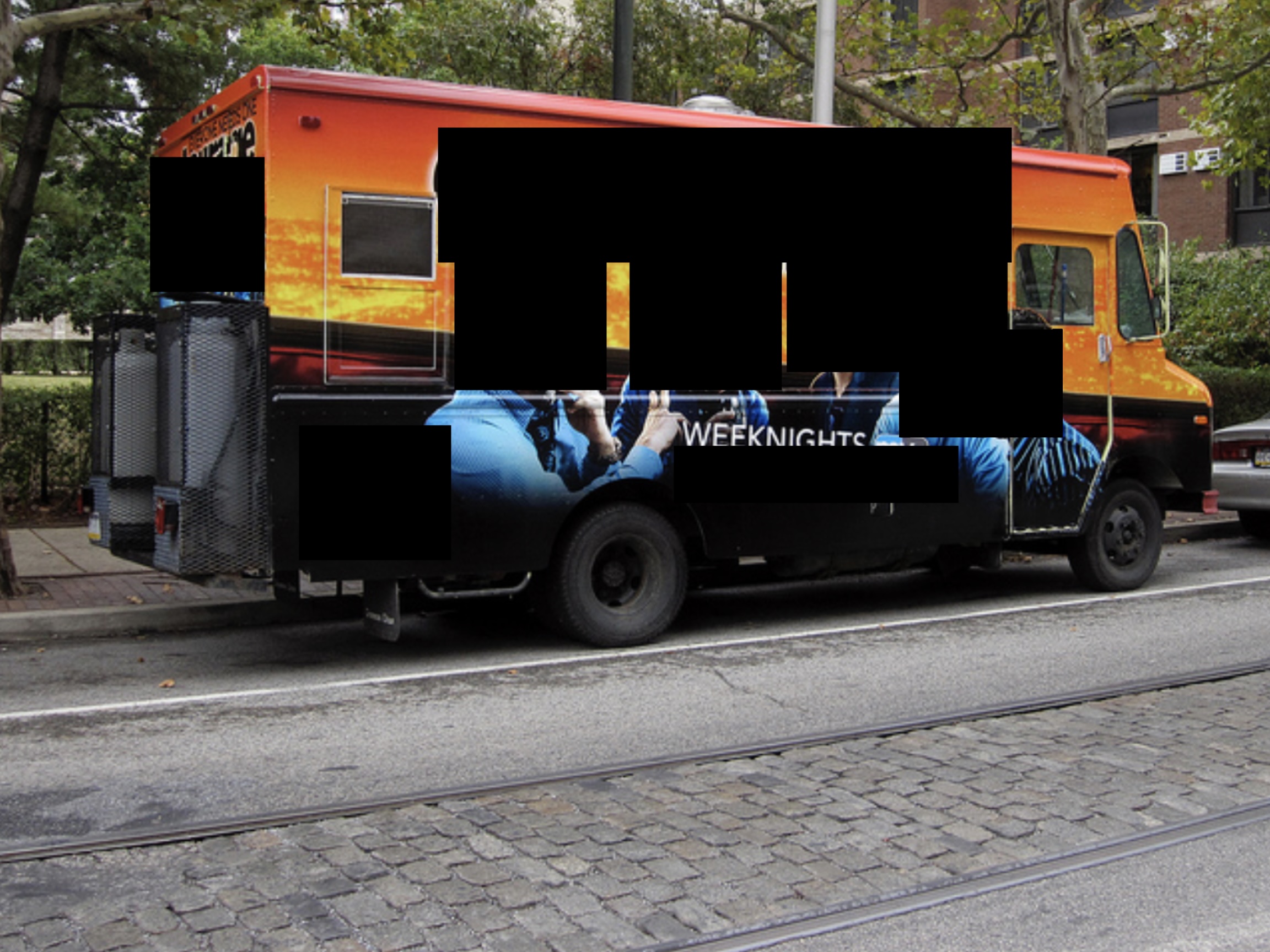}}
    };

\end{tikzpicture}

%% file: figures/mmhal.tex
\begin{tikzpicture}[font=\fontsize{7pt}{7pt}\selectfont]

\tikzstyle{combined_block} = [rectangle, draw, rounded corners, inner sep=10pt, align=left, minimum width=10cm]

    \node[combined_block] (upper_block) at (0,0) {
    \raisebox{-0.7\height}{image ID: \href{https://huggingface.co/datasets/Shengcao1006/MMHal-Bench/blob/main/test_data.zip}{16189396430\_4dce91a9d7\_o}}
        \hspace{10pt}
        \begin{minipage}[t]{0.48\textwidth}
            \textbf{Prompt:} How many people are there in the image?\\
            \textbf{MMHAL reference response:} There are four people in the image.\\
            \textbf{MMHAL categories}: Girl, Human head, Human hair, Mammal, Dog, Person, Human face\\
            \textbf{LLaVA 1.6-7B aligned:} There are four people in the image: a man, a woman, and two children.\\~\\
            \textbf{GPT-4 response:} total score=1, hallucination=1, helpfulness=1
        \end{minipage}
    };

    \node[combined_block] (middle_block) at (0,-3.3) {
        \raisebox{-0.7\height}{image ID: \href{https://huggingface.co/datasets/Shengcao1006/MMHal-Bench/blob/main/test_data.zip}{12392564684\_ffd19b5b55\_o}}
        \hspace{10pt}
        \begin{minipage}[t]{0.48\textwidth}
            \textbf{Prompt :} Compare the two different views of the backpack.\\
            \textbf{MMHAL reference response:} The image shows two different views of the same black backpack. On the left side, it shows the inside of the bag, containing a camera, two lenses an other accessories. On the right side, it shows the outside of the bag, and there is a "Lowepro" logo on it.\\
            \textbf{MMHAL categories}: Hiking equipment, Clothing, Backpack\\
            \textbf{Aligned LLaVA 1.6-7B response:} The image shows a backpack with two different views: the top view and the side view. In the top view, we can see (...) \\~\\
            \textbf{GPT-4 response:} total score=4, hallucination=0, helpfulness=1
        \end{minipage}
    };

\end{tikzpicture}

%% file: figures/plot_impact_size_llava.tex
{
\pgfplotsset{compat=newest}
\begin{tikzpicture}
  \begin{axis}[
    xlabel=Preference training set size,
    ylabel style={align=center}, ylabel=Llava-Bench-in-the-Wild $\uparrow$,
    grid=both,
    grid style={line width=.1pt, draw=gray!10},
    major grid style={line width=.2pt,draw=gray!50},
    minor tick num=5,
    axis line style={latex-latex},
    xmin=-100,
    xmax=5500,
    ymin=73,
    ymax=88,
    height=1\textwidth,
    width=1\textwidth,
    xtick={100,500,1000,2000,3000,4000,5000},
    xticklabels={100,500,1000,2000,3000,4000,5000},
    xticklabel style={font=\normalsize},
    tick label style={font=\normalsize},
        yticklabel style = {font=\normalsize},
        xticklabel style = {font=\normalsize},
        ylabel style = {font=\normalsize},
        xlabel style = {font=\normalsize},
    xticklabel style={rotate=45},
    xlabel shift=-3pt,
    legend style={at={(0.35,0.295)}, anchor=north, legend cell align=left, align=left, draw=white!15!black, font=\tiny},
    ]
    \addplot +[line width=1.5pt,color=blue,mark options={fill=blue}] coordinates {
        (100,	79.55)
        (500,	81.031)
        (1000,	79.231)
        (2000,	77.896)
        (3000,	81.044)
        (4000,	80.476)
        (5000,	81.792)
    };
    \addlegendentry{POVID}
    \addplot +[line width=1.5pt,color=green,mark options={fill=green}] coordinates {
        (100,	81.092)
        (500,	79.883)
        (1000,  81.456)
        (2000,  80.294)
        (3000,  80.784)
        (5000,  82.91)
    };
    \addlegendentry{VLFeedback}
    \addplot +[line width=1.5pt,color=orange,mark options={fill=orange}] coordinates {
        (100,	79.63)
        (500,	79.167)
        (1000,  82.171)
        (2000,  81.274)
        (3000,  86.381)
        (4000,  83.301)
        (5000,  87.50)
    };
    \addlegendentry{VLFeedbackCorrupted (Ours)}
    \addplot [dashed, line width=1.0pt, draw=gray] coordinates {(0,80.85) (6000,80.85)};
    \addlegendentry{LLaVA 1.6-7B}
  \end{axis}
\end{tikzpicture}\\
}

%% file: figures/plot_impact_size_recall.tex
{
\pgfplotsset{compat=newest}
\begin{tikzpicture}
  \begin{axis}[
    xlabel=Preference training set size,
    ylabel style={align=center}, ylabel=Recall$^{\text{coco}}\uparrow$,
    grid=both,
    grid style={line width=.1pt, draw=gray!10},
    major grid style={line width=.2pt,draw=gray!50},
    minor tick num=5,
    axis line style={latex-latex},
    xmin=-100,
    xmax=5500,
    ymax=73,
    ymin=62.0,
    height=1\textwidth,
    width=1\textwidth,
    xtick={100,500,1000,2000,3000,4000,5000},
    xticklabels={100,500,1000,2000,3000,4000,5000},
    xticklabel style={font=\normalsize},
    tick label style={font=\normalsize},
        yticklabel style = {font=\normalsize},
        xticklabel style = {font=\normalsize},
        ylabel style = {font=\normalsize},
        xlabel style = {font=\normalsize},
    xticklabel style={rotate=45},
    xlabel shift=-3pt,
    legend style={at={(0.35,0.295)}, anchor=north, legend cell align=left, align=left, draw=white!15!black, font=\tiny},
    ]
    \addplot +[line width=1.5pt,color=blue,mark options={fill=blue}] coordinates {
        (100,	68.34)
        (500,	67.71)
        (1000,	65.41)
        (2000,	69.29)
        (3000,	70.23)
        (4000,	71.70)
        (5000,	72.43)
    };
    \addlegendentry{POVID}
    \addplot +[line width=1.5pt,color=green,mark options={fill=green}] coordinates {
        (100,	67.09)
        (500,	67.61)
        (1000,  67.40)
        (2000,  67.19)
        (3000,  68.55)
        (5000,  67.19)
    };
    \addlegendentry{VLFeedback}
    \addplot +[line width=1.5pt,color=orange,mark options={fill=orange}] coordinates {
        (100,	67.40)
        (500,	67.82)
        (1000,  68.34)
        (2000,  68.87)
        (3000,  68.34)
        (4000,  69.29)
        (5000,  67.61)
    };
    \addlegendentry{VLFeedbackCorrupted (Ours)}
    \addplot [dashed, line width=1.0pt, draw=gray] coordinates {(0,68.13) (6000,68.13)};
    \addlegendentry{LLaVA 1.6-7B}
  \end{axis}
\end{tikzpicture}\\
}

%% file: figures/plot_impact_size_helpfulness.tex
{
\pgfplotsset{compat=newest}
\begin{tikzpicture}
  \begin{axis}[
    xlabel=Preference training set size,
    ylabel style={align=center}, ylabel=MMHALBench Helpfulness rate$\uparrow$,
    grid=both,
    grid style={line width=.1pt, draw=gray!10},
    major grid style={line width=.2pt,draw=gray!50},
    minor tick num=5,
    axis line style={latex-latex},
    xmin=-100,
    xmax=5500,
    ymax=1.4,
    ymin=1.1,
    height=1\textwidth,
    width=1\textwidth,
    xtick={100,500,1000,2000,3000,4000,5000},
    xticklabels={100,500,1000,2000,3000,4000,5000},
    xticklabel style={font=\normalsize},
    tick label style={font=\normalsize},
        yticklabel style = {font=\normalsize},
        xticklabel style = {font=\normalsize},
        ylabel style = {font=\normalsize},
        xlabel style = {font=\normalsize},
    xticklabel style={rotate=45},
    xlabel shift=-3pt,
    legend style={at={(0.35,0.295)}, anchor=north, legend cell align=left, align=left, draw=white!15!black, font=\tiny},
    ]
    \addplot +[line width=1.5pt,color=blue,mark options={fill=blue}] coordinates {
        (100,	1.24)
        (500,	1.23)
        (1000,	1.28)
        (2000,	1.29)
        (3000,	1.21)
        (4000,	1.24)
        (5000,	1.20)
    };
    \addlegendentry{POVID}
    \addplot +[line width=1.5pt,color=green,mark options={fill=green}] coordinates {
        (100,	1.25)
        (500,	1.29)
        (1000,  1.26)
        (2000,  1.25)
        (3000,  1.30)
        (5000,  1.38)
    };
    \addlegendentry{VLFeedback}
    \addplot +[line width=1.5pt,color=orange,mark options={fill=orange}] coordinates {
        (100,	1.24)
        (500,	1.28)
        (1000,  1.25)
        (2000,  1.29)
        (3000,  1.36)
        (4000,  1.34)
        (5000,  1.36)
    };
    \addlegendentry{VLFeedbackCorrupted (Ours)}
    \addplot [dashed, line width=1.0pt, draw=gray] coordinates {(0,1.271) (6000,1.271)};
    \addlegendentry{LLaVA 1.6-7B}
  \end{axis}
\end{tikzpicture}\\
}

%% file: figures/plot_impact_size_hallucinations.tex
{
\pgfplotsset{compat=newest}
\begin{tikzpicture}
  \begin{axis}[
    xlabel=Preference training set size,
    ylabel style={align=center}, ylabel=MMHALBench Hallucination rate$\downarrow$,
    grid=both,
    grid style={line width=.1pt, draw=gray!10},
    major grid style={line width=.2pt,draw=gray!50},
    minor tick num=5,
    axis line style={latex-latex},
    xmin=-100,
    xmax=5500,
    ymax=0.5,
    ymin=0.3,
    height=1\textwidth,
    width=1\textwidth,
    xtick={100,500,1000,2000,3000,4000,5000},
    xticklabels={100,500,1000,2000,3000,4000,5000},
    xticklabel style={font=\normalsize},
    tick label style={font=\normalsize},
        yticklabel style = {font=\normalsize},
        xticklabel style = {font=\normalsize},
        ylabel style = {font=\normalsize},
        xlabel style = {font=\normalsize},
    xticklabel style={rotate=45},
    xlabel shift=-3pt,
    legend style={at={(0.35,0.295)}, anchor=north, legend cell align=left, align=left, draw=white!15!black, font=\tiny},
    ]
    \addplot +[line width=1.5pt,color=blue,mark options={fill=blue}] coordinates {
        (100,	0.45)
        (500,	0.44)
        (1000,	0.43)
        (2000,	0.44)
        (3000,	0.46)
        (4000,	0.41)
        (5000,	0.44)
    };
    \addlegendentry{POVID}
    \addplot +[line width=1.5pt,color=green,mark options={fill=green}] coordinates {
        (100,	0.42)
        (500,	0.43)
        (1000,  0.41)
        (2000,  0.47)
        (3000,  0.42)
        (5000,  0.46)
    };
    \addlegendentry{VLFeedback}
    \addplot +[line width=1.5pt,color=orange,mark options={fill=orange}] coordinates {
        (100,	0.42)
        (500,	0.44)
        (1000,  0.44)
        (2000,  0.44)
        (3000,  0.36)
        (4000,  0.35)
        (5000,  0.35)
    };
    \addlegendentry{VLFeedbackCorrupted (Ours)}
    \addplot [dashed, line width=1.0pt, draw=gray] coordinates {(0,0.4375) (6000,0.4375)};
    \addlegendentry{LLaVA 1.6-7B}
  \end{axis}
\end{tikzpicture}\\
}

%% file: tables/annotator_prompt.tex
\begin{table}[h]
\centering
\begin{minipage}{\textwidth}
\begin{tcolorbox}[colback=white!95!gray, colframe=black]
The most important part of this task is to choose a response that contains less hallucination. Everything in the answer should be based on the contents of the image. You are given an image, a question, and two responses. If the context is about something practical, a helpful response is a concise response, and not one with irrelevant questions and comments. You are an expert annotator, and you rate the answer with less hallucination and more helpful information about the image as the better answer. Less hallucination means every object or attribute of the object, like color and relationship, is described accurately and as it appears in the image. If something is ambiguous in the image, the answer should avoid including any details that are not clearly visible in the image. In your response, you should generate an answer where you indicate whether Response 1 or Response 2 is better and explain the reason.
\end{tcolorbox}
\captionof{table}{The Prompt Used for the Annotator.}
\label{table:online_prompt}
\end{minipage}
\end{table}

%% file: figures/annotator.tex
\begin{tikzpicture}[font=\fontsize{7pt}{7pt}\selectfont]

\tikzstyle{block} = [rectangle, draw, minimum width=4cm, minimum height=5cm, rounded corners, inner sep=10pt, label={[anchor=north, inner sep=3pt, align=center, yshift=0pt]north:\textbf{Prompt}~~How many people are in the pool?}]
\tikzstyle{text_block} = [rectangle, draw, minimum width=3cm, text width=6cm, rounded corners, align=justify, inner sep=5pt]

    \node[block, fill=white] (img_block) at (0,0) {
        \includegraphics[width=5cm, height=4cm]{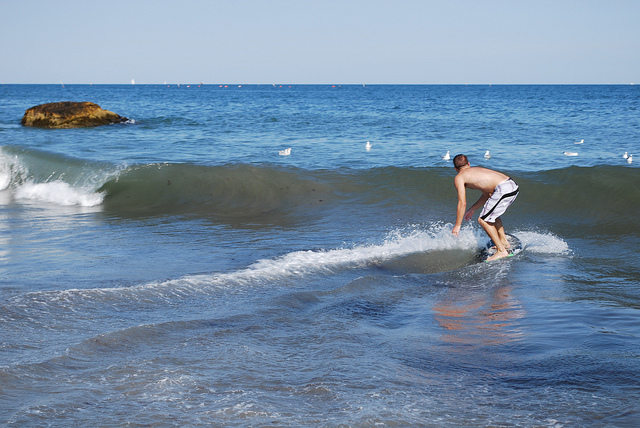} 
    };
    \node[text_block] (text1) at (6.25cm, 1.8cm) {\textbf{Response 1:} There is a person in the pool, a young man enjoying his time surfing the small waves.};
    \node[text_block, below=0.15cm of text1] (text2) {\textbf{Response 2:} This is not a swimming pool but the ocean, and on the ocean, there is a person. A young man is enjoying his time surfing.};
    \node[text_block, below=0.15cm of text2] (text3) {\textbf{Annotator:} Based on the image content, Response 2 is better and more helpful. The image shows a person surfing in the ocean, not a swimming pool. The surfer is riding a wave, and there are no other people visible in the image. The description accurately reflects the content of the image and does not include any hallucinations or irrelevant details.};
\end{tikzpicture}

%% file: figures/diffusion_beta.tex
\begin{tikzpicture}

\begin{axis}[
height=3cm,
font=\scriptsize,
width=0.9\textwidth,
yticklabel style={
    /pgf/number format/fixed,
    /pgf/number format/precision=5
},
scaled y ticks=false,
tick align=outside,
tick pos=left,
axis lines = left,
x grid style={gray},
xlabel={Step $k$},
xmajorgrids,
xmin=0, xmax=1000,
xtick style={color=black},
y grid style={gray},
ylabel={$\beta_k$},
ymajorgrids,
ymin=0, ymax=0.006,
ytick style={color=black}
]
\addplot [thick, black, each nth point={9}] 
table {%
0 2.23383885895601e-05
1 2.2487118258141e-05
2 2.26376359933056e-05
3 2.27899618039373e-05
4 2.29441211558878e-05
5 2.31001322390512e-05
6 2.32580205192789e-05
7 2.34178060054546e-05
8 2.35795123444404e-05
9 2.37431595451199e-05
10 2.39087748923339e-05
11 2.40763802139554e-05
12 2.42460009758361e-05
13 2.44176571868593e-05
14 2.45913779508555e-05
15 2.47671850956976e-05
16 2.49451022682479e-05
17 2.51251585723367e-05
18 2.53073776548263e-05
19 2.54917849815683e-05
20 2.56784041994251e-05
21 2.58672680502059e-05
22 2.60583983617835e-05
23 2.62518224189989e-05
24 2.64475693256827e-05
25 2.66456663666759e-05
26 2.68461408268195e-05
27 2.70490199909545e-05
28 2.72543366008904e-05
29 2.74621143034892e-05
30 2.76723894785391e-05
31 2.78851930488599e-05
32 2.81005322904093e-05
33 2.83184708678164e-05
34 2.85390106000705e-05
35 2.8762211513822e-05
36 2.89880772470497e-05
37 2.9216655093478e-05
38 2.94479705189588e-05
39 2.96820635412587e-05
40 2.99189596262295e-05
41 3.01586969726486e-05
42 3.04013083223253e-05
43 3.06468282360584e-05
44 3.08952876366675e-05
45 3.11467229039408e-05
46 3.14011740556452e-05
47 3.165867383359e-05
48 3.19192549795844e-05
49 3.2182961149374e-05
50 3.24498250847682e-05
51 3.27198831655551e-05
52 3.29931717715226e-05
53 3.32697381963953e-05
54 3.35496115440037e-05
55 3.38328354700934e-05
56 3.41194536304101e-05
57 3.44094951287843e-05
58 3.47030108969193e-05
59 3.50000336766243e-05
60 3.53006180375814e-05
61 3.56047894456424e-05
62 3.59126024704892e-05
63 3.62240971298888e-05
64 3.65393134416081e-05
65 3.68583023373503e-05
66 3.71811038348824e-05
67 3.75077579519711e-05
68 3.78383192582987e-05
69 3.81728314096108e-05
70 3.85113380616531e-05
71 3.88538901461288e-05
72 3.92005313187838e-05
73 3.95513125113212e-05
74 3.99062846554443e-05
75 4.02654914068989e-05
76 4.06289946113247e-05
77 4.0996827010531e-05
78 4.13690540881362e-05
79 4.17457158619072e-05
80 4.21268778154626e-05
81 4.25125836045481e-05
82 4.29028877988458e-05
83 4.32978486060165e-05
84 4.36975169577636e-05
85 4.41019474237692e-05
86 4.45112018496729e-05
87 4.49253311671782e-05
88 4.534441177384e-05
89 4.5768447307637e-05
90 4.61975723737851e-05
91 4.66317724203691e-05
92 4.70711747766472e-05
93 4.75157721666619e-05
94 4.79657028336078e-05
95 4.8420944949612e-05
96 4.88816440338269e-05
97 4.93477782583795e-05
98 4.98194967804011e-05
99 5.02967886859551e-05
100 5.07797922182363e-05
101 5.12684964633081e-05
102 5.17630542162806e-05
103 5.22634363733232e-05
104 5.27698139194399e-05
105 5.32821759406943e-05
106 5.38006315764505e-05
107 5.4325231758412e-05
108 5.485606379807e-05
109 5.5393193179043e-05
110 5.59366962988861e-05
111 5.64866422791965e-05
112 5.70431111555081e-05
113 5.76061793253757e-05
114 5.81759195483755e-05
115 5.87524118600413e-05
116 5.9335732657928e-05
117 5.99259619775694e-05
118 6.05231762165204e-05
119 6.11274736002088e-05
120 6.1738901422359e-05
121 6.23575833742507e-05
122 6.29835776635446e-05
123 6.36169716017321e-05
124 6.42578525003046e-05
125 6.49063149467111e-05
126 6.55624462524429e-05
127 6.62263191770762e-05
128 6.68980355840176e-05
129 6.75776827847585e-05
130 6.82653480907902e-05
131 6.89611333655193e-05
132 6.96651259204373e-05
133 7.0377420342993e-05
134 7.10981257725507e-05
135 7.18273076927289e-05
136 7.25650897948071e-05
137 7.33115593902767e-05
138 7.40668183425441e-05
139 7.48309685150161e-05
140 7.56041117710993e-05
141 7.63863645261154e-05
142 7.71777995396405e-05
143 7.79785477789119e-05
144 7.87886965554208e-05
145 7.96083768364042e-05
146 8.04376904852688e-05
147 8.12767393654212e-05
148 8.2125625340268e-05
149 8.2984464825131e-05
150 8.38533887872472e-05
151 8.47325136419386e-05
152 8.56219121487811e-05
153 8.65217880345881e-05
154 8.74321267474443e-05
155 8.83531902218238e-05
156 8.92849639058113e-05
157 9.02277097338811e-05
158 9.11814131541178e-05
159 9.21463215490803e-05
160 9.31224494706839e-05
161 9.41100443014875e-05
162 9.51090914895758e-05
163 9.61198675213382e-05
164 9.71423723967746e-05
165 9.81768680503592e-05
166 9.92233399301767e-05
167 0.000100282064522617
168 0.00010135310003534
169 0.000102436606539413
170 0.000103532715002075
171 0.000104641570942476
172 0.000105763305327855
173 0.000106898078229278
174 0.000108046035165899
175 0.000109207307104953
176 0.000110382083221339
177 0.000111570450826548
178 0.000112772599095479
179 0.000113988658995368
180 0.000115218805149198
181 0.000116463183076121
182 0.00011772196739912
183 0.000118995274533518
184 0.000120283322758041
185 0.000121586228488013
186 0.000122904195450246
187 0.000124237325508147
188 0.000125585836940445
189 0.000126949889818206
190 0.000128329658764414
191 0.000129725303850137
192 0.000131136985146441
193 0.000132564920932055
194 0.0001340092276223
195 0.000135470138047822
196 0.000136947797727771
197 0.000138442395837046
198 0.00013995413610246
199 0.000141483164043166
200 0.000143029697937891
201 0.000144593941513449
202 0.000146176040288992
203 0.000147776227095164
204 0.000149394647451118
205 0.000151031548739411
206 0.000152687120134942
207 0.000154361521708779
208 0.000156054986291565
209 0.000157767761265859
210 0.000159499933943152
211 0.000161251809913665
212 0.000163023549248464
213 0.000164815370226279
214 0.000166627505677752
215 0.000168460159329697
216 0.000170313549460843
217 0.000172187879798003
218 0.000174083383171819
219 0.000176000263309106
220 0.000177938782144338
221 0.000179899114300497
222 0.000181881550815888
223 0.000183886266313493
224 0.000185913493623957
225 0.000187963480129838
226 0.000190036473213695
227 0.000192132705706172
228 0.000194252410437912
229 0.000196395820239559
230 0.000198563211597502
231 0.000200754788238555
232 0.000202970782993361
233 0.000205211530555971
234 0.000207477176445536
235 0.000209768040804192
236 0.000212084356462583
237 0.000214426399907097
238 0.000216794403968379
239 0.000219188630580902
240 0.000221609399886802
241 0.000224056886509061
242 0.000226531410589814
243 0.00022903319040779
244 0.000231562575208955
245 0.000234119754168205
246 0.000236705091083422
247 0.00023931878968142
248 0.000241961141000502
249 0.000244632450630888
250 0.000247332995058969
251 0.000250063079874963
252 0.000252822908805683
253 0.000255612860200927
254 0.000258433195995167
255 0.000261284207226709
256 0.000264166155830026
257 0.000267079361947253
258 0.000270024116616696
259 0.000273000769084319
260 0.000276009581284598
261 0.000279050815152004
262 0.000282124819932505
263 0.000285231915768236
264 0.0002883723936975
265 0.000291546573862433
266 0.000294754776405171
267 0.000297997234156355
268 0.000301274383673444
269 0.000304586486890912
270 0.000307933834847063
271 0.000311316805891693
272 0.000314735691063106
273 0.000318190839607269
274 0.000321682542562485
275 0.00032521114917472
276 0.000328776921378449
277 0.000332380295731127
278 0.000336021592374891
279 0.000339701044140384
280 0.000343419087585062
281 0.0003471759846434
282 0.000350972171872854
283 0.000354807911207899
284 0.000358683551894501
285 0.000362599443178624
286 0.000366555963410065
287 0.000370553403627127
288 0.000374592171283439
289 0.000378672528313473
290 0.000382794853067026
291 0.000386959552997723
292 0.000391166948247701
293 0.000395417388062924
294 0.000399711192585528
295 0.000404048711061478
296 0.00040843038004823
297 0.00041285646148026
298 0.000417327333707362
299 0.000421843404183164
300 0.000426405022153631
301 0.000431012449553236
302 0.000435666152043268
303 0.000440366478869691
304 0.000445113779278472
305 0.000449908373411745
306 0.000454750610515475
307 0.000459640868939459
308 0.000464579527033493
309 0.000469566934043542
310 0.000474603468319401
311 0.000479689508210868
312 0.000484825315652415
313 0.000490011356305331
314 0.00049524795031175
315 0.000500535476021469
316 0.000505874282680452
317 0.000511264603119344
318 0.000516707077622414
319 0.000522201822604984
320 0.000527749245520681
321 0.000533349753823131
322 0.000539003638550639
323 0.000544711365364492
324 0.000550473167095333
325 0.00055628945119679
326 0.000562160566914827
327 0.000568086805287749
328 0.000574068631976843
329 0.000580106338020414
330 0.000586200214456767
331 0.000592350726947188
332 0.000598557991907001
333 0.000604822591412812
334 0.000611144700087607
335 0.00061752472538501
336 0.000623963074758649
337 0.000630459864623845
338 0.000637015677057207
339 0.000643630686681718
340 0.000650305242743343
341 0.000657039578072727
342 0.000663834216538817
343 0.000670689274556935
344 0.000677605159580708
345 0.000684582162648439
346 0.000691620574798435
347 0.00069872074527666
348 0.000705882848706096
349 0.000713107292540371
350 0.000720394309610128
351 0.000727744249161333
352 0.000735157635062933
353 0.000742634118068963
354 0.000750174338463694
355 0.000757778529077768
356 0.000765447039157152
357 0.000773179926909506
358 0.000780977657996118
359 0.000788840290624648
360 0.000796768174041063
361 0.000804761482868344
362 0.000812820566352457
363 0.000820945540908724
364 0.000829136639367789
365 0.000837394152767956
366 0.000845718255732208
367 0.000854109006468207
368 0.000862566812429577
369 0.00087109167361632
370 0.00087968388106674
371 0.000888343609403819
372 0.00089707103325054
373 0.000905866269022226
374 0.000914729433134198
375 0.0009236607584171
376 0.000932660303078592
377 0.000941728241741657
378 0.000950864690821618
379 0.000960069766733795
380 0.00096934346947819
381 0.000978686148300767
382 0.000988097628578544
383 0.000997578026726842
384 0.00100712769199163
385 0.00101674627512693
386 0.00102643412537873
387 0.00103619124274701
388 0.00104601739440113
389 0.00105591292958707
390 0.00106587773188949
391 0.00107591180130839
392 0.00108601525425911
393 0.00109618785791099
394 0.00110642961226404
395 0.00111674075014889
396 0.00112712103873491
397 0.00113757024519145
398 0.00114808848593384
399 0.00115867576096207
400 0.0011693318374455
401 0.00118005683179945
402 0.00119085039477795
403 0.00120171229355037
404 0.00121264264453202
405 0.00122364109847695
406 0.00123470777180046
407 0.00124584219884127
408 0.00125704437959939
409 0.00126831408124417
410 0.00127965107094496
411 0.00129105523228645
412 0.00130252621602267
413 0.00131406367290765
414 0.00132566748652607
415 0.00133733754046261
416 0.00134907325264066
417 0.00136087508872151
418 0.00137274153530598
419 0.00138467259239405
420 0.00139666849281639
421 0.00140872853808105
422 0.00142085237894207
423 0.00143303954973817
424 0.0014452898176387
425 0.00145760283339769
426 0.00146997801493853
427 0.00148241512943059
428 0.00149491359479725
429 0.00150747282896191
430 0.00152009259909391
431 0.00153277243953198
432 0.0015455117681995
433 0.00155831000301987
434 0.00157116679474711
435 0.00158408132847399
436 0.00159705337136984
437 0.00161008222494274
438 0.00162316742353141
439 0.00163630803581327
440 0.00164950406178832
441 0.00166275433730334
442 0.00167605851311237
443 0.00168941589072347
444 0.00170282565522939
445 0.00171628757379949
446 0.0017298007151112
447 0.00174336426425725
448 0.00175697752274573
449 0.00177064002491534
450 0.0017843508394435
451 0.00179810950066894
452 0.00181191484443843
453 0.00182576628867537
454 0.00183966336771846
455 0.00185360491741449
456 0.0018675901228562
457 0.00188161840196699
458 0.00189568905625492
459 0.00190980068873614
460 0.00192395318299532
461 0.0019381451420486
462 0.00195237575098872
463 0.00196664454415441
464 0.00198095012456179
465 0.00199529179371893
466 0.00200966908596456
467 0.00202408037148416
468 0.00203852518461645
469 0.0020530023612082
470 0.00206751120276749
471 0.00208205077797174
472 0.00209661945700645
473 0.00211121700704098
474 0.00212584272958338
475 0.00214049476198852
476 0.0021551726385951
477 0.00216987542808056
478 0.00218460173346102
479 0.00219935062341392
480 0.00221412186510861
481 0.0022289133630693
482 0.00224372441880405
483 0.00225855456665158
484 0.0022734017111361
485 0.00228826585225761
486 0.00230314512737095
487 0.00231803930364549
488 0.00233294651843607
489 0.00234786630608141
490 0.00236279726959765
491 0.00237773847766221
492 0.00239268876612186
493 0.00240764720365405
494 0.00242261285893619
495 0.00243758410215378
496 0.00245256070047617
497 0.00246754055842757
498 0.00248252344317734
499 0.00249750772491097
500 0.00251249223947525
501 0.00252747652120888
502 0.00254245917312801
503 0.00255743926391006
504 0.0025724156294018
505 0.00258738710545003
506 0.00260235276073217
507 0.00261731119826436
508 0.00263226125389338
509 0.00264720292761922
510 0.00266213365830481
511 0.00267705367878079
512 0.00269196089357138
513 0.00270685460418463
514 0.00272173434495926
515 0.00273659825325012
516 0.00275144539773464
517 0.00276627507992089
518 0.00278108683414757
519 0.00279587809927762
520 0.0028106493409723
521 0.00282539869658649
522 0.00284012476913631
523 0.00285482732579112
524 0.0028695052023977
525 0.00288415746763349
526 0.00289878249168396
527 0.00291338027454913
528 0.00292794941924512
529 0.00294248876161873
530 0.00295699760317802
531 0.00297147454693913
532 0.00298591959290206
533 0.00300033064559102
534 0.00301470817066729
535 0.00302904960699379
536 0.00304335565306246
537 0.0030576242133975
538 0.00307185482233763
539 0.00308604678139091
540 0.00310019892640412
541 0.00311431079171598
542 0.00312838121317327
543 0.00314240949228406
544 0.00315639493055642
545 0.00317033659666777
546 0.00318423355929554
547 0.00319808488711715
548 0.00321189034730196
549 0.0032256490085274
550 0.00323935993947089
551 0.00325302220880985
552 0.00326663581654429
553 0.00328019936569035
554 0.00329371239058673
555 0.00330717372708023
556 0.00332058384083211
557 0.00333394133485854
558 0.00334724551066756
559 0.0033604959025979
560 0.00337369204498827
561 0.00338683230802417
562 0.00339991762302816
563 0.00341294636018574
564 0.00342591828666627
565 0.00343883293680847
566 0.00345168984495103
567 0.0034644880797714
568 0.00347722740843892
569 0.00348990713246167
570 0.00350252725183964
571 0.00351508636958897
572 0.00352758495137095
573 0.00354002183303237
574 0.00355239724740386
575 0.00356471003033221
576 0.00357696018181741
577 0.00358914746902883
578 0.00360127119347453
579 0.00361333135515451
580 0.00362532702274621
581 0.00363725842908025
582 0.00364912464283407
583 0.00366092659533024
584 0.00367266219109297
585 0.00368433236144483
586 0.00369593617506325
587 0.00370747386477888
588 0.00371894473209977
589 0.00373034877702594
590 0.00374168553389609
591 0.00375295523554087
592 0.00376415764912963
593 0.00377529230900109
594 0.00378635874949396
595 0.00379735720343888
596 0.00380828743800521
597 0.00381914968602359
598 0.00382994301617146
599 0.00384066812694073
600 0.00385132408700883
601 0.00386191159486771
602 0.00387242948636413
603 0.00388287892565131
604 0.00389325921423733
605 0.00390357011929154
606 0.00391381187364459
607 0.00392398471012712
608 0.00393408816307783
609 0.00394412223249674
610 0.00395408691838384
611 0.00396398268640041
612 0.00397380907088518
613 0.00398356560617685
614 0.00399325368925929
615 0.00400287238880992
616 0.00401242170482874
617 0.00402190210297704
618 0.00403131358325481
619 0.00404065614566207
620 0.00404993025586009
621 0.0040591349825263
622 0.00406827172264457
623 0.00407733954489231
624 0.00408633938059211
625 0.00409527029842138
626 0.004104133695364
627 0.00411292910575867
628 0.00412165652960539
629 0.00413031643256545
630 0.00413890834897757
631 0.00414743321016431
632 0.00415589101612568
633 0.00416428176686168
634 0.0041726054623723
635 0.00418086303398013
636 0.00418905401602387
637 0.00419717933982611
638 0.00420523853972554
639 0.00421323208138347
640 0.00422115949913859
641 0.00422902218997478
642 0.00423682015389204
643 0.00424455292522907
644 0.00425222143530846
645 0.0042598252184689
646 0.00426736567169428
647 0.00427484232932329
648 0.00428225565701723
649 0.0042896056547761
650 0.00429689278826118
651 0.00430411705747247
652 0.00431127892807126
653 0.00431837933138013
654 0.0043254173360765
655 0.00433239480480552
656 0.00433931080624461
657 0.00434616580605507
658 0.00435296026989818
659 0.00435969466343522
660 0.0043663689866662
661 0.00437298463657498
662 0.0043795402161777
663 0.00438603665679693
664 0.00439247488975525
665 0.00439885491505265
666 0.00440517719835043
667 0.00441144220530987
668 0.00441764900460839
669 0.00442379992455244
670 0.00442989310249686
671 0.00443593133240938
672 0.00444191321730614
673 0.00444783922284842
674 0.00445371074602008
675 0.00445952638983727
676 0.00446528848260641
677 0.00447099655866623
678 0.00447665015235543
679 0.00448225066065788
680 0.00448779808357358
681 0.0044932933524251
682 0.00449873507022858
683 0.00450412556529045
684 0.00450946437194943
685 0.00451475149020553
686 0.00451998878270388
687 0.00452517438679934
688 0.00453031016513705
689 0.00453539658337831
690 0.00454043317586184
691 0.0045454204082489
692 0.00455035921186209
693 0.00455524912104011
694 0.00456009153276682
695 0.00456488644704223
696 0.00456963339820504
697 0.00457433378323913
698 0.00457898760214448
699 0.0045835948549211
700 0.00458815647289157
701 0.00459267245605588
702 0.00459714373573661
703 0.00460156938061118
704 0.00460595125332475
705 0.00461028842255473
706 0.004614582285285
707 0.00461883284151554
708 0.00462304055690765
709 0.00462720543146133
710 0.00463132746517658
711 0.00463540758937597
712 0.00463944626972079
713 0.00464344397187233
714 0.00464740069583058
715 0.00465131597593427
716 0.00465519214048982
717 0.00465902779251337
718 0.0046628238633275
719 0.00466658035293221
720 0.00467029912397265
721 0.00467397784814239
722 0.00467761978507042
723 0.00468122307211161
724 0.00468478910624981
725 0.00468831742182374
726 0.00469180895015597
727 0.0046952641569078
728 0.00469868304207921
729 0.0047020660713315
730 0.00470541324466467
731 0.00470872549340129
732 0.00471200235188007
733 0.0047152447514236
734 0.00471845315769315
735 0.00472162757068872
736 0.00472476799041033
737 0.00472787534818053
738 0.00473094917833805
739 0.00473398994654417
740 0.00473699951544404
741 0.00473997555673122
742 0.00474292086437345
743 0.00474583357572556
744 0.00474871601909399
745 0.00475156679749489
746 0.00475438730791211
747 0.00475717661902308
748 0.00475993705913424
749 0.00476266676560044
750 0.00476536713540554
751 0.00476803863421082
752 0.00477068079635501
753 0.00477329455316067
754 0.0047758799046278
755 0.00477843731641769
756 0.00478096632286906
757 0.00478346878662705
758 0.00478594284504652
759 0.00478839036077261
760 0.00479081133380532
761 0.00479320576414466
762 0.00479557365179062
763 0.00479791546240449
764 0.00480023166164756
765 0.00480252271518111
766 0.00480478815734386
767 0.00480702891945839
768 0.00480924500152469
769 0.00481143686920404
770 0.00481360405683517
771 0.00481574749574065
772 0.00481786718592048
773 0.00481996312737465
774 0.00482203625142574
775 0.00482408609241247
776 0.00482611311599612
777 0.00482811871916056
778 0.00483010103926063
779 0.0048320610076189
780 0.00483400002121925
781 0.00483591621741652
782 0.00483781192451715
783 0.00483968621119857
784 0.00484154000878334
785 0.00484337285161018
786 0.00484518427401781
787 0.00484697613865137
788 0.00484874797984958
789 0.00485049979761243
790 0.00485223205760121
791 0.00485394522547722
792 0.00485563836991787
793 0.00485731288790703
794 0.00485896877944469
795 0.00486060511320829
796 0.00486222375184298
797 0.00486382376402617
798 0.00486540561541915
799 0.0048669702373445
800 0.00486851669847965
801 0.00487004593014717
802 0.00487155746668577
803 0.00487305223941803
804 0.00487452978268266
805 0.00487599056214094
806 0.00487743504345417
807 0.00487886276096106
808 0.00488027464598417
809 0.00488166976720095
810 0.00488304998725653
811 0.00488441390916705
812 0.00488576246425509
813 0.00488709611818194
814 0.00488841347396374
815 0.00488971639424562
816 0.00489100441336632
817 0.0048922779969871
818 0.0048935366794467
819 0.00489478092640638
820 0.00489601120352745
821 0.0048972275108099
822 0.00489842984825373
823 0.00489961775019765
824 0.00490079261362553
825 0.00490195397287607
826 0.00490310182794929
827 0.00490423664450645
828 0.00490535842254758
829 0.00490646716207266
830 0.00490756286308169
831 0.00490864692255855
832 0.00490971794351935
833 0.0049107763916254
834 0.00491182319819927
835 0.00491285743191838
836 0.00491388002410531
837 0.00491489097476006
838 0.00491588981822133
839 0.00491687748581171
840 0.00491785397753119
841 0.00491881836205721
842 0.0049197725020349
843 0.00492071453481913
844 0.00492164678871632
845 0.00492256786674261
846 0.0049234782345593
847 0.00492437789216638
848 0.00492526730522513
849 0.00492614647373557
850 0.00492701539769769
851 0.00492787454277277
852 0.00492872344329953
853 0.00492956209927797
854 0.00493039144203067
855 0.00493121100589633
856 0.00493202125653625
857 0.00493282172828913
858 0.00493361335247755
859 0.00493439612910151
860 0.00493516866117716
861 0.00493593327701092
862 0.00493668811395764
863 0.0049374345690012
864 0.00493817264214158
865 0.0049389018677175
866 0.00493962224572897
867 0.00494033470749855
868 0.00494103832170367
869 0.0049417344853282
870 0.00494242226704955
871 0.00494310166686773
872 0.00494377361610532
873 0.00494443764910102
874 0.00494509376585484
875 0.00494574196636677
876 0.0049463827162981
877 0.00494701601564884
878 0.00494764233008027
879 0.00494826072826982
880 0.00494887260720134
881 0.00494947703555226
882 0.0049500735476613
883 0.00495066400617361
884 0.00495124701410532
885 0.00495182396844029
886 0.00495239347219467
887 0.00495295692235231
888 0.00495351338759065
889 0.00495406333357096
890 0.00495460676029325
891 0.00495514366775751
892 0.00495567498728633
893 0.00495619885623455
894 0.00495671760290861
895 0.00495723029598594
896 0.00495773646980524
897 0.00495823659002781
898 0.00495873158797622
899 0.0049592200666666
900 0.00495970295742154
901 0.00496018072590232
902 0.00496065197512507
903 0.00496111810207367
904 0.00496157864108682
905 0.0049620340578258
906 0.00496248388662934
907 0.00496292905882001
908 0.00496336864307523
909 0.00496380217373371
910 0.00496423151344061
911 0.00496465526521206
912 0.00496507482603192
913 0.00496548879891634
914 0.0049658976495266
915 0.00496630277484655
916 0.00496670184656978
917 0.00496709672734141
918 0.00496748741716146
919 0.00496787298470736
920 0.00496825389564037
921 0.00496863061562181
922 0.00496900314465165
923 0.00496937101706862
924 0.00496973469853401
925 0.00497009372338653
926 0.00497044855728745
927 0.00497079966589808
928 0.00497114565223455
929 0.0049714888446033
930 0.00497182691469789
931 0.00497216172516346
932 0.00497249234467745
933 0.00497281877323985
934 0.00497314194217324
935 0.00497346092015505
936 0.00497377570718527
937 0.00497408723458648
938 0.00497439503669739
939 0.004974699113518
940 0.00497499946504831
941 0.0049752970226109
942 0.00497559038922191
943 0.0049758804962039
944 0.00497616734355688
945 0.00497644999995828
946 0.00497672986239195
947 0.0049770069308579
948 0.00497727980837226
949 0.0049775498919189
950 0.00497781671583652
951 0.00497808028012514
952 0.00497834105044603
953 0.0049785990267992
954 0.00497885327786207
955 0.00497910473495722
956 0.00497935293242335
957 0.00497959880158305
958 0.00497984094545245
959 0.00498008076101542
960 0.00498031778261065
961 0.00498055154457688
962 0.00498078297823668
963 0.00498101208359003
964 0.00498123746365309
965 0.00498146051540971
966 0.00498168170452118
967 0.00498189963400364
968 0.00498211430385709
969 0.00498232757672668
970 0.00498253805562854
971 0.00498274527490139
972 0.00498295109719038
973 0.00498315365985036
974 0.00498335435986519
975 0.00498355226591229
976 0.00498374830931425
977 0.0049839410930872
978 0.00498413294553757
979 0.00498432153835893
980 0.00498450826853514
981 0.00498469267040491
982 0.00498487474396825
983 0.00498505495488644
984 0.0049852323718369
985 0.00498540885746479
986 0.00498558161780238
987 0.00498575391247869
988 0.00498592341318727
989 0.00498609151691198
990 0.00498625682666898
991 0.00498642027378082
992 0.00498658232390881
993 0.00498674204573035
994 0.00498689990490675
995 0.00498705543577671
996 0.0049872100353241
997 0.00498736230656505
998 0.00498751224949956
999 0.00498766172677279
};
\end{axis}

\end{tikzpicture}

%% file: figures/bdhs_similarity.tex
\begin{tikzpicture}

\definecolor{crimson2143940}{RGB}{214,39,40}
\definecolor{darkgray176}{RGB}{176,176,176}
\definecolor{darkorange25512714}{RGB}{255,127,14}
\definecolor{forestgreen4416044}{RGB}{44,160,44}
\definecolor{lightgray204}{RGB}{204,204,204}
\definecolor{mediumpurple148103189}{RGB}{148,103,189}
\definecolor{sienna1408675}{RGB}{140,86,75}
\definecolor{steelblue31119180}{RGB}{31,119,180}

\newcommand{\figurewidth}{0.9\textwidth}
\newcommand{\figureheight}{4.5cm}

\begin{axis}[
height=\figureheight,
font=\scriptsize,
width=\figurewidth,
axis lines = left,
legend columns = 3,
legend pos = outer north,
legend style={draw=none,fill=none,legend cell align=left, column sep=4pt,at={(\figurewidth/2-1cm,\figureheight-1.5cm)}},
tick align=outside,
tick pos=left,
xtick=data,
x grid style={gray},
xlabel={BDHS Iteration},
xmajorgrids,
xmin=1, xmax=5,
xtick style={color=black},
y grid style={gray},
ylabel={\% Non-Similar Responses},
ymajorgrids,
ymin=75, ymax=100,
extra y ticks={75},
extra y tick labels={75},
ytick style={color=black}
]
\addplot [thick, forestgreen4416044]
table {%
1 80.84
2 87.18
3 89.14
4 90.22
5 93.38
};
\addlegendentry{$\text{BDHS}_{\text{noise}}, N{=}300$}
\addplot [thick, crimson2143940]
table {%
1 87.72
2 94.32
3 96.24
4 97.38
5 99.94
};
\addlegendentry{$\text{BDHS}_{\text{noise}}, N{=}500$}
\addplot [thick, steelblue31119180]
table {%
1 77.96
2 85.9
3 88.8
4 90.34
5 93.52
};
\addlegendentry{$\text{BDHS}_{\text{attn}}, \rho_{\text{th}}{=}0.98$}
\addplot [thick, darkorange25512714]
table {%
1 74.6
2 80.4
3 82.14
4 83
5 85.3
};
\addlegendentry{$\text{BDHS}_{\text{attn}}, \rho_{\text{th}}{=}0$}
\addplot [thick, mediumpurple148103189]
table {%
1 79.94
2 87.82
3 90.44
4 91.8
5 94.6
};
\addlegendentry{$\text{BDHS}_{\text{attn}}, \rho_{\text{th}}{=}0.99$}
\addplot [thick, dashed, sienna1408675]
table {%
1 82.34
2 88.18
3 90.08
4 90.82
5 92.68
};
\addlegendentry{$\text{BDHS}_{\text{attn}}, \rho_{\text{th}}{=}1$}
\end{axis}
\end{tikzpicture}

%% file: tables/llm_bias_appendix.tex
\begin{table*}[ht]
\centering
\resizebox{\textwidth}{!}{%
\begin{tabular}{c |c c c c c c c c}
\toprule
$\tilde{y}^-$ derived from policy                  & POPE$\uparrow$ & MMHAL$\uparrow$ & MMHAL$^V\uparrow$ & LLaVA$^\text{W}$$\uparrow$ & VQA$^T\uparrow$ & GQA$\uparrow$  & MMVet$\uparrow$ & Recall$^{\text{coco}}\uparrow$\\

\midrule
-- (Baseline)                                       & 86.40          & \textbf{2.95}   & 2.75      & 80.85                     & 64.85           & {64.23}        & \textbf{43.94}  & 68.13 \\
-- (Plain DPO)                                     & 88.18          & \uline{2.93}    & \textbf{2.93}      & 81.89                      & 64.90           & 64.34          & 43.39           & 71.80 \\

\midrule
Attention Masking, $\rho_\text{th}=0.98$           & 88.61          & 2.25            & 2.25              & 82.25                      & 64.92           & 64.04          & 42.75           & \textbf{77.46} \\
Attention Masking, $\rho_\text{th}=0.99$           & 88.70          & 2.52            & 2.51              & 86.08                      & 65.07           & 64.06          & 42.02           & \uline{77.04} \\
\midrule
$\text{BDHS}_{\text{attn}}$, $\rho_\text{th}=0.98$ & \textbf{88.80} & 2.56            & 2.68              & \textbf{86.54}             & 65.02           & 64.03          & 43.03           & 76.10 \\
$\text{BDHS}_{\text{attn}}$, $\rho_\text{th}=0.99$ & \uline{88.75}  & 2.61            & 2.71              & \uline{86.33}              & 65.07           & 63.97          & \uline{43.39}   & 75.58 \\
$\text{BDHS}_{\text{attn}}$, $\rho_\text{th}=1.00$ & 88.70          & 2.63    & \uline{2.80}     & 84.15                      & \textbf{65.18}  & 63.93          & 43.12           & 75.37 \\
\midrule
$\text{BDHS}_{\text{noise}}$, $N=100$              & 88.50          & 2.58            & 2.48              & 82.46                      & 64.96           & \textbf{64.34} & 40.14           & 75.47 \\
$\text{BDHS}_{\text{noise}}$, $N=200$              & 88.55          & 2.49            & 2.38              & 83.43                      & 65.10           & 64.24          & 38.76           & 74.53 \\
$\text{BDHS}_{\text{noise}}$, $N=300$              & 88.59          & 2.43            & 2.45              & 85.16                      & \uline{65.11}   & 64.18          & 40.69           & 76.10 \\
$\text{BDHS}_{\text{noise}}$, $N=400$              & 88.66          & 2.39            & 2.42              & 83.72                      & 65.09           & \uline{64.29}  & 40.41           & 75.16 \\
$\text{BDHS}_{\text{noise}}$, $N=500$              & 88.59          & 2.36            & 2.49              & 84.53                      & 65.05           & 64.14          & 41.38           & 75.16 \\
\bottomrule
\end{tabular}
}
\caption{Additional ablation results for Offline-BDHS. All results are based on LLaVA 1.6-7B, using DPO and the POVID (5k) sample for the source of images and prompts.
and prompt.}
\label{tab:additional_bdhs_ablation}
\end{table*}

%% file: main.bbl
\begin{thebibliography}{45}
\providecommand{\natexlab}[1]{#1}
\providecommand{\url}[1]{\texttt{#1}}
\expandafter\ifx\csname urlstyle\endcsname\relax
  \providecommand{\doi}[1]{doi: #1}\else
  \providecommand{\doi}{doi: \begingroup \urlstyle{rm}\Url}\fi

\bibitem[Ahmadian et~al.(2024)Ahmadian, Cremer, Gall{\'e}, Fadaee, Kreutzer, {\"U}st{\"u}n, and Hooker]{Ahmadian_2024}
Arash Ahmadian, Chris Cremer, Matthias Gall{\'e}, Marzieh Fadaee, Julia Kreutzer, Ahmet {\"U}st{\"u}n, and Sara Hooker.
\newblock Back to basics: Revisiting reinforce style optimization for learning from human feedback in llms.
\newblock \emph{arXiv preprint arXiv:2402.14740}, 2024.

\bibitem[Azar et~al.(2024)Azar, Guo, Piot, Munos, Rowland, Valko, and Calandriello]{Azar_2024}
Mohammad~Gheshlaghi Azar, Zhaohan~Daniel Guo, Bilal Piot, Remi Munos, Mark Rowland, Michal Valko, and Daniele Calandriello.
\newblock A general theoretical paradigm to understand learning from human preferences.
\newblock In \emph{International Conference on Artificial Intelligence and Statistics}, pp.\  4447--4455. PMLR, 2024.

\bibitem[Bai et~al.(2023)Bai, Bai, Yang, Wang, Tan, Wang, Lin, Zhou, and Zhou]{Bai_2023b}
Jinze Bai, Shuai Bai, Shusheng Yang, Shijie Wang, Sinan Tan, Peng Wang, Junyang Lin, Chang Zhou, and Jingren Zhou.
\newblock Qwen-vl: A versatile vision-language model for understanding, localization, text reading, and beyond.
\newblock 2023.

\bibitem[Chen et~al.(2023)Chen, Sikka, Cogswell, Ji, and Divakaran]{Chen_2023e}
Yangyi Chen, Karan Sikka, Michael Cogswell, Heng Ji, and Ajay Divakaran.
\newblock Dress: Instructing large vision-language models to align and interact with humans via natural language feedback.
\newblock \emph{arXiv preprint arXiv:2311.10081}, 2023.

\bibitem[Christiano et~al.(2017)Christiano, Leike, Brown, Martic, Legg, and Amodei]{Christiano_2017}
Paul~F Christiano, Jan Leike, Tom Brown, Miljan Martic, Shane Legg, and Dario Amodei.
\newblock Deep reinforcement learning from human preferences.
\newblock \emph{Advances in neural information processing systems}, 30, 2017.

\bibitem[Cui et~al.(2023)Cui, Zhou, Yang, Wu, Zhang, Zou, and Yao]{Cui_2023}
Chenhang Cui, Yiyang Zhou, Xinyu Yang, Shirley Wu, Linjun Zhang, James Zou, and Huaxiu Yao.
\newblock Holistic analysis of hallucination in gpt-4v (ision): Bias and interference challenges.
\newblock \emph{arXiv preprint arXiv:2311.03287}, 2023.

\bibitem[Deng et~al.(2024)Deng, Lu, Yin, Hu, Shen, Zou, Chang, and Wang]{deng2024enhancing}
Yihe Deng, Pan Lu, Fan Yin, Ziniu Hu, Sheng Shen, James Zou, Kai-Wei Chang, and Wei Wang.
\newblock Enhancing large vision language models with self-training on image comprehension.
\newblock \emph{arXiv preprint arXiv:2405.19716}, 2024.

\bibitem[Gao et~al.(2023{\natexlab{a}})Gao, Schulman, and Hilton]{Gao_2023}
Leo Gao, John Schulman, and Jacob Hilton.
\newblock Scaling laws for reward model overoptimization.
\newblock In \emph{International Conference on Machine Learning}, pp.\  10835--10866. PMLR, 2023{\natexlab{a}}.

\bibitem[Gao et~al.(2023{\natexlab{b}})Gao, Tow, Abbasi, Biderman, Black, DiPofi, Foster, Golding, Hsu, Le~Noac'h, Li, McDonell, Muennighoff, Ociepa, Phang, Reynolds, Schoelkopf, Skowron, Sutawika, Tang, Thite, Wang, Wang, and Zou]{eval-harness}
Leo Gao, Jonathan Tow, Baber Abbasi, Stella Biderman, Sid Black, Anthony DiPofi, Charles Foster, Laurence Golding, Jeffrey Hsu, Alain Le~Noac'h, Haonan Li, Kyle McDonell, Niklas Muennighoff, Chris Ociepa, Jason Phang, Laria Reynolds, Hailey Schoelkopf, Aviya Skowron, Lintang Sutawika, Eric Tang, Anish Thite, Ben Wang, Kevin Wang, and Andy Zou.
\newblock A framework for few-shot language model evaluation, 12 2023{\natexlab{b}}.
\newblock URL \url{https://zenodo.org/records/10256836}.

\bibitem[Guo et~al.(2024)Guo, Zhang, Liu, Liu, Khalman, Llinares, Rame, Mesnard, Zhao, Piot, et~al.]{Guo_2024}
Shangmin Guo, Biao Zhang, Tianlin Liu, Tianqi Liu, Misha Khalman, Felipe Llinares, Alexandre Rame, Thomas Mesnard, Yao Zhao, Bilal Piot, et~al.
\newblock Direct language model alignment from online ai feedback.
\newblock \emph{arXiv preprint arXiv:2402.04792}, 2024.

\bibitem[Hester et~al.(2018)Hester, Vecerik, Pietquin, Lanctot, Schaul, Piot, Horgan, Quan, Sendonaris, Osband, et~al.]{hester2018deep}
Todd Hester, Matej Vecerik, Olivier Pietquin, Marc Lanctot, Tom Schaul, Bilal Piot, Dan Horgan, John Quan, Andrew Sendonaris, Ian Osband, et~al.
\newblock Deep q-learning from demonstrations.
\newblock In \emph{Proceedings of the AAAI conference on artificial intelligence}, volume~32, 2018.

\bibitem[Hudson \& Manning(2019)Hudson and Manning]{Hudson_2019}
Drew~A Hudson and Christopher~D Manning.
\newblock Gqa: A new dataset for real-world visual reasoning and compositional question answering.
\newblock \emph{Conference on Computer Vision and Pattern Recognition (CVPR)}, 2019.

\bibitem[Li et~al.(2024)Li, Zhang, Zhang, Pu, Du, Dong, Liu, Zhang, Zhang, Li, and Liu]{Li_2024}
Bo~Li, Peiyuan Zhang, Kaichen Zhang, Fanyi Pu, Xinrun Du, Yuhao Dong, Haotian Liu, Yuanhan Zhang, Ge~Zhang, Chunyuan Li, and Ziwei Liu.
\newblock Lmms-eval: Accelerating the development of large multimodal models, March 2024.
\newblock URL \url{https://github.com/EvolvingLMMs-Lab/lmms-eval}.

\bibitem[Li et~al.(2023{\natexlab{a}})Li, Xie, Li, Chen, Wang, Chen, Yang, Wang, and Kong]{Li_2023d}
Lei Li, Zhihui Xie, Mukai Li, Shunian Chen, Peiyi Wang, Liang Chen, Yazheng Yang, Benyou Wang, and Lingpeng Kong.
\newblock Silkie: Preference distillation for large visual language models, 2023{\natexlab{a}}.

\bibitem[Li et~al.(2023{\natexlab{b}})Li, Du, Zhou, Wang, Zhao, and Wen]{Li_2023f}
Yifan Li, Yifan Du, Kun Zhou, Jinpeng Wang, Wayne~Xin Zhao, and Ji-Rong Wen.
\newblock Pope: Evaluating object hallucination in large vision-language models, 2023{\natexlab{b}}.

\bibitem[Lin et~al.(2014)Lin, Maire, Belongie, Hays, Perona, Ramanan, Doll{\'a}r, and Zitnick]{Lin_2014}
Tsung-Yi Lin, Michael Maire, Serge Belongie, James Hays, Pietro Perona, Deva Ramanan, Piotr Doll{\'a}r, and C~Lawrence Zitnick.
\newblock Microsoft coco: Common objects in context.
\newblock In \emph{European conference on computer vision}, pp.\  740--755. Springer, 2014.

\bibitem[Liu et~al.(2023{\natexlab{a}})Liu, Lin, Li, Wang, Yacoob, and Wang]{liu2023mitigating}
Fuxiao Liu, Kevin Lin, Linjie Li, Jianfeng Wang, Yaser Yacoob, and Lijuan Wang.
\newblock Mitigating hallucination in large multi-modal models via robust instruction tuning.
\newblock In \emph{The Twelfth International Conference on Learning Representations}, 2023{\natexlab{a}}.

\bibitem[Liu et~al.(2023{\natexlab{b}})Liu, Li, Li, and Lee]{Liu_2023b}
Haotian Liu, Chunyuan Li, Yuheng Li, and Yong~Jae Lee.
\newblock Llava-1.5: Improved baselines with visual instruction tuning.
\newblock \emph{arXiv preprint arXiv:2310.03744}, 2023{\natexlab{b}}.

\bibitem[Liu et~al.(2023{\natexlab{c}})Liu, Li, Wu, and Lee]{Liu_2023}
Haotian Liu, Chunyuan Li, Qingyang Wu, and Yong~Jae Lee.
\newblock Llava: Visual instruction tuning, 2023{\natexlab{c}}.

\bibitem[Liu et~al.(2024)Liu, Li, Li, Li, Zhang, Shen, and Lee]{Liu_2024b}
Haotian Liu, Chunyuan Li, Yuheng Li, Bo~Li, Yuanhan Zhang, Sheng Shen, and Yong~Jae Lee.
\newblock Llava-next: Improved reasoning, ocr, and world knowledge, January 2024.
\newblock URL \url{https://llava-vl.github.io/blog/2024-01-30-llava-next/}.

\bibitem[McKinzie et~al.(2024)McKinzie, Gan, Fauconnier, Dodge, Zhang, Dufter, Shah, Du, Peng, Weers, et~al.]{Mckinzie_2024}
Brandon McKinzie, Zhe Gan, Jean-Philippe Fauconnier, Sam Dodge, Bowen Zhang, Philipp Dufter, Dhruti Shah, Xianzhi Du, Futang Peng, Floris Weers, et~al.
\newblock Mm1: Methods, analysis \& insights from multimodal llm pre-training.
\newblock \emph{arXiv preprint arXiv:2403.09611}, 2024.

\bibitem[Ouyang et~al.(2022)Ouyang, Wu, Jiang, Almeida, Wainwright, Mishkin, Zhang, Agarwal, Slama, Ray, et~al.]{Ouyang_2022}
Long Ouyang, Jeffrey Wu, Xu~Jiang, Diogo Almeida, Carroll Wainwright, Pamela Mishkin, Chong Zhang, Sandhini Agarwal, Katarina Slama, Alex Ray, et~al.
\newblock Training language models to follow instructions with human feedback.
\newblock \emph{Advances in neural information processing systems}, 35:\penalty0 27730--27744, 2022.

\bibitem[Qian et~al.(2024)Qian, Zhang, Yang, and Gan]{qian2024easy}
Yusu Qian, Haotian Zhang, Yinfei Yang, and Zhe Gan.
\newblock How easy is it to fool your multimodal llms? an empirical analysis on deceptive prompts.
\newblock \emph{arXiv preprint arXiv:2402.13220}, 2024.

\bibitem[Rafailov et~al.(2023)Rafailov, Sharma, Mitchell, Ermon, Manning, and Finn]{Rafailov_2023}
Rafael Rafailov, Archit Sharma, Eric Mitchell, Stefano Ermon, Christopher~D. Manning, and Chelsea Finn.
\newblock Direct preference optimization: Your language model is secretly a reward model, 2023.

\bibitem[Reimers \& Gurevych(2019)Reimers and Gurevych]{Reimers_2019}
Nils Reimers and Iryna Gurevych.
\newblock Sentence-bert: Sentence embeddings using siamese bert-networks.
\newblock In \emph{Proceedings of the 2019 Conference on Empirical Methods in Natural Language Processing}. Association for Computational Linguistics, 11 2019.
\newblock URL \url{https://arxiv.org/abs/1908.10084}.

\bibitem[Rohrbach et~al.(2018)Rohrbach, Hendricks, Burns, Darrell, and Saenko]{Rohrbach_2018}
Anna Rohrbach, Lisa~Anne Hendricks, Kaylee Burns, Trevor Darrell, and Kate Saenko.
\newblock Object hallucination in image captioning.
\newblock In \emph{Proceedings of the 2018 Conference on Empirical Methods in Natural Language Processing}, pp.\  4035--4045, 2018.

\bibitem[Schulman et~al.(2017)Schulman, Wolski, Dhariwal, Radford, and Klimov]{Schulman_2017}
John Schulman, Filip Wolski, Prafulla Dhariwal, Alec Radford, and Oleg Klimov.
\newblock Proximal policy optimization algorithms.
\newblock \emph{arXiv preprint arXiv:1707.06347}, 2017.

\bibitem[Singh et~al.(2019)Singh, Natarajan, Shah, Jiang, Chen, Batra, Parikh, and Rohrbach]{Singh_2019}
Amanpreet Singh, Vivek Natarajan, Meet Shah, Yu~Jiang, Xinlei Chen, Dhruv Batra, Devi Parikh, and Marcus Rohrbach.
\newblock Towards vqa models that can read.
\newblock In \emph{Proceedings of the IEEE/CVF conference on computer vision and pattern recognition}, pp.\  8317--8326, 2019.

\bibitem[Stiennon et~al.(2020)Stiennon, Ouyang, Wu, Ziegler, Lowe, Voss, Radford, Amodei, and Christiano]{Stiennon_2020}
Nisan Stiennon, Long Ouyang, Jeffrey Wu, Daniel Ziegler, Ryan Lowe, Chelsea Voss, Alec Radford, Dario Amodei, and Paul~F Christiano.
\newblock Learning to summarize with human feedback.
\newblock \emph{Advances in Neural Information Processing Systems}, 33:\penalty0 3008--3021, 2020.

\bibitem[Sun et~al.(2023)Sun, Shen, Cao, Liu, Li, Shen, Gan, Gui, Wang, Yang, et~al.]{Sun_2023}
Zhiqing Sun, Sheng Shen, Shengcao Cao, Haotian Liu, Chunyuan Li, Yikang Shen, Chuang Gan, Liang-Yan Gui, Yu-Xiong Wang, Yiming Yang, et~al.
\newblock Llava rhlf: Aligning large multimodal models with factually augmented rlhf.
\newblock \emph{arXiv preprint arXiv:2309.14525}, 2023.

\bibitem[Tang et~al.(2024{\natexlab{a}})Tang, Guo, Zheng, Calandriello, Cao, Tarassov, Munos, Pires, Valko, Cheng, and Dabney]{Tang_2024}
Yunhao Tang, Daniel~Zhaohan Guo, Zeyu Zheng, Daniele Calandriello, Yuan Cao, Eugene Tarassov, R{\'e}mi Munos, Bernardo~{\'A}vila Pires, Michal Valko, Yong Cheng, and Will Dabney.
\newblock Understanding the performance gap between online and offline alignment algorithms, 2024{\natexlab{a}}.

\bibitem[Tang et~al.(2024{\natexlab{b}})Tang, Guo, Zheng, Calandriello, Munos, Rowland, Richemond, Valko, Ávila Pires, and Piot]{tang2024generalized}
Yunhao Tang, Zhaohan~Daniel Guo, Zeyu Zheng, Daniele Calandriello, Rémi Munos, Mark Rowland, Pierre~Harvey Richemond, Michal Valko, Bernardo Ávila Pires, and Bilal Piot.
\newblock Generalized preference optimization: A unified approach to offline alignment.
\newblock \emph{arXiv preprint arXiv:2402.05749}, 2024{\natexlab{b}}.

\bibitem[Touvron et~al.(2023)Touvron, Martin, Stone, Albert, Almahairi, Babaei, Bashlykov, Batra, Bhargava, Bhosale, et~al.]{Touvron_2023}
Hugo Touvron, Louis Martin, Kevin Stone, Peter Albert, Amjad Almahairi, Yasmine Babaei, Nikolay Bashlykov, Soumya Batra, Prajjwal Bhargava, Shruti Bhosale, et~al.
\newblock Llama 2: Open foundation and fine-tuned chat models.
\newblock \emph{arXiv preprint arXiv:2307.09288}, 2023.

\bibitem[Williams(1992)]{Williams_1992}
Ronald~J Williams.
\newblock Simple statistical gradient-following algorithms for connectionist reinforcement learning.
\newblock \emph{Machine learning}, 8:\penalty0 229--256, 1992.

\bibitem[Xu et~al.(2024)Xu, Fu, Gao, Ye, Liu, Mei, Wang, Yu, and Wu]{xu2024dpo}
Shusheng Xu, Wei Fu, Jiaxuan Gao, Wenjie Ye, Weilin Liu, Zhiyu Mei, Guangju Wang, Chao Yu, and Yi~Wu.
\newblock Is dpo superior to ppo for llm alignment? a comprehensive study.
\newblock \emph{arXiv preprint arXiv:2404.10719}, 2024.

\bibitem[Yu et~al.(2023{\natexlab{a}})Yu, Hu, Yao, Zhang, Zhao, Wang, Wang, Pan, Xue, Li, Liu, Zheng, and Sun]{Yu_2023b}
Tianyu Yu, Jinyi Hu, Yuan Yao, Haoye Zhang, Yue Zhao, Chongyi Wang, Shan Wang, Yinxv Pan, Jiao Xue, Dahai Li, Zhiyuan Liu, Hai-Tao Zheng, and Maosong Sun.
\newblock Reformulating vision-language foundation models and datasets towards universal multimodal assistants, 2023{\natexlab{a}}.

\bibitem[Yu et~al.(2023{\natexlab{b}})Yu, Yao, Zhang, He, Han, Cui, Hu, Liu, Zheng, Sun, and Chua]{Yu_2023}
Tianyu Yu, Yuan Yao, Haoye Zhang, Taiwen He, Yifeng Han, Ganqu Cui, Jinyi Hu, Zhiyuan Liu, Hai-Tao Zheng, Maosong Sun, and Tat-Seng Chua.
\newblock Rlhf-v: Towards trustworthy mllms via behavior alignment from fine-grained correctional human feedback, 2023{\natexlab{b}}.

\bibitem[Yu et~al.(2024)Yu, Zhang, Yao, Dang, Chen, Lu, Cui, He, Liu, Chua, et~al.]{Yu_2024}
Tianyu Yu, Haoye Zhang, Yuan Yao, Yunkai Dang, Da~Chen, Xiaoman Lu, Ganqu Cui, Taiwen He, Zhiyuan Liu, Tat-Seng Chua, et~al.
\newblock Rlaif-v: Aligning mllms through open-source ai feedback for super gpt-4v trustworthiness.
\newblock \emph{arXiv preprint arXiv:2405.17220}, 2024.

\bibitem[Yu et~al.(2023{\natexlab{c}})Yu, Yang, Li, Wang, Lin, Liu, Wang, and Wang]{Yu_2023c}
Weihao Yu, Zhengyuan Yang, Linjie Li, Jianfeng Wang, Kevin Lin, Zicheng Liu, Xinchao Wang, and Lijuan Wang.
\newblock Mm-vet: Evaluating large multimodal models for integrated capabilities.
\newblock \emph{arXiv preprint arXiv:2308.02490}, 2023{\natexlab{c}}.

\bibitem[Yuan et~al.(2024)Yuan, Pang, Cho, Sukhbaatar, Xu, and Weston]{Yuan_2024}
Weizhe Yuan, Richard~Yuanzhe Pang, Kyunghyun Cho, Sainbayar Sukhbaatar, Jing Xu, and Jason Weston.
\newblock Self-rewarding language models.
\newblock \emph{arXiv preprint arXiv:2401.10020}, 2024.

\bibitem[Zhang et~al.(2023)Zhang, Zhang, Gu, Zhou, Lipka, Yang, and Sun]{Zhang_2023b}
Yanzhe Zhang, Ruiyi Zhang, Jiuxiang Gu, Yufan Zhou, Nedim Lipka, Diyi Yang, and Tong Sun.
\newblock Llavar: Enhanced visual instruction tuning for text-rich image understanding.
\newblock \emph{arXiv preprint arXiv:2306.17107}, 2023.

\bibitem[Zhao et~al.(2023{\natexlab{a}})Zhao, Wu, and Huang]{Zhao_2023b}
Bo~Zhao, Boya Wu, and Tiejun Huang.
\newblock Svit: Scaling up visual instruction tuning, 2023{\natexlab{a}}.

\bibitem[Zhao et~al.(2023{\natexlab{b}})Zhao, Joshi, Liu, Khalman, Saleh, and Liu]{Zhao_2023e}
Yao Zhao, Rishabh Joshi, Tianqi Liu, Misha Khalman, Mohammad Saleh, and Peter~J Liu.
\newblock Slic-hf: Sequence likelihood calibration with human feedback.
\newblock \emph{arXiv preprint arXiv:2305.10425}, 2023{\natexlab{b}}.

\bibitem[Zhou et~al.(2024)Zhou, Cui, Rafailov, Finn, and Yao]{Zhou_2024}
Yiyang Zhou, Chenhang Cui, Rafael Rafailov, Chelsea Finn, and Huaxiu Yao.
\newblock Aligning modalities in vision large language models via preference fine-tuning, 2024.

\bibitem[Zhu et~al.(2023)Zhu, Chen, Shen, Li, and Elhoseiny]{Zhu_2023c}
Deyao Zhu, Jun Chen, Xiaoqian Shen, Xiang Li, and Mohamed Elhoseiny.
\newblock Minigpt-4: Enhancing vision-language understanding with advanced large language models.
\newblock \emph{arXiv preprint arXiv:2304.10592}, 2023.

\end{thebibliography}
